\documentclass[10pt]{article} %
\usepackage[accepted]{tmlr}

\usepackage{amsmath,amsfonts,bm}

\def\eqref#1{equation~\ref{#1}}

\def\1{\bm{1}}

\DeclareMathAlphabet{\mathsfit}{\encodingdefault}{\sfdefault}{m}{sl}
\SetMathAlphabet{\mathsfit}{bold}{\encodingdefault}{\sfdefault}{bx}{n}

\usepackage{hyperref}
\usepackage{url}

\usepackage[bottom]{footmisc}

\usepackage{latexsym}
\usepackage{graphicx}
\usepackage{algorithm}
\usepackage{algorithmic}
\usepackage{verbatim}
\usepackage{newfloat}
\usepackage{listings}
\floatstyle{ruled}
\newfloat{listing}{tb}{lst}{}
\floatname{listing}{Listing}
\usepackage{microtype}
\usepackage{booktabs}
\usepackage{xcolor}
\usepackage{multirow}
\usepackage{amsmath}
\usepackage{dsfont}
\usepackage{subcaption}
\usepackage{adjustbox}
\usepackage{wrapfig} 

\definecolor{bestcolor}{RGB}{0,128,0}
\newcommand{\best}[1]{\textcolor{bestcolor}{\textbf{#1}}}

\usepackage{lineno}

\definecolor{darkblue}{rgb}{0, 0, 0.5}

\usepackage{cleveref}
\crefname{section}{Section}{Sections}

\usepackage{pdflscape}

\usepackage{tikz}

\lstdefinestyle{noindent}{
  basicstyle=\ttfamily\footnotesize,
  keywordstyle=\ttfamily,
  breaklines=true,
  aboveskip=5pt,
  belowskip=0pt,
  showspaces=false,
  showtabs=false,
  showstringspaces=false,
  xleftmargin=-\parindent,
  resetmargins=true,
  gobble=0
}
\newcommand{\itcomment}[1]{\textnormal{\itshape #1}}
\makeatletter
\providecommand{\needspace}[1]{\par\penalty-100\begingroup
  \setlength{\dimen@}{#1}\dimen@ii\pagegoal\advance\dimen@ii-\pagetotal
  \ifdim\dimen@>\dimen@ii\ifdim\dimen@ii>\z@\vfil\fi\break\fi\endgroup}
\makeatother
\lstdefinestyle{smtcode}{
  language=Python,
  basicstyle=\ttfamily\scriptsize,
  keywordstyle=\color{blue},
  morekeywords={True,False,None},
  stringstyle=\color{red!60!black},
  commentstyle=\color{green!45!black},
  numbers=none,
  xleftmargin=0pt,
  aboveskip=0pt,
  belowskip=0pt,
  showstringspaces=false,
  breaklines=false,
  columns=fullflexible,
  keepspaces=true,
  upquote=true,
  escapeinside={(*@}{@*)}
}

\usepackage{enumitem}
\usepackage{caption}

\Crefname{equation}{Eq.}{Eqs.}
\Crefname{figure}{Fig.}{Figs.}
\Crefname{tabular}{Tab.}{Tabs.}
\Crefname{section}
{Sec.}{Secs.}
\Crefname{appendix}
{App.}{Apps.}
\crefname{appendix}{app.}{apps.}
\Crefname{theorem}{Thm.}{Thms.}

\title{Uncertainty Quantification for LLM Function-Calling}

\author{\name Zihuiwen Ye\begin{NoHyper}%
\thanks{~Equal contribution, ~ $^+$ Joint senior authorship} 
\end{NoHyper} \\
\name Lukas Aichberger$^*$ \\
\addr University of Oxford
\ANDAUTHOR
\name Michael Kirchhof \\
\name Sinead Williamson \\
\name Luca Zappella \\
\addr Apple
\ANDAUTHOR
\name Yarin Gal \\
\addr University of Oxford
\ANDAUTHOR
\name Arno Blaas$^+$ \\
\name Adam Goli{\'n}ski$^+$ \\
\addr Apple 
}

\newcommand{\method}{\textbf{G-NLL}$_{SMT}$ }

\def\month{10}  %
\def\year{2026} %
\def\openreview{\url{https://openreview.net/forum?id=5ah8wJpDRt}} %

\definecolor{pink1}{RGB}{239, 156, 157}
\definecolor{yellow2}{RGB}{240, 227, 172}
\definecolor{green3}{RGB}{159, 209, 173}
\definecolor{blue4}{RGB}{166, 189, 219}
\definecolor{cyan5}{RGB}{187, 226, 244}
\definecolor{purple6}{RGB}{203, 158, 186}

\begin{document}

\maketitle

\begin{abstract}

Large Language Models (LLMs) are increasingly deployed to autonomously solve real-world tasks. 
A key ingredient for this is the \emph{LLM Function-Calling} paradigm, a widely used approach for equipping LLMs with tool-use capabilities.
However, an LLM calling functions incorrectly can have severe implications, especially when their effects are irreversible, e.g., transferring money or deleting data.
Hence, it is of paramount importance to consider the LLM's confidence that a function call solves the task correctly prior to executing it.
Uncertainty Quantification (UQ) methods can be used to quantify this confidence and prevent potentially incorrect function calls.
In this work, we present what is, to our knowledge, the first evaluation of UQ methods for LLM Function-Calling (FC).
While multi-sample UQ methods, such as Semantic Entropy, show strong performance for natural language Q\&A tasks, we find that in the FC setting the entropy-based multi-sample methods we evaluate offer no clear advantage over simple single-sample UQ methods.
Additionally, we find that the particularities of FC outputs can be leveraged to improve the performance of existing UQ methods in this setting.
Specifically, multi-sample UQ methods benefit from clustering FC outputs based on their abstract syntax tree parsing, while for single-sample UQ methods, restricting the logit-based score to semantically meaningful tokens leaves discrimination essentially unchanged but improves calibration and, by smaller margins, the utility of the score under abstention.

\end{abstract}

\section{Introduction}
\label{sec:introduction}

Large Language Models (LLMs) have undergone remarkable advancements in recent years, evolving from text generators to systems capable of interacting with their environment through API calls and function execution. 
This capability, commonly referred to as ``tool use'' \citep{schick2023toolformer, qin2024tooll, mialon2023augmented}, enables LLMs to perform actions ranging from simple information retrieval to complex operations that modify the state of the systems they interact with, such as performing monetary transactions or modifying databases.
Major language models \citep{achiam2023gpt, grattafiori2024llama, jiang2024mixtral, team2024gemini} 
now allow their users to integrate them with external tools, databases, and services through their \emph{Function-Calling} (FC) interface \citep{openai2025fc,gemini2025fc}. Such interfaces have become the most widespread approach for accessing the LLM's tool-use capabilities \citep{carrigan2024tool}.
The integration of tool-use capabilities represents a leap toward autonomous AI systems that can take actions on behalf of users. 

However, with this increased autonomy come increased risks and responsibilities. 
Unlike errors in text generation that may be less consequential, errors in tool use can result in irreversible changes, leading to data corruption, financial losses, or safety incidents \citep{andriushchenko2024, aichberger2025attacking}. In this context, Uncertainty Quantification (UQ) emerges as a critical component for responsible deployment of LLM-based systems with tool-use capabilities. 
UQ methods aim to estimate a model's confidence in its decisions,
enabling systems to abstain from actions or defer to human oversight when uncertainty is high. 
Effective UQ methods can serve as guardrails, preventing models from executing potentially harmful actions when they lack sufficient confidence. 

Despite the clear importance of UQ in FC scenarios, there is a notable absence of systematic evaluations for assessing UQ methods in this specific context, as well as of UQ methods developed specifically for it. 
While substantial research has addressed uncertainty estimation for classification and question-answering tasks \citep{ kuhn2023semantic, ye2024benchmarking, kirchhof2024pretrained, ye2025uncertaintyaware}, the tool-use setting introduces unique challenges. 
FC requires structured outputs that adhere to predefined syntax, often involving complex reasoning about parameter selection and the appropriateness of specific API calls. 
It is therefore unclear how existing UQ methods fare in this setting.

In this paper we aim to address this gap. We  conduct, to our knowledge, the first comprehensive evaluation of uncertainty quantification methods in LLM tool-use scenarios. 
To do so, we create a benchmark based upon the Berkeley Function Calling Leaderboard dataset (BFCL) \citep{bfcl,patil2025bfcl}, which provides a collection of FC examples, mostly in Python.
Our benchmark evaluates various UQ approaches based on how well their confidence estimates predict the correctness of function calls outputted by LLMs.
We reimplement several UQ methods, ranging from simple single-sample approaches based on model logits to more sophisticated multi-sample approaches that incorporate semantic equivalence between samples, and test how effectively they discriminate correct from incorrect function calls in a variety of setups.

Our findings indicate that in the FC setting, unlike in natural language Q\&A settings, the more involved entropy-based multi-sample methods we evaluate (predictive and semantic entropy) do not outperform simple single-sample UQ methods, such as the Greedy Negative Log-Likelihood (G-NLL).
We then investigate if the performance of existing UQ methods in the FC setting can be increased by tailoring them to the characteristics of FC outputs. By adapting the clustering strategy in Semantic Entropy and the token selection in G-NLL, we improve the discrimination of multi-sample UQ methods and the calibration of G-NLL.

\noindent In summary, our contributions are as follows:
\begin{enumerate}[leftmargin=*, topsep=0pt, itemsep=2pt, partopsep=0pt, parsep=0pt]
    \item We build the first benchmark for UQ methods in Function-Calling settings. 
    \item We find that
    Semantic Entropy (SE), the best-performing of the entropy-based multi-sample UQ methods we evaluate, offers no clear advantage over G-NLL as the best single-sample method.
    \item We analyze the properties of FC outputs
    and use them to devise variants of existing UQ methods that are tailored to the FC setting and improve their performance in this setting. Namely, we adapt the semantic-entailment method of SE to the FC setting by using Abstract Syntax Trees (AST) to cluster FC outputs, and adjust G-NLL and other single-sample UQ methods to the FC setting by defining semantically meaningful tokens and including only those in  calculating UQ scores. AST-based clustering consistently improves the multi-sample methods, while restricting single-sample scores to semantically meaningful tokens leaves their ranking performance essentially unchanged and instead improves the calibration of G-NLL and, to a smaller extent, its utility under abstention.
\end{enumerate}

\textbf{Artifact availability.} We release the benchmark harness (the BFCL-derived task splits and our selective-prediction evaluation) and our implementations of all UQ methods evaluated here (including the AST-based clustering and the semantically meaningful token extractors for both output formats): \url{https://github.com/daniellaye/uncertainty_tool_call/}.

\section{Background and Related Work}
\label{sec:related_work}

\subsection{LLM Tool-Use and Function-Calling}

A precursor of LLM tool-use, i.e., explicitly generating calls to external tools, which are then executed, is Retrieval-Augmented Generation \citep{lewis2020retri},
where the model is allowed to retrieve information from an external source (e.g., a database) before yielding a response.
Since then, various LLM tool-use approaches have been explored in the literature 
\citep{yao2023react,schick2023toolformer,qin2023tool}. The most widespread approach---which has been implemented in a very similar fashion in the majority of open- and closed-source LLMs
\citep{carrigan2024tool,openai2025fc,gemini2025fc}---is the Function-Calling (FC) interface.
In this framework, 
the LLM is explicitly provided with a list of functions in context.
Then, the LLM can call one or more of those functions to address the user's request.

To evaluate tool-use capabilities of LLMs, many benchmarks have been proposed 
\citep{bfcl,patil2025bfcl,qin2024tooll,li2023api-b,lu2024tools}.
However, all of them focus primarily on the accuracy-oriented performance of LLMs on these tool-use-related tasks. 
While accuracy is important, it is an orthogonal question to UQ, which tells us which function calls the LLM is uncertain about. 
This, in turn, can be used to refuse to execute certain  (potentially erroneous) function calls. 
This aspect, which is crucial from the reliability perspective, to the best of our knowledge, so far has been neglected in tool-use evaluation.

\subsection{Uncertainty Quantification Methods for Natural Language Generation}
\label{sec:uq-methods}

An established way to prevent incorrect model outputs before they can cause harm is to detect them via uncertainty estimates \citep{chow1957optim}. 
In problems like classification, where the output space is confined and well-structured, this can be as simple as returning the negative log-likelihood (NLL) of the predicted class %
\citep{galil2023what}. When working with language models, there are at least two additional challenges.
First, rather than a single class label with a single NLL value, we now have sequences of tokens, each with their own NLL value. Second, multiple distinct sequences can have the same semantic meaning: the probability of generating a correct response is the probability of generating any one of a large set of semantically equivalent correct responses \citep{kuhn2023semantic}.
In FC, such semantic equivalences might stem from, e.g., additional whitespaces or a different order of arguments. There are two main families of UQ methods that tackle these problems. Below we give an overview of both and refer to \Cref{app:uq} for more details.

The first family uses only a single greedily decoded sequence (which is likely close to being the highest-probability sequence), 
and aggregates its individual token uncertainties, in different ways: 
\textbf{MAX} reports the highest NLL of all predicted tokens (i.e., highest per-token uncertainty), 
\textbf{AVG} the average,
and \textbf{G-NLL} the sum \citep{aichberger2024rethi, fadeeva2023, vashurin2025benchmarkinguncertaintyquantificationmethods}. 
In addition to these,
we also evaluate the na\"{i}ve baseline \textbf{LEN} which reports the number of tokens of the greedy answer, as correctness metrics might spuriously correlate with shorter answers \citep{santilli2024on}. 
In case our benchmarking metrics were affected, it would be reflected in a high performance of this baseline.

The second family samples multiple sequences from the LLM to estimate the entropy of the LLM's distribution over whole sequences. 
A simple way of doing this is to directly estimate the Predictive Entropy (\textbf{PE}) of the distribution, based on the individual sequences' aggregated probabilities. 
However, this also captures uncertainty arising from sequences that are semantically identical. 
To account for this, Semantic Entropy (SE) \citep{kuhn2023semantic,farquhar2024detec} clusters samples that are different in form but have the same meaning, then estimates the entropy of the distribution of semantically different clusters. 
The original implementation of SE uses NLI-based entailment methods to cluster semantically-equivalent generations.
However, in initial experiments we found this to lead to bad results for the highly structured FC outputs (see \Cref{app:SEentail}), and furthermore it is computationally expensive. 
Thus, we use exact string matching (EXM), where two sequences get assigned the same cluster if they are exactly equal. 
Once the clusters are determined, we compute the entropy over the cluster distribution, weighting each cluster (1) by its relative frequency among all sequences (\textbf{DSE}$_{EXM}$) or (2) by the cluster probabilities computed as the sum of probabilities of its member sequences (\textbf{SE}$_{EXM}$), both following \citet{farquhar2024detec}. 

Lastly, we also include \textbf{P}$_{true}$ \citep{kadavath2022languagemodelsmostlyknow}, a prompting-based method, see \Cref{app:uq}.

\section{Benchmarking UQ Performance in Function-Calling}
\label{sec:benchmark}

We build upon the Berkeley Function Calling Leaderboard (BFCL) benchmark \citep{bfcl,patil2025bfcl}, which provides a collection of FC examples, mostly in Python with small parts in Java and JavaScript.
The BFCL benchmark consists of multiple tasks, with a separate dataset for evaluating each one of them. Beyond being able to compare different UQ methods for FC purposes, developing a UQ benchmark based on the BFCL allows users to simultaneously evaluate model accuracy---how often the LLM is right---and model UQ---how well the model's uncertainty predicts its correctness. %

To quantify this, we evaluate how well uncertainty estimates discriminate correct from incorrect function calls in the selective prediction setting \citep{el-yaniv2010found}.
First, we determine whether a given response is correct with Abstract Syntax Tree (AST) matching w.r.t.\ the ground truth response, as is the default in BFCL. 
Then, the problem is interpreted as a binary classification of whether the response is correct or not, using the UQ score calculated as the classifier-score. 
The UQ-performance is then quantified as the performance of this classifier. The most appropriate metric to use for this is
the area under the receiver operating characteristic (AUROC) \citep{fawcett2006introduction}. 
It yields a value of 0.5 for random uncertainty estimates and 1.0 for uncertainty estimates that perfectly discriminate correct from incorrect outputs. 
In \Cref{sec:risk_coverage}, we also include risk-coverage (rejection-accuracy) curves 
to illustrate what the AUROC values correspond to in terms of potential accuracy gains thanks to UQ-based abstention.

We evaluate the UQ methods of \Cref{sec:uq-methods} across eight LLMs:
Qwen2.5-\{0.5,3,7\}B-Instruct \citep{qwen2025qwen25}, 
Ministral-8B-Instruct-2410 \citep{mistral2024ministral}, Qwen3-4B-Instruct-2507 \citep{yang2025qwen3}, gemma-2-9b-it \citep{gemma2024gemma2}, and gemma-3-\{4,12\}b-it \citep{gemma2025gemma3}. We evaluate the correctness of the greedy decoding response (temperature $T\!=\!0$) and draw 10 samples at $T\!=\!1$ to compute the multi-sample UQ methods. We begin in \Cref{sec:individual_tasks} by directly adapting the well-defined single-turn tasks from the BFCL to measure the utility of UQ methods.
These tasks vary in difficulty, allowing us to progressively test the LLMs' and UQ methods' performances on increasingly harder tasks.
Next, in \Cref{sec:combinations}, we explain why such a straightforward analysis on each individual task, while useful, is insufficient for comprehensive UQ evaluation, and present our adaptations that allow for a more realistic evaluation.
In \Cref{sec:irrelevance_combinations}, we evaluate how well the UQ methods work for requests that cannot be solved by the LLM given the available tools.

\begin{figure}[t]
\centering
\begin{tikzpicture}[
  font=\small,
  box/.style={draw, rounded corners=2pt, minimum width=3.05cm, minimum height=1.1cm, align=center, thick},
  lbl/.style={font=\small\itshape},
]
\draw[->, thick] (-0.35,-0.35) -- (-0.35,2.85) node[midway, above, rotate=90, yshift=6pt, lbl] {\# functions provided (A)};
\draw[->, thick] (-0.35,-0.35) -- (7.0,-0.35) node[midway, below, yshift=-2pt, lbl] {\# calls required (B)};
\node[box, fill=cyan5]   at (1.4,0.4) {\textbf{Simple}\\[-1pt]{\footnotesize 1 function, 1 call}};
\node[box, fill=blue4]   at (5.0,0.4) {\textbf{Parallel}\\[-1pt]{\footnotesize 1 function, $n$ calls}};
\node[box, fill=green3]  at (1.4,2.1) {\textbf{Multiple}\\[-1pt]{\footnotesize $m$ functions, 1 call}};
\node[box, fill=yellow2] at (5.0,2.1) {\textbf{Parallel-Multiple}\\[-1pt]{\footnotesize $m$ functions, $n$ calls}};
\node[box, fill=pink1] at (10.5,1.25) {\textbf{Irrelevance}\\[-1pt]{\footnotesize 1 insufficient function,}\\[-2pt]{\footnotesize correct action: refuse}};
\draw[dashed, thick] (7.8,-0.5) -- (7.8,3.0);
\node[lbl] at (10.5,-0.35) {unanswerable};
\node[lbl] at (3.2,3.15) {answerable};
\end{tikzpicture}
\caption{The BFCL task regimes we evaluate, arranged along the two axes we vary: the number of functions made available in the request (A) and the number of calls needed to satisfy it (B). \emph{Irrelevance} sits outside this grid, since the provided function is insufficient and the correct behavior is to emit no call at all. \Cref{sec:combinations} combines the four answerable regimes to obtain multi-modal task distributions, and \Cref{sec:irrelevance_combinations} adds \emph{Irrelevance}.}
\label{fig:regimes}
\end{figure}

\setlength{\tabcolsep}{9pt}
\renewcommand{\arraystretch}{1.3}

\begin{table*}[!htbp]
\centering
\resizebox{\textwidth}{!}{\begin{tabular}{llcccccccc}
\toprule
 &  & \multicolumn{3}{c}{\textbf{Single-sample}} & \multicolumn{3}{c}{\textbf{Multi-sample}} & \multicolumn{2}{c}{\textbf{Other}}  \\
\cmidrule(lr){3-5} \cmidrule(lr){6-8} \cmidrule(lr){9-10}
\textbf{Category} & \textbf{Task} & \textbf{MAX} & \textbf{AVG} & \textbf{G-NLL} &  \textbf{SE}$_{EXM}$ & \textbf{DSE}$_{EXM}$ & \textbf{PE} & \textbf{P}$_{true}$ & \textbf{LEN} \\
\hline
\multirow{4}{*}{\textbf{Individual}}
& \textbf{Simple} & $0.74$ \tiny{±$.045$} & \underline{$0.76$} \tiny{±$.046$} & \best{$0.77$} \tiny{±$.045$} & $0.73$ \tiny{±$.049$} & $0.73$ \tiny{±$.049$} & $0.72$ \tiny{±$.045$} & $0.69$ \tiny{±$.052$} & $0.56$ \tiny{±$.052$} \\
& \textbf{Multiple} & $0.71$ \tiny{±$.065$} & \underline{$0.74$} \tiny{±$.066$} & \best{$0.76$} \tiny{±$.065$} & $0.73$ \tiny{±$.073$} & $0.73$ \tiny{±$.073$} & $0.71$ \tiny{±$.071$} & $0.67$ \tiny{±$.075$} & $0.58$ \tiny{±$.081$} \\
& \textbf{Parallel} & $0.66$ \tiny{±$.059$} & \best{$0.72$} \tiny{±$.058$} & \underline{$0.71$} \tiny{±$.058$} & $0.66$ \tiny{±$.059$} & $0.66$ \tiny{±$.059$} & $0.67$ \tiny{±$.061$} & $0.62$ \tiny{±$.063$} & $0.49$ \tiny{±$.064$} \\
& \textbf{Parallel-Multiple} & $0.72$ \tiny{±$.048$} & \underline{$0.74$} \tiny{±$.048$} & \best{$0.76$} \tiny{±$.048$} & $0.73$ \tiny{±$.051$} & $0.73$ \tiny{±$.051$} & $0.72$ \tiny{±$.052$} & $0.67$ \tiny{±$.059$} & $0.55$ \tiny{±$.060$} \\
\hline
\multirow{4}{*}{\textbf{Combinations}}
& \textbf{Simple+Multiple} & $0.73$ \tiny{±$.036$} & \underline{$0.75$} \tiny{±$.037$} & \best{$0.77$} \tiny{±$.037$} & $0.73$ \tiny{±$.041$} & $0.73$ \tiny{±$.041$} & $0.72$ \tiny{±$.038$} & $0.68$ \tiny{±$.042$} & $0.57$ \tiny{±$.044$} \\
& \textbf{Simple+Parallel} & $0.70$ \tiny{±$.037$} & \underline{$0.71$} \tiny{±$.039$} & \best{$0.75$} \tiny{±$.036$} & \underline{$0.71$} \tiny{±$.038$} & $0.70$ \tiny{±$.038$} & $0.67$ \tiny{±$.040$} & $0.68$ \tiny{±$.038$} & $0.59$ \tiny{±$.039$} \\
& \textbf{Multiple+Parallel-Multiple} & $0.73$ \tiny{±$.037$} & $0.71$ \tiny{±$.039$} & \best{$0.77$} \tiny{±$.037$} & \underline{$0.74$} \tiny{±$.041$} & $0.73$ \tiny{±$.041$} & $0.68$ \tiny{±$.041$} & $0.69$ \tiny{±$.045$} & $0.64$ \tiny{±$.045$} \\
        & \textbf{All Combined}  & \underline{$0.72$} \tiny{±$.026$} & $0.71$ \tiny{±$.027$} & \best{$0.76$} \tiny{±$.025$} & \underline{$0.72$} \tiny{±$.027$} & \underline{$0.72$} \tiny{±$.027$} & $0.68$ \tiny{±$.028$} & $0.68$ \tiny{±$.029$} & $0.61$ \tiny{±$.030$} \\
\hline
\multirow{2}{*}{\textbf{Unanswerable}}
& \textbf{Simple+Irrelevance} & $0.78$ \tiny{±$.026$} & \underline{$0.80$} \tiny{±$.025$} & \best{$0.81$} \tiny{±$.025$} & $0.76$ \tiny{±$.026$} & $0.76$ \tiny{±$.026$} & $0.79$ \tiny{±$.026$} & $0.76$ \tiny{±$.026$} & $0.50$ \tiny{±$.032$} \\
& \textbf{All Combined+Irrelevance} & $0.74$ \tiny{±$.020$} & \underline{$0.77$} \tiny{±$.020$} & \best{$0.78$} \tiny{±$.020$} & $0.74$ \tiny{±$.020$} & $0.74$ \tiny{±$.020$} & $0.75$ \tiny{±$.021$} & $0.71$ \tiny{±$.022$} & $0.49$ \tiny{±$.024$} \\
\bottomrule
\end{tabular}}

\caption{
Mean AUROC $\pm$ mean standard error (mean over eight LLMs) for individual tasks, task combinations, and unanswerable requests described in \Cref{sec:individual_tasks}, \Cref{sec:combinations}, and \Cref{sec:irrelevance_combinations} respectively.
The best result in \best{green}, \underline{underlined} second best.
For per-LLM results, see \Cref{fig:individual_tasks,fig:combinations,fig:irrelevance_combinations}
(\Cref{app:detailed_results}).
}
\label{tab:combined_all_tasks}
\end{table*}

\subsection{How Well Do the UQ Methods Work for Individual Tasks?}
\label{sec:individual_tasks}
First, we evaluate how well the UQ methods work when we individually consider the tasks defined in BFCL.
To this end, we use the following individual Python task datasets from BFCL in their standard form: 
\emph{Simple}\footnote{BFCL's \emph{Simple} split consists of 3 parts: Python, Java, and JavaScript. To be able to compare with the other three Python-based tasks (which do not have Java and JavaScript counterparts), we focus on the Python part of \emph{Simple}.},
\emph{Multiple}, \emph{Parallel}, and \emph{Parallel-Multiple} (see \Cref{fig:regimes}).
\emph{Simple} consists of 400 requests where only one function is made available to the LLM. 
It tests whether the model is able to call the function provided, together with the correct argument values, to solve the request.
\emph{Multiple} provides definitions of multiple functions and tests whether the model can choose the correct function as well as its correct arguments to solve the request (200 requests). 
\emph{Parallel} requires the model to call a function more than once in parallel, including the correct arguments (200 requests). 
Like in \emph{Simple}, the model is only provided with the functions it needs to use.
\emph{Parallel-Multiple} combines the challenges of \emph{Parallel} and \emph{Multiple}: 
the model is provided with multiple functions, some of which are unnecessary to solve the request.
The model needs to choose which functions to call, and set their parameters correctly (200 requests).
Importantly, all of the requests in these four tasks are solvable given the functions provided as part of the request.

Throughout our analysis, we exclude the requests for which the greedy decoding results in an output that could not be interpreted by the Python interpreter, i.e., that result in an Abstract Syntax Tree (AST) decoding error.
These cases are clearly incorrect, since they do not produce valid function calls; however, since they do not result in an execution, they also cannot cause irreversible undesired outcomes.
In a practical context, we would thus not need model uncertainty to flag potentially erroneous function calls for these cases.
These errors are rare (on average $3.4\%$ of outputs on the \emph{All Combined} split), and including them barely changes the relative ranking of UQ methods (Spearman correlation $0.94$) and preserves the ordering of G-NLL and the entropy-based methods; see \Cref{app:ast_errors} for effective sample sizes per model and further details.

In \Cref{tab:combined_all_tasks} (top), we report the mean (over eight LLMs) AUROC for each combination of task and UQ method, along with the mean of the per-LLM standard error.
Standard errors were estimated per LLM via bootstrap sampling with 1000 samples; their mean indicates the typical estimation uncertainty of a single LLM's AUROC, rather than a confidence interval on the mean over LLMs.
\Cref{fig:individual_tasks} of \Cref{app:detailed_results} breaks down the results in \Cref{tab:combined_all_tasks} to show the per-model performances.

The results show that the sequence likelihood (G-NLL) %
generally performs best (often by a small margin)
and outperforms the entropy-based multi-sample methods we evaluate.
Overall, the level of performance achieved by the best methods on these individual tasks (AUROC of $\sim$0.7--0.8) resembles the performance achieved by these methods on natural language Q\&A tasks \citep{aichberger2024rethi,farquhar2024detec}.

\subsection{How Well Do UQ Methods Work on More 
Realistic Multi-Modal Task Distributions? }
\label{sec:combinations}
In accuracy-oriented evaluation, it is possible to evaluate a model on individual tasks and then obtain a valid estimator of the mixture's performance by averaging the individual task performance estimates, weighted by the mixture's component weights, because accuracy is a linear metric.

However, unlike accuracy, AUROC measures ranking performance rather than raw correctness counts, and so is not a linear metric.
This implies that if we consider a mixture of multiple different task distributions, we might obtain a lower AUROC performance than for any individual task alone. This is important when using a confidence threshold to decide whether to execute a function call; in general, it would not be practical to set a different threshold for different modes of the task distribution.
For this reason, unlike in the accuracy-oriented benchmarks like BFCL, in order to estimate the UQ performance in a way that informs practical applications, we need to evaluate the UQ methods on more complex, multi-modal data distributions.
To this end, we combine the individual BFCL task data sets described in \Cref{sec:individual_tasks} into four multi-modal combinations, and evaluate the UQ methods on each of these combinations.

The four BFCL tasks we consider offer an opportunity to gradually increase the multi-modality of the input and output distributions along two axes:
A) number of functions at the model's disposal;
B)~number of function calls the model has to produce in order to solve the task.
\textit{Simple+Multiple} varies A, while keeping B fixed at 1.
\textit{Simple+Parallel} varies B between 1 and $n$, while A is always kept at 1, i.e.\ only the function the request actually needs is provided; in \emph{Parallel} this is a single function that has to be called several times.
\textit{Multiple+Parallel-Multiple} is analogous to \textit{Simple+Parallel} in B, but with A set to a larger number such that there are always some unnecessary functions present in the input.
\textit{All Combined} combines both axes of variation, yielding the most heterogeneous of the four combinations we evaluate.%

We present the results in \Cref{tab:combined_all_tasks} (middle) and \Cref{fig:combinations}.
At a high level, the results are similar to the individual task results: G-NLL performs best out of all of the considered methods, and on average, performance levels remain roughly the same as those observed in \Cref{sec:individual_tasks}.

\subsection{Do UQ Methods Allow for Identifying Requests That Are Impossible to Answer?}
\label{sec:irrelevance_combinations}

Beyond the four task datasets introduced in \Cref{sec:individual_tasks}, the BFCL benchmark also includes an \emph{Irrelevance} task to test whether models can recognize and explicitly communicate when they are not able to solve the task given the provided functions. 
Each request in this task includes a single function definition, which is not sufficient to solve the request. 
Thus, a successful answer by the model for this task consists of refusing to output any function call. 
Note that a model is able to get perfect accuracy on this task by always refusing (independent of the request). 
Hence, evaluating on this task alone can yield misleading results about UQ methods. 
We thus combine \emph{Irrelevance} with the other tasks, once with the \emph{Simple} data only (\emph{Simple+Irrelevance}), and once with all other datasets combined (\emph{All Combined+Irrelevance}).
A correct refusal contains no function call whose uncertainty could be scored, so from \emph{Irrelevance} we keep only the requests for which the model emitted a call anyway; all of these are incorrect and enter the evaluation as negatives. Depending on the LLM, this retains between 37 and 215 of the 240 \emph{Irrelevance} requests (\Cref{tab:effective_sample_sizes}). These splits thus test whether a UQ method flags function calls that should not have been made at all. %

We present the results in \Cref{tab:combined_all_tasks} (bottom) and \Cref{fig:irrelevance_combinations}.
Again, we observe good and broadly comparable performance of most UQ methods. 
We also find that the single-sample methods (G-NLL, AVG) slightly outperform the multi-sample methods (SE$_{EXM}$, DSE$_{EXM}$, PE).

\subsection{Are UQ Methods Calibrated for FC?}
\label{sec:calibration}

\begin{wrapfigure}[13]{r}{0.44\linewidth}
    \centering
    \vspace{-23pt}
    \includegraphics[width=\linewidth, trim=10 13 10 26, clip]{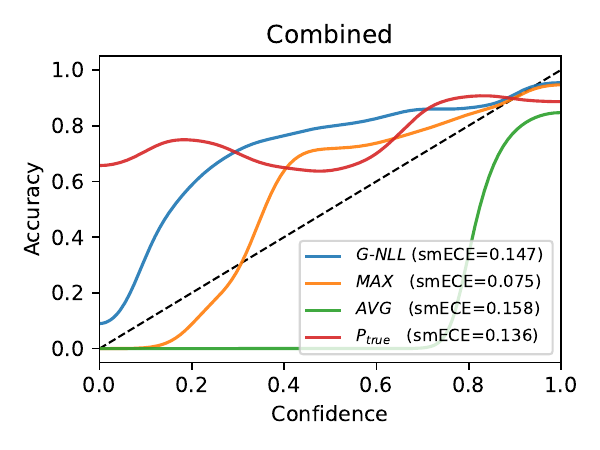}
    \caption{smECE for \textit{All Combined} task, aggregated over all models.}
    \label{fig:calibration}
\end{wrapfigure}
Beyond being able to distinguish incorrect from correct outputs, which we have measured with AUROC in the previous sections, a desirable quality of a UQ score is to be calibrated, 
i.e., to correctly predict the probability of a sample being correct, which is important in decision-making settings \citep{kiyani2025robustdecisionmakingpartially}. 
Among several calibration measures proposed, we choose smooth ECE (smECE) \citep{blasiok2023smooth}, which is a smoothed version of expected calibration error (ECE) that does not exhibit discontinuous behavior. 

Since not all of the UQ methods we evaluate yield a valid probability score, we can only evaluate calibration for the following subset:  {MAX}, {AVG}, {G-NLL}, and {P}$_{true}$. 
In \Cref{fig:calibration}, we show the results for the \textit{All Combined} task as representative for all tasks. 
While quantitative results differ per task, qualitatively, they stay largely the same across tasks, so we relegate the detailed plots to \Cref{app:calibration}.

We make several observations. 
Firstly, {AVG} achieves the worst calibration among all methods, being largely overconfident with confidence values (x-axis) being higher than the accuracy (y-axis).
{P}$_{true}$ and {G-NLL}, on the other hand, are underconfident, as indicated by their confidence being mostly lower than the accuracy. 
For {G-NLL}, this would be expected from natural language, where the multiplication of per-token probabilities, each $<1$, tends to lead to small sequence probabilities. 
However, in function calling, this phenomenon is not as extreme since most probabilities are $\approx 1.0$ (\Cref{sec:discussion_se}). 
Lastly, {MAX} achieves the best calibration. 
It is still slightly underconfident for most parts, but since by definition $\min_i \log p_i \geq \sum_i \log p_i$, it is less underconfident than {G-NLL}, especially in low-confidence regions.
These insights complement the previous AUROC results, and illustrate that ECE does not necessarily correlate with AUROC.\footnote{For another illustrative example, consider the extreme case of a model with accuracy $0.5$ for which a UQ method assigns a correctness probability of $0.5$ for all samples. It has a perfect ECE of $0$ but an AUROC of $0.5$.}
In the setting of selective prediction---generating a function call with an option to abstain---the crucial property to assess is the ability to discriminate between correct and incorrect outputs, which is measured by AUROC. ECE measures whether confidence values are calibrated as probabilities, which is a secondary consideration in this context. Furthermore, none of the multi-sample entropy-based methods output probabilities, and hence their calibration cannot be evaluated.
We note that for methods with high AUROC but poor calibration, post-hoc calibration techniques (e.g., Platt scaling) can be applied to improve calibration without affecting discriminative performance.
We make this precise in \Cref{app:posthoc}: all of the discrimination and selective-prediction metrics we report depend on the scores only through the ranking they induce, so a strictly monotone recalibration of the score leaves them invariant. Post-hoc calibration therefore changes \emph{how} a score should be thresholded, but not \emph{which} score is preferable. \Cref{app:posthoc} also discusses which of the evaluated methods are amenable to it.

\subsection{From AUROC to Abstention}
\label{sec:selective_prediction}

AUROC measures how well a UQ score ranks correct above incorrect function calls, but it does not by itself say how much accuracy a system gains by declining to execute the calls it is least sure about. Since this is the quantity a practitioner cares about, we summarize it here and relegate the full results to \Cref{sec:risk_coverage,app:selection_pred_metrics}.

\begin{figure*}[h!]
    \centering
    \begin{subfigure}[c]{0.48\textwidth}
        \centering
        \includegraphics[width=\textwidth]{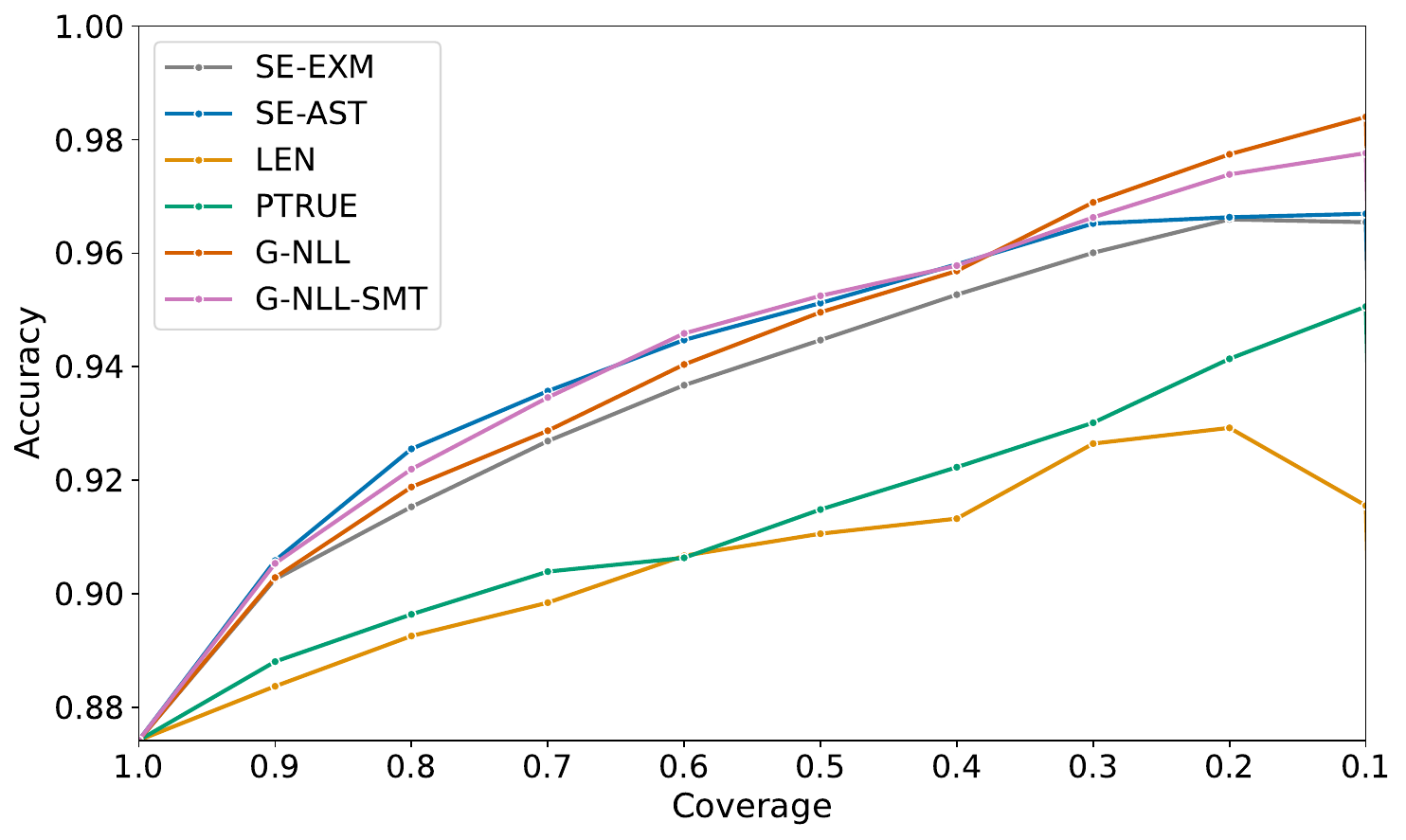}
        \caption{All Combined}
        \label{fig:rc_main_all}
    \end{subfigure}
    \hfill
    \begin{subfigure}[c]{0.48\textwidth}
        \centering
        \includegraphics[width=\textwidth]{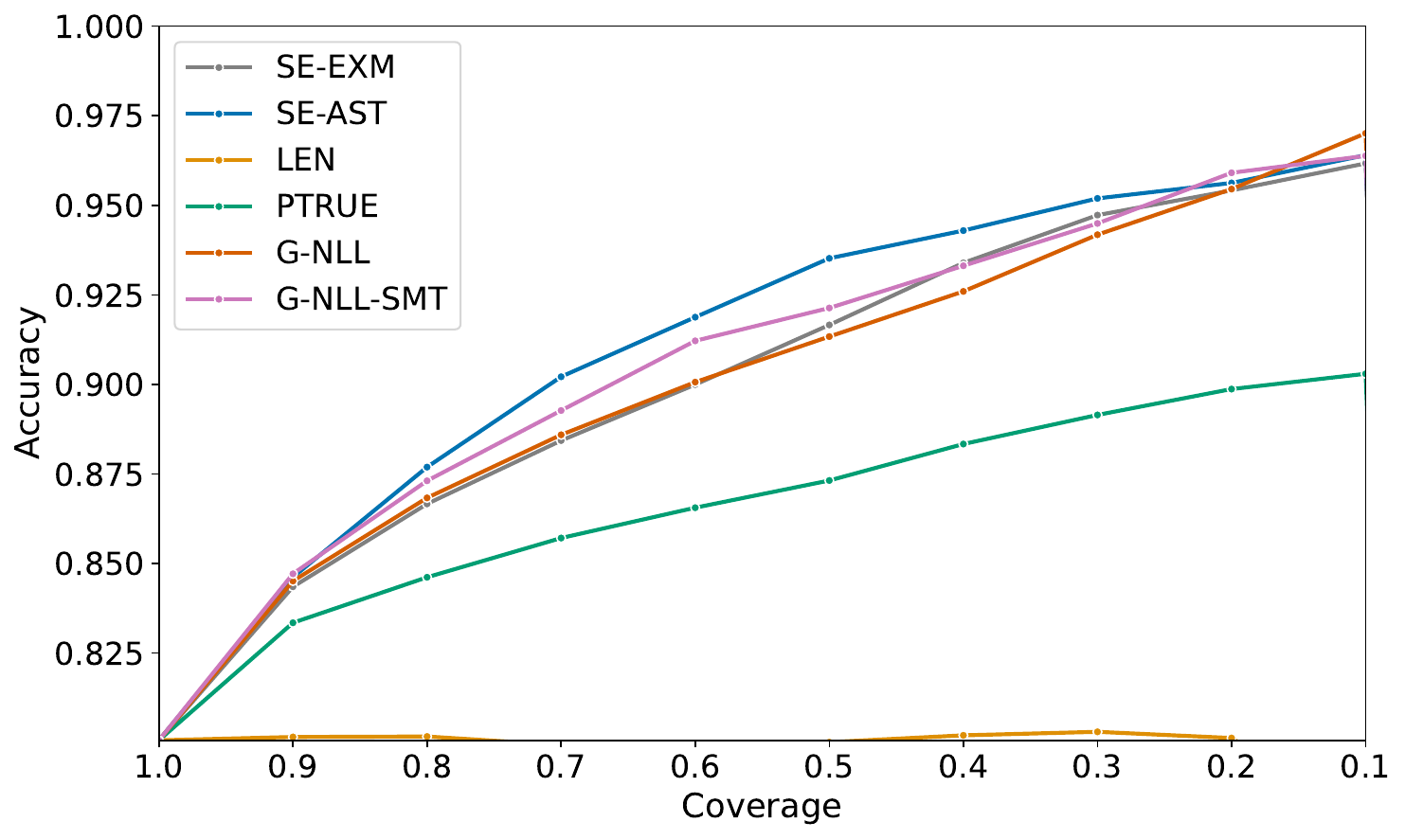}
        \caption{All Combined+Irrelevance}
        \label{fig:rc_main_allirr}
    \end{subfigure}
    \caption{Risk-coverage curves on the two largest splits, averaged over the eight LLMs. Higher is better. The full set of splits is in \Cref{sec:risk_coverage}.}
    \label{fig:riskcoverage_main}
\end{figure*}

\Cref{fig:riskcoverage_main} shows risk-coverage curves for \emph{All Combined} and for \emph{All Combined+Irrelevance}, averaged over models. Across tasks (\Cref{sec:risk_coverage}), abstaining on the $30\%$ of requests with the highest G-NLL$_{SMT}$ uncertainty (an FC-adapted variant of G-NLL introduced in \Cref{sec:new_method}) raises accuracy from ${<}92\%$ to $96\%$ on \emph{Simple} and from $79\%$ to $86\%$ on \emph{Parallel-Multiple} (\Cref{fig:individual_tasks_riskcoverage}), and by roughly $10$ percentage points on the splits that include \emph{Irrelevance} (\Cref{fig:irrelevance_combinations_riskcoverage}). UQ-based abstention is therefore not a marginal effect in this setting. Which operating point to choose depends on the cost of executing an incorrect call relative to abstaining, which is application-specific.

\Cref{tab:selective_main} compares the four headline methods on \emph{All Combined} under AUROC and under two selective-prediction metrics: the normalized area under the risk-coverage curve (NAURC, lower is better) and the prediction-rejection ratio at a $30\%$ and $50\%$ rejection rate (PRR, higher is better); both are defined in \Cref{app:selection_pred_metrics}. Three points stand out. First, the best single-sample method (G-NLL$_{SMT}$) outperforms the best multi-sample method (SE$_{AST}$) under all four metrics, so the advantage of the single-sample family does not depend on the choice of metric. Second, the two FC-adapted variants (SE$_{AST}$ and G-NLL$_{SMT}$, introduced in \Cref{sec:SE_AST,sec:new_method}) are the best or tied-best in their respective families throughout. Third, the metrics are not redundant: at a $30\%$ rejection rate SE$_{AST}$ overtakes plain G-NLL, which the AUROC column does not reveal, and it is the FC-adapted single-sample score G-NLL$_{SMT}$ that closes that gap. We return to this in \Cref{app:selection_pred_metrics}.

\begin{table*}[h!]
\centering
\setlength{\tabcolsep}{10pt}
\footnotesize
\begin{tabular}{lcccc}
\toprule
\textbf{Method} & \textbf{AUROC} $\uparrow$ & \textbf{NAURC} $\downarrow$ & \textbf{PRR@30\%} $\uparrow$ & \textbf{PRR@50\%} $\uparrow$ \\
\midrule
G-NLL & $0.76$ & $0.37$ & $0.40$ & $0.46$ \\
\textbf{G-NLL}$_{SMT}$ & \best{$0.78$} & \best{$0.35$} & \best{$0.44$} & \best{$0.51$} \\
\midrule
SE$_{EXM}$ & $0.72$ & $0.53$ & $0.38$ & $0.41$ \\
\textbf{SE}$_{AST}$ & $0.74$ & $0.52$ & $0.43$ & $0.45$ \\
\bottomrule
\end{tabular}
\caption{G-NLL and SE$_{EXM}$ and their FC-adapted variants G-NLL$_{SMT}$ (\Cref{sec:new_method}) and SE$_{AST}$ (\Cref{sec:SE_AST}) on \emph{All Combined}, under AUROC and the selective-prediction metrics of \Cref{app:selection_pred_metrics} (mean over eight LLMs). Arrows indicate the direction of improvement; best result in \best{green}. Per-split results for all methods are in \Cref{tab:combined_all_tasks} and \Cref{tab:naurc,tab:prr30,tab:prr50}.}
\label{tab:selective_main}
\end{table*}

\section{Adapting UQ Methods to FC}
\label{sec:discussion}

The results from \Cref{sec:benchmark} suggest that the more involved entropy-based multi-sample UQ methods we evaluate underperform compared to simple single-sample methods. 
This contrasts with observations in natural language generation tasks. 
In this section, we investigate whether the performance of existing UQ methods in the FC setting can be improved by adapting them to the particular structure of FC outputs.

\subsection{Adapting Multi-Sample UQ Methods} 
\label{sec:SE_AST}

In the following, we investigate whether multi-sample UQ methods can be tailored to the FC setting. Specifically, we focus on the two variants of Semantic Entropy (SE and DSE) as the highest-performing multi-sample methods.

The original SE implementation clusters generations with an NLI-based entailment model. We report in \Cref{app:SEentail} that this does not transfer to FC outputs: entailment-based clustering is substantially worse than exact string matching, which we attribute to textual entailment being the wrong equivalence relation for structured outputs. This motivates the structural alternative we consider here.

We hypothesize that the clustering based on exact string matching (EXM) used in our experiments might be overly strict, as it does not take permutation invariance in the function arguments into account.
To circumvent this, we utilize the Abstract Syntax Trees (\textbf{AST}) to cluster FC outputs based on \citet{bfcl}. Two outputs get assigned the same cluster if their AST matches. This approach effectively treats the arguments to the function as a dictionary that is permutation-invariant.

\paragraph{Results}
\Cref{tab:new_method_se} suggests that AST clustering improves performance over simple EXM clustering in eight out of ten splits. However, these gains remain insufficient to establish an advantage over single-sample methods.
We note that multi-sample methods also require proportionally higher throughput (proportional to the number of samples $J$), which is non-negligible especially in on-device scenarios.
We further ablate the choice of temperature $T$ and number of samples $J$ in \Cref{app:sensitivity}: increasing $T$ to $1.5$ or $J$ to $30$ lets SE$_{AST}$ match or slightly exceed G-NLL on the answerable splits and roughly match G-NLL$_{SMT}$ at $T=1.5$ (but not at $J=30$); it remains clearly behind both on the \emph{Irrelevance} splits, at $J$ times the inference cost.
This raises the question of why multi-sample UQ methods do not perform better in comparison, despite incorporating multiple samples. To better understand this limitation, we conduct a more detailed analysis in the following.

\begin{table*}[!htbp]
\centering
\setlength{\tabcolsep}{10pt}
\footnotesize
\begin{tabular}{ll|cc|cc}
\toprule
\textbf{Category} & \textbf{Task} & \textbf{SE}$_{EXM}$ & \textbf{SE}$_{AST}$ & \textbf{DSE}$_{EXM}$ & \textbf{DSE}$_{AST}$ \\ 
\hline
\multirow{4}{*}{\textbf{Individual}}
& \textbf{Simple} & $0.73$ \tiny{±$.049$} & \best{$0.75$} \tiny{±$.047$} & $0.73$ \tiny{±$.049$} & \best{$0.75$} \tiny{±$.047$} \\
& \textbf{Multiple} & $0.73$ \tiny{±$.073$} & \best{$0.75$} \tiny{±$.067$} & $0.73$ \tiny{±$.073$} & \best{$0.74$} \tiny{±$.067$} \\
& \textbf{Parallel} & $0.66$ \tiny{±$.059$} & \best{$0.69$} \tiny{±$.056$} & $0.66$ \tiny{±$.059$} & \best{$0.69$} \tiny{±$.056$} \\
& \textbf{Parallel-Multiple} & $0.73$ \tiny{±$.051$} & \best{$0.74$} \tiny{±$.049$} & \best{$0.73$} \tiny{±$.051$} & \best{$0.73$} \tiny{±$.049$} \\
\hline
\multirow{4}{*}{\textbf{Combinations}}
& \textbf{Simple+Multiple} & $0.73$ \tiny{±$.041$} & \best{$0.75$} \tiny{±$.038$} & $0.73$ \tiny{±$.041$} & \best{$0.75$} \tiny{±$.038$} \\
& \textbf{Simple+Parallel} & $0.71$ \tiny{±$.038$} & \best{$0.73$} \tiny{±$.036$} & $0.70$ \tiny{±$.038$} & \best{$0.73$} \tiny{±$.036$} \\
& \textbf{Multiple+Parallel-Multiple} & $0.74$ \tiny{±$.041$} & \best{$0.76$} \tiny{±$.038$} & $0.73$ \tiny{±$.041$} & \best{$0.75$} \tiny{±$.038$} \\
& \textbf{All Combined} & $0.72$ \tiny{±$.027$} & \best{$0.74$} \tiny{±$.026$} & $0.72$ \tiny{±$.027$} & \best{$0.74$} \tiny{±$.026$} \\
\hline
\multirow{2}{*}{\textbf{Unanswerable}}
& \textbf{Simple+Irrelevance} & \best{$0.76$} \tiny{±$.026$} & $0.75$ \tiny{±$.025$} & \best{$0.76$} \tiny{±$.026$} & $0.74$ \tiny{±$.025$} \\
& \textbf{All Combined+Irrelevance} & \best{$0.74$} \tiny{±$.020$} & \best{$0.74$} \tiny{±$.020$} & \best{$0.74$} \tiny{±$.020$} & $0.73$ \tiny{±$.020$} \\
\bottomrule
\end{tabular}

\caption{
Mean AUROC, $\pm$ mean standard error (mean taken over the eight LLMs) for all splits from \Cref{sec:benchmark}. Within each EXM/AST pair, the better result is in \best{green} (both if tied).
}
\label{tab:new_method_se}
\end{table*}

\subsection{Patterns of Token Probabilities in FC vs.\ Natural Language Tasks}
\label{sec:discussion_se}

To better understand why SE and DSE still do not outperform single-sample methods like G-NLL, we investigate an example of a wrong response by Qwen2.5-7B-Instruct to a request from the \emph{Parallel-Multiple} task in \Cref{fig:example_SE_failure1} 
(full description of question and available functions in \Cref{app:parallel-multiple_example}). 
Here, the LLM makes the mistake of adding an unexpected parameter at the end of the last function call (\texttt{year=1882}). 
This example illustrates two patterns that we find to hold in general:
\begin{enumerate}[leftmargin=*, topsep=0pt, itemsep=2pt, partopsep=0pt, parsep=0pt]
\item The LLM assigns high probabilities (probability $\approx 1$) to most tokens and exhibits only a minimal degree of uncertainty at a small number of tokens ($<5$), including those associated with the error.
\item Even for tokens the LLM is slightly uncertain about (probability $<1$), the second most likely token is either semantically equivalent (\texttt{=["} vs. \texttt{=['}) or is significantly less likely (\texttt{year} vs. \texttt{specific}).
\end{enumerate}

As a result, the samples for computing the multi-sample methods are all semantically identical to the greedy-decoded one, differing only in syntactically equivalent tokens \texttt{=["} vs. \texttt{=['}.
Consequently, only a single semantic cluster is formed, resulting in low SE, even though the response is incorrect.

We compare this to a wrong response the same LLM gives to a natural language question from the Natural Questions (NQ) dataset \citep{kwiatkowski2019natural} in \Cref{app:nq}. 
In this setting, neither of the above two patterns holds.
In the FC setting, where available functions are provided in-context and the valid output space is restricted, the probabilities of individual tokens are higher than in natural language settings. 
Indeed, across the eight LLMs, we find that the average token probability on \emph{All Combined} is $0.97$,\footnote{This also explains the large overconfidence of AVG in \Cref{sec:calibration}, as the average accuracy on \textit{All Combined} is $<0.9$.} while for the NQ dataset it is only $0.71$. 
Similarly, the mean number of semantic clusters formed is substantially lower on BFCL than on NQ ($3.4$ vs. $6.4$).%

In contrast, single-sample methods like G-NLL rely directly on the sequence probability, making them sensitive to variations in the few tokens where the LLM assigns probabilities lower than 1.
However, \Cref{fig:example_SE_failure1} highlights a limitation of single-sample methods in their current form: The impact of lower probabilities for semantically unimportant tokens (the token \texttt{=["} is required by the function description) might distort their score.
Our knowledge about the structure of function calls and which tokens carry semantic weight raises the question of whether single-sample methods can be adapted to better leverage these characteristics of FC outputs.

\subsection{Adapting Single-Sample UQ Methods}
\label{sec:new_method}

In the following, we implement a method that aims to address the limitation of single-sample methods. 
We first analyze which tokens are semantically meaningful in the FC context (as opposed to syntactically required ones) and derive a general algorithm to extract those from an FC output. 

\begin{wrapfigure}[41]{r}{0.49\linewidth}
\vspace{0.5pt}
\centering
\includegraphics[width=0.6\linewidth]{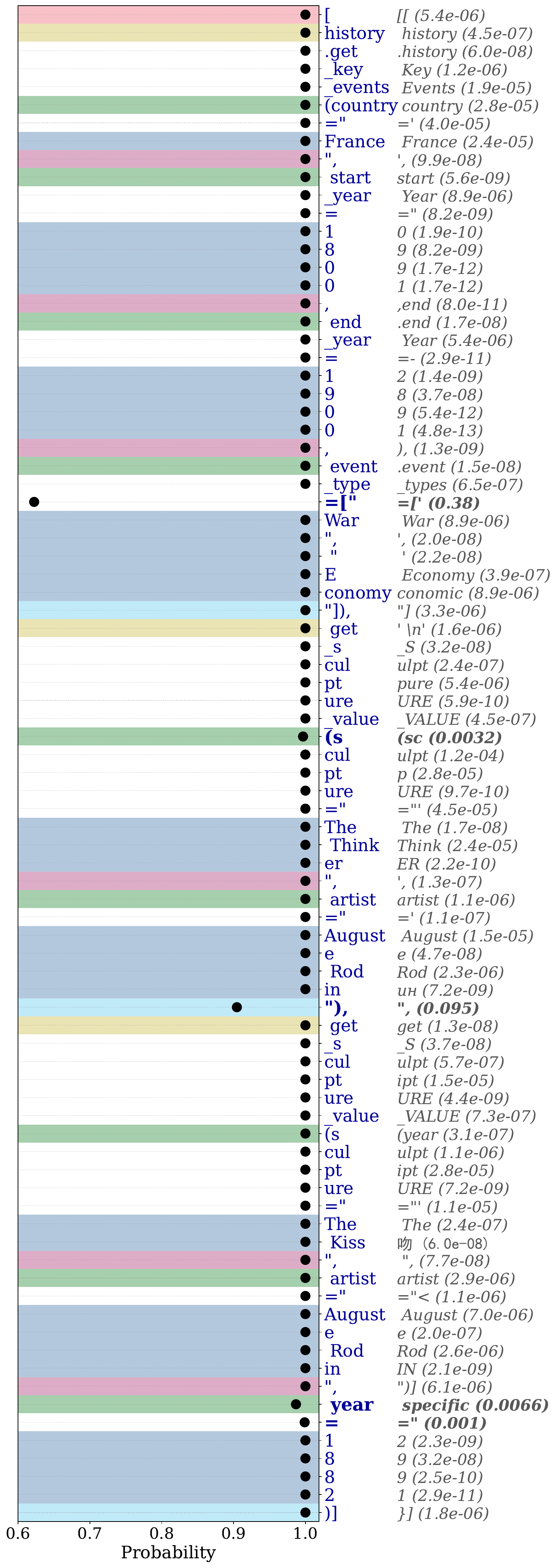}
\caption{\raggedright{
Top token probabilities for an incorrect Qwen2.5-7B-Instruct response (lowest 5 in bold) in the \emph{Parallel-Multiple} task (details in \Cref{app:parallel-multiple_example}). Second-highest tokens shown in gray. Color coding explained in \Cref{sec:new_method}. The ground-truth answer does not contain the argument
\texttt{year=1882}.
}}
\label{fig:example_SE_failure1}
\end{wrapfigure}

\paragraph{Semantically Meaningful Tokens}
We define Semantically Meaningful Tokens in the FC context as those corresponding to positions where the LLM has to decide between tokens that likely lead to either correct or incorrect outputs. Using the example response in \Cref{fig:example_SE_failure1}, these are:
\begin{enumerate}[leftmargin=*, topsep=0pt, itemsep=2pt, partopsep=0pt, parsep=0pt]
    \item \textcolor{pink1}{\rule{1.2ex}{1.2ex}} Tokens deciding if functions are called or the task is refused (the first token: \texttt{[})
    \item \textcolor{yellow2}{\rule{1.2ex}{1.2ex}} Tokens deciding if the right function is called (\texttt{history}, \texttt{get}, \texttt{get})
    \item \textcolor{green3}{\rule{1.2ex}{1.2ex}} Tokens deciding if the right parameter is used (\texttt{(country}, \texttt{start}, \texttt{end}, \texttt{event}, \texttt{(s}, \texttt{artist}, \texttt{(s}, \texttt{artist}, \texttt{year})
    \item \textcolor{blue4}{\rule{1.2ex}{1.2ex}} Tokens deciding if the right parameter value is chosen (\texttt{France}, \texttt{1}, \texttt{8}, \texttt{0}, \texttt{0}, \texttt{1}, \texttt{9}, \texttt{0}, \texttt{0}, \texttt{War}, \texttt{",}, \texttt{"}, \texttt{E}, \texttt{conomy}, \texttt{The}, \texttt{Think}, \texttt{er}, \texttt{August}, \texttt{e}, \texttt{Rod}, \texttt{in}, \texttt{The}, \texttt{Kiss}, \texttt{August}, \texttt{e}, \texttt{Rod}, \texttt{in}, \texttt{1}, \texttt{8}, \texttt{8}, \texttt{2})
    \item \textcolor{cyan5}{\rule{1.2ex}{1.2ex}} Tokens deciding if the right number of functions is called (\texttt{"]),}, \texttt{"),}, \texttt{)]})
    \item \textcolor{purple6}{\rule{1.2ex}{1.2ex}} Tokens deciding if the right number of parameters are included in a call (\texttt{",}, \texttt{,}, \texttt{,}, \texttt{",}, \texttt{",}, \texttt{",})
\end{enumerate}

All remaining tokens are primarily syntactically required to form a correct function call output (such as \texttt{=}), or continuations of function names according to the available names in the context (such as \texttt{cul}, \texttt{pt}, \texttt{ure}). 
Under this definition, in the above example, the semantically meaningful tokens would include the token \texttt{year}, which has higher uncertainty and is related to the LLM's mistake. In contrast, the token \texttt{=["} would not be included, which should not have high uncertainty, since the function format requires the parameter value in the form of an array, making this token syntactically necessary.
While this logic can be argued about and other ways of defining semantically meaningful tokens could be considered, we find that even this simple heuristic already yields improvements in UQ performance. Lastly, we highlight that this approach is related to previous work that proposes to focus on relevant tokens when computing UQ scores \citep{duan2023shifting, fadeeva2024fact, bakman2024mars}. However, these methods are tailored to the context of natural language and no such approach has been proposed for the FC setting so far.

\textbf{Implementing SMT Extraction}\quad
Once defined, there are many options to extract the semantically meaningful tokens from a given LLM function-call output. We opt for a simple, rule-based approach that is both computationally efficient and easy to implement. Nonetheless, it requires a separate implementation for different FC output formats, depending on the LLM's specific instruction used for finetuning. For the models evaluated, gemma and Qwen models share the same FC output format corresponding to a list of Python calls (\Cref{fig:example_SE_failure1}), while Ministral outputs its function calls in a JSON format and thus requires a separate implementation of the token extraction function. We include the algorithms for both in \Cref{app:new_method}.
Note that this extraction operates purely on the syntactic structure of the model's own output tokens and never accesses the gold function call, so it is label-free and cannot leak ground-truth information into the UQ score.

We note that the extraction logic follows from the formal grammar of the output schema/format used for AST parsing, and only a handful of FC output schemas account for the majority of LLM FC use. To further demonstrate the ease of adapting the SMT extraction to new formats, we prompted GPT-5.2 with 10 examples each of Qwen2.5-7B-Instruct and Ministral function-call outputs together with our Python-format SMT algorithm, asking it to produce the corresponding algorithm for the Ministral JSON format. The resulting function selected on average $57\%$ of Ministral-generated tokens as semantically meaningful, compared to $54\%$ by our manually written algorithm, selecting nearly identical tokens. The performance impact was minimal: on average $0.001$ AUROC difference over the 10 splits, with a maximum divergence of $0.004$ (on \emph{Simple}).

Once the semantically meaningful tokens are defined and extracted, we aggregate the corresponding log-probabilities in the same way as for the whole token sequence (MAX, G-NLL, AVG). This leads to an adaptation of single-sample methods to the FC setting we refer to as \textbf{SMT}.

\begin{table*}[t]
\centering
\setlength{\tabcolsep}{10pt}
\footnotesize
\begin{tabular}{ll|cc}
\toprule
\textbf{Category} & \textbf{Task} & \textbf{G-NLL} & \method  \\
\hline
\multirow{4}{*}{\textbf{Individual}}
& \textbf{Simple} & {$0.77$} \tiny{±$.045$} & \best{$0.78$} \tiny{±$.047$} \\
& \textbf{Multiple} & \best{$0.76$} \tiny{±$.065$} & \best{$0.76$} \tiny{±$.065$} \\
& \textbf{Parallel} & {$0.71$} \tiny{±$.058$} & \best{$0.73$} \tiny{±$.057$} \\
& \textbf{Parallel-Multiple} & {$0.76$} \tiny{±$.048$} & \best{$0.78$} \tiny{±$.044$} \\
\hline
\multirow{4}{*}{\textbf{Combinations}}
& \textbf{Simple+Multiple} & {$0.77$} \tiny{±$.037$} & \best{$0.78$} \tiny{±$.038$} \\
& \textbf{Simple+Parallel} & {$0.75$} \tiny{±$.036$} & \best{$0.76$} \tiny{±$.036$} \\
& \textbf{Multiple+Parallel-Multiple} & {$0.77$} \tiny{±$.037$} & \best{$0.79$} \tiny{±$.035$} \\
& \textbf{All Combined} & {$0.76$} \tiny{±$.025$} & \best{$0.78$} \tiny{±$.025$} \\
\hline
\multirow{2}{*}{\textbf{Unanswerable}}
& \textbf{Simple+Irrelevance} & {$0.81$} \tiny{±$.025$} & \best{$0.82$} \tiny{±$.025$} \\
& \textbf{All Combined+Irrelevance} & {$0.78$} \tiny{±$.020$} & \best{$0.80$} \tiny{±$.019$} \\

\bottomrule
\end{tabular}
\caption{
Mean AUROC, $\pm$ mean standard error (mean taken over the eight LLMs) for all splits from \Cref{sec:benchmark}. Within each row, the better result is in \best{green} (both if tied).%
}
\label{tab:new_method}
\end{table*}

\vspace{-5pt}
\paragraph{Results}

\Cref{tab:new_method} shows the performance of G-NLL$_{SMT}$ compared to G-NLL, which has emerged as the dominant method in \Crefrange{sec:individual_tasks}{sec:irrelevance_combinations}.
On three out of the four individual tasks, G-NLL$_{SMT}$ performs better than G-NLL. 
It also performs better on all of the combinations of tasks, including those that involve unanswerable questions. While the observed gains are only small, they are consistent in the across-model mean.
We caution, however, that these AUROC gains are not statistically significant: a paired Wilcoxon signed-rank test across models (\Cref{app:paired_significance}) finds no systematic advantage of G-NLL$_{SMT}$ over G-NLL, with the per-split median $\Delta$AUROC small in magnitude and split in both directions.
In \Cref{app:detailed_results} we show that the same is true for MAX$_{SMT}$ and AVG$_{SMT}$.
Furthermore, SMT also improves calibration of G-NLL on 9 of the 10 splits and of MAX on 6 of the 10; for MAX, smECE increases on \textit{Parallel}, \textit{Parallel-Multiple}, and both splits with irrelevance. For example, on \textit{All Combined}, the smECE drops from $0.147$ for G-NLL to $0.091$ for G-NLL$_{SMT}$ and from $0.075$ for MAX to $0.053$ for MAX$_{SMT}$ (the impact on AVG is insignificant). The full calibration results can be found in \Cref{app:calibration}.

The differences are somewhat more consistent under the selective-prediction metrics, which measure the utility of a score under abstention rather than its ranking quality alone. In \Cref{app:selection_pred_metrics}, G-NLL$_{SMT}$ attains the best NAURC on 9 of the 10 splits (counting its ties with G-NLL on \emph{Simple} and \emph{Simple+Parallel}) and the best prediction-rejection ratio on all 10 splits at PRR$@50\%$, and it is at least as good as G-NLL on every split at both rejection rates. The margins are small, however, and we do not test them for significance, so we read this as a consistent trend in direction rather than a demonstrated improvement. A possible explanation is that AUROC summarizes the ranking over the whole score range, whereas PRR$@k$ depends only on the $k\%$ most uncertain outputs, i.e., on the high-uncertainty region in which abstention decisions are made, so a modification that mainly affects the ordering among the most uncertain outputs would be more visible under the latter. Overall we regard SMT as a cheap modification that does not hurt and tends to help slightly---most clearly in calibration---rather than one that reliably improves UQ performance in this setting.

Overall, we thus find that the particularities of FC outputs can be leveraged when designing UQ methods for this setting, though to different degrees. AST-based clustering yields a consistent improvement over exact-match clustering for the multi-sample methods, whereas restricting single-sample scores to semantically meaningful tokens leaves discrimination essentially unchanged, improves the calibration of G-NLL, and yields small, untested gains in the selective-prediction metrics.

\section{Conclusion}
We present the first benchmark for UQ methods in LLM Function-Calling (FC) settings. %
We show that in FC settings, simple logit-based UQ methods largely perform better than the more sophisticated entropy-based multi-sample methods we evaluate. %
We further show that the particularities of the FC setting can be utilized to improve existing UQ methods. Specifically, we adapt the semantic-entailment method of multi-sample UQ methods by using Abstract Syntax Trees. For single-sample UQ methods, we use only semantically meaningful tokens, which we define by analyzing FC outputs, when calculating their UQ scores. AST-based clustering consistently improves over exact-match clustering, while the semantically-meaningful-token variant leaves discrimination essentially unchanged and instead improves the calibration of G-NLL, with smaller gains in the selective-prediction metrics. %

\section{Limitations}

We see three main limitations to this work. First, it would be interesting to compare how our analysis holds for the frontier models that top the BFCL benchmark. However, these tend to not disclose token probabilities through their API, and for those which are open-source, they are computationally infeasible to run on our hardware. For such models, scores that rely only on sampled outputs, such as DSE and self-consistency (SC), remain applicable; we report them in \Cref{app:black_box}. Also, we believe that our findings for smaller and medium-sized models are still very relevant, especially in the context of on-device function calling, which would need to rely on such smaller models.
Second, our benchmark is single-turn and derived from BFCL, with all candidate functions provided in the context. It would be interesting to extend this analysis to more challenging function-calling settings, such as multi-turn or agent settings. Since the mechanism behind our findings, i.e., the LLM assigning high probability to most output tokens, with the remaining uncertainty concentrated in a few semantically meaningful tokens, follows from the constrained FC output format rather than from BFCL specifically, we expect the advantage of G-NLL and the motivation for AST-based clustering to carry over. However, multi-turn settings additionally introduce errors that compound across turns and require aggregating uncertainty over trajectories, which we leave for future research.

Third, the multi-sample baselines we evaluate are entropy-based: predictive entropy and the SE/DSE family. Other families of sampling-based estimators that we do not evaluate here include similarity-graph methods such as DegMat, Eccentricity and EigenValue \citep{lin2024gener}, hybrid scores that combine sample similarity with sequence likelihood such as CoCoA \citep{vashurin2025uncer}, and representation-based estimators computed from hidden states such as EigenScore \citep{chen2024insid}. Since the output-space signal is compressed in the FC setting (\Cref{sec:discussion_se}), these estimators may behave differently from the entropy-based ones, and evaluating them---which can reuse the same sample budget---is an important direction for future work. Our conclusions should therefore be read as applying to the entropy-based multi-sample methods we evaluate rather than to sampling-based UQ in general.

\newpage
\bibliography{references}
\bibliographystyle{tmlr}

\newpage
\appendix
\crefalias{section}{appendix}
\crefalias{subsection}{appendix}

\section{Examples of BFCL Tasks}
\label{app:bfcl_examples}

In this section, we provide illustrative examples of different tasks from the BFCL dataset.

\subsection{Simple} \mbox{}

\begin{lstlisting}[style=noindent, language=Python, escapeinside={(*@}{@*)}, numbers=none]
{
 "id": "simple_0",
 "question": [
  [
   {
    "role": "user",
    "content": "Find the area of a triangle with a base of 10 units and height of 5 units."
   }
  ]
 ],
 "function": [
  {
   "name": "calculate_triangle_area",
   "description": "Calculate the area of a triangle given its base and height.",
   "parameters": {
    "type": "dict",
    "properties": {
     "base": {
      "type": "integer",
      "description": "The base of the triangle."
     },
     "height": {
      "type": "integer",
      "description": "The height of the triangle."
     },
     "unit": {
      "type": "string",
      "description": "The unit of measure (defaults to 'units' if not specified)"
     }
    },
    "required": [
     "base",
     "height"
    ]
   }
  }
 ]
}
\end{lstlisting}

\subsection{Multiple} \mbox{}

\begin{lstlisting}[style=noindent, language=Python, escapeinside={(*@}{@*)}, numbers=none]
{
 "id": "multiple_0",
 "question": [
  [
   {
    "role": "user",
    "content": "Can I find the dimensions and properties of a triangle, if I know its three sides are 5 units, 4 units and 3 units long?"
   }
  ]
 ],
 "function": [
  {
   "name": "triangle_properties.get",
   "description": "Retrieve the dimensions, such as area and perimeter, of a triangle if lengths of three sides are given.",
   "parameters": {
    "type": "dict",
    "properties": {
     "side1": {
      "type": "integer",
      "description": "The length of first side of the triangle."
     },
     "side2": {
      "type": "integer",
      "description": "The length of second side of the triangle."
     },
     "side3": {
      "type": "integer",
      "description": "The length of third side of the triangle."
     },
     "get_area": {
      "type": "boolean",
      "description": "A flag to determine whether to calculate the area of triangle. Default is true.",
      "default": true,
      "optional": true
     },
     "get_perimeter": {
      "type": "boolean",
      "description": "A flag to determine whether to calculate the perimeter of triangle. Default is true.",
      "default": true,
      "optional": true
     },
     "get_angles": {
      "type": "boolean",
      "description": "A flag to determine whether to calculate the internal angles of triangle. Default is true.",
      "default": true,
      "optional": true
     }
    },
    "required": [
     "side1",
     "side2",
     "side3"
    ]
   }
  },
  {
   "name": "circle_properties.get",
   "description": "Retrieve the dimensions, such as area and circumference, of a circle if radius is given.",
   "parameters": {
    "type": "dict",
    "properties": {
     "radius": {
      "type": "float",
      "description": "The length of radius of the circle."
     },
     "get_area": {
      "type": "boolean",
      "description": "A flag to determine whether to calculate the area of circle. Default is true.",
      "default": true,
      "optional": true
     },
     "get_circumference": {
      "type": "boolean",
      "description": "A flag to determine whether to calculate the circumference of circle. Default is true.",
      "default": true,
      "optional": true
     }
    },
    "required": [
     "radius"
    ]
   }
  }
 ]
}
\end{lstlisting}

\subsection{Parallel} \mbox{}

\begin{lstlisting}[style=noindent, language=Python, escapeinside={(*@}{@*)}, numbers=none]
{
 "id": "parallel_0",
 "question": [
  [
   {
    "role": "user",
    "content": "Play songs from the artists Taylor Swift and Maroon 5, with a play time of 20 minutes and 15 minutes respectively, on Spotify."
   }
  ]
 ],
 "function": [
  {
   "name": "spotify.play",
   "description": "Play specific tracks from a given artist for a specific time duration.",
   "parameters": {
    "type": "dict",
    "properties": {
     "artist": {
      "type": "string",
      "description": "The artist whose songs you want to play."
     },
     "duration": {
      "type": "integer",
      "description": "The duration for which the songs should be played, in minutes."
     }
    },
    "required": [
     "artist",
     "duration"
    ]
   }
  }
 ]
}
\end{lstlisting}

\subsection{Parallel-Multiple} \mbox{}
\label{app:parallel-multiple_example}

\begin{lstlisting}[style=noindent, language=Python, escapeinside={(*@}{@*)}, numbers=none]
{
 "id": "parallel_multiple_177",
 "question": [
  [
   {
    "role": "user",
    "content": "Can you first find out the key historical events related to 'War' and 'Economy' that took place in France between the years 1800 and 1900? After that, could you please tell me the current market value of the sculpture 'The Thinker' created by the artist 'Auguste Rodin'? Lastly, I would also like to know the market value of the sculpture 'The Kiss', also created by 'Auguste Rodin', in the year 1882."
   }
  ]
 ],
 "function": [
  {
   "name": "get_sculpture_value",
   "description": "Retrieve the current market value of a particular sculpture by a specific artist.",
   "parameters": {
    "type": "dict",
    "properties": {
     "sculpture": {
      "type": "string",
      "description": "The name of the sculpture."
     },
     "artist": {
      "type": "string",
      "description": "The name of the artist who created the sculpture."
     }
    },
    "required": [
     "sculpture",
     "artist"
    ]
   }
  },
  {
   "name": "history.get_key_events",
   "description": "Retrieve key historical events within a specific period for a certain country.",
   "parameters": {
    "type": "dict",
    "properties": {
     "country": {
      "type": "string",
      "description": "The name of the country for which history is queried."
     },
     "start_year": {
      "type": "integer",
      "description": "Start year of the period for which history is queried."
     },
     "end_year": {
      "type": "integer",
      "description": "End year of the period for which history is queried."
     },
     "event_type": {
      "type": "array",
      "items": {
        "type": "string",
        "enum": ["War", "Revolutions", "Diplomacy", "Economy"]
        },
      "description": "Types of event. If none is provided, default that all types will be considered."
     }
    },
    "required": [
     "country",
     "start_year", 
     "end_year"
    ]
   }
  }
 ]
}
\end{lstlisting}

\needspace{8\baselineskip}
\subsection{Irrelevance} \mbox{}

\begin{lstlisting}[style=noindent, language=Python, escapeinside={(*@}{@*)}, numbers=none]
{
 "id": "irrelevance_0",
 "question": [
  [
   {
    "role": "user",
    "content": "Calculate the area of a triangle given the base is 10 meters and height is 5 meters."
   }
  ]
 ],
 "function": [
  {
   "name": "determine_body_mass_index",
   "description": "Calculate body mass index given weight and height.",
   "parameters": {
    "type": "dict",
    "properties": {
     "weight": {
      "type": "float",
      "description": "Weight of the individual in kilograms."
     },
     "height": {
      "type": "float",
      "description": "Height of the individual in meters."
     }
    },
    "required": [
     "weight",
     "height"
    ]
   }
  }
 ] 
}
\end{lstlisting}

\newpage
\section{Example of Natural Questions Answer}
\label{app:nq}

\begin{figure}[h!]
  \centering

    \includegraphics[width=0.4\linewidth]{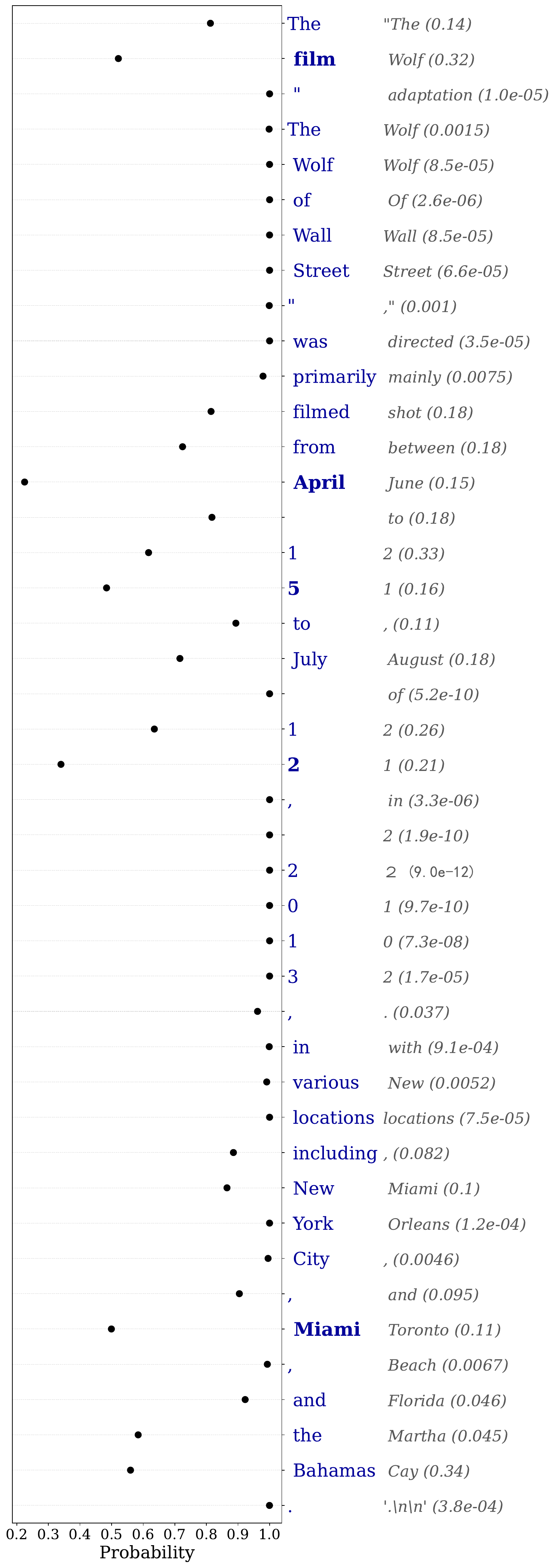}
    \caption{
    Qwen2.5-7B-Instruct response top token probabilities (lowest 5 bolded) for a question wrongly answered in the Natural Questions dataset. \\Question:  \scriptsize \emph{When was the Wolf of Wall Street filmed?} \\Ground truth answer: \scriptsize \emph{ The Wolf of Wall Street (2013 film) Filming began on August 8, 2012, in New York. Jonah Hill announced that his first day of shooting was September 4, 2012. Filming also took place in Closter, New Jersey and Harrison, New York. In January 2013, additional scenes were shot at a set built in an abandoned office building in Ardsley, New York. Scenes at the beach house were filmed in Sands Point, New York.}}
   \label{fig:example_SE_failure2}

\end{figure}

\newpage
\section{Uncertainty Estimators} \label{app:uq}

In response to an input question $x$, LLMs output sequences of tokens, so let $y = (y_1, \dotsc, y_L)$ denote an output sequence of length $L$. The first family of uncertainty estimation methods uses the greedily decoded sequence, where the LLM always chooses the most likely next token. We denote this sequence with $y^G$. Each token $y_i^G$ comes with a negative log-likelihood value (NLL, i.e., token uncertainty) that the LLM has assigned to it during generation based on the previous tokens $y_{<i} = (y_1, \dots, y_{i-1})$, which we denote as $- \log p(y^G_i|y^G_{<i}, x)$. Single-sample uncertainty methods differ in how they aggregate these probabilities across the sequence into a single uncertainty $u(y^G)$ value for the whole sequence.

\textbf{MAX} outputs the highest NLL value in the sequence \begin{equation}
u_{\text{MAX}}(y^G) := \max_i -\log p(y^G_i|y^G_{<i}, x).\end{equation}

\textbf{G-NLL} sums up the NLLs of the sequence (name as per \citet{aichberger2024rethi}): 
\begin{equation}
\label{eq:G-NLL}
    u_{\text{G-NLL}}(y^G) := \sum_{i=1}^L -\log p(y^G_i|y^G_{<i}, x).
\end{equation}

\textbf{AVG} averages the NLLs. It thus normalizes G-NLL by length and can also be interpreted as the logarithm of the sequence perplexity \begin{equation}u_{\text{AVG}}(y^G) := -1/L \sum_{i=1}^L \log p(y^G_i|y^G_{<i}, x).\end{equation}

\textbf{LEN} is a baseline that simply outputs the number of tokens in the answer, treating longer sequences as inherently more uncertain. This is an obviously flawed uncertainty estimator and is used more as a sanity-checking baseline that should perform close to random: 
\begin{equation}u_{\text{LEN}}(y^G) := L.\end{equation}

The second family of uncertainty estimators samples $J=10$ responses $(y^1, \dots, y^J)$ from the LLM, i.e., based on the next-token probabilities with temperature 1. It then combines them into one uncertainty estimate $u((y^1, \dots, y^J))$ based on different ways of clustering the responses $y^j$.

\textbf{PE} does not perform any clustering and treats all outputs as different, even if their decoded strings and even their chosen tokens are the same. This gives a naïve estimate of the predictive entropy of the sequence distribution.
\begin{align}
&u_{\text{PE}}((y^1, \dots, y^J))
:= H[p(y|x)]
\\
&= -E_{y \sim p(y|x)}[\log p(y|x)] \\
&\approx -1/J \sum_{j=1}^{J} 1/L_j\sum_{i=1}^{L_j} \log p(y^{j}_i|y^{j}_{<i}, x)
\end{align}
As for AVG, each sampled sequence is normalized by its own length $L_j$, so PE is estimated on a per-token scale; like all the estimators in this section, larger values indicate higher uncertainty.

\textbf{SE}$_{EXM}$: Semantic Entropy \citep{farquhar2024detec} methods first cluster the output sequences $(y^1, \dots, y^J)$ into $K$ clusters $c^1, \dots, c^K$ and then compute an entropy over the cluster distribution $p(c | x)$. SE$_{EXM}$ uses exact string matching to determine whether two outputs belong to the same cluster, i.e., their decoded text must be exactly equal. 
\begin{align}
\label{eq:se}
&u_{\text{SE}}((c^1, \dots, c^K))
:= H[p(c | x)]
\\
&= -E_{c \sim p(c|x)}[\log p(c | x)] \\
&\approx - \sum_{k=1}^K p(c^k | x) \log p(c^k | x) ,
\end{align}
with 
\begin{equation}  \label{eq:se_clusterprob}
p(c | x) = \sum_{j=1}^J \mathds{1} \! \{ y^j \in c \} p(y^j | x) \ .
\end{equation}
The cluster probabilities are then normalized to 1, following~\citet{kuhn2023semantic}.

\textbf{DSE}$_{EXM}$: Discrete Semantic Entropy uses the same clustering algorithm but does not consider the individual sequence likelihoods to compute the cluster distribution $p(c | x)$:
\begin{equation}  \label{eq:dse}
p(c | x) = \frac{1}{J} \sum_{j=1}^J \mathds{1} \! \{ y^j \in c \} \ .
\end{equation}
\textbf{SE}$_{AST}$ uses a clustering algorithm that is more suited to the FC setup. In particular, it checks whether two decoded function calls are equal up to permutations in their (named) arguments. This is called abstract syntax tree (AST) matching, because it treats two function calls as equivalent if their call will indeed lead to the same inputs to the same function. We first use the AST parser implemented in BFCL to extract the AST of each function call, cast as a dictionary. We then group the calls into clusters based on their key-value pairs' equivalence.  

\textbf{DSE}$_{AST}$ uses abstract syntax tree matching to form clusters.

\textbf{P}$_{true}$ \citep{kadavath2022languagemodelsmostlyknow} looks at the probability with which an LLM predicts that $y^G$ is true, given few-shot examples and the set of all generated responses $\{y^1, \dots, y^J\}$ serving as ``brainstormed'' alternatives:
\begin{equation}
    u_{\text{P}_{true}}(y^G) := p(y = \text{A} \mid x_{\text{P}_{true}}) \ ,
\end{equation}
where $x_{\text{P}_{true}}$ consists of the following parts:
\begin{enumerate}[leftmargin=*]
    \item System prompt:
    \begin{lstlisting}[style=noindent, language=Python, escapeinside={(*@}{@*)}, numbers=none, lineskip=-1pt]
    <|im_start|>system
    You are an expert in composing functions. You are given a question and a set of possible functions. You are also given brainstormed ideas and a possible answer. Based on the question, you have to assess if the possible answer achieves the purpose.
    
    If none of the functions can be used, it should be stated out in the answer. If the given question lacks the parameters required by the function, it should also be pointed out in the answer. Otherwise, only function calls should be included in the answer.
    
    Any invoked function(s) MUST be put it in the format of [func_name1(params_name1=params_value1, params_name2=params_value2...), func_name2(params)]
    <|im_end|>
    \end{lstlisting}
    \item Incorrect few-shot example: 
    \begin{lstlisting}[style=noindent, language=Python, escapeinside={(*@}{@*)}, numbers=none, lineskip=-1pt]
    <|im_start|>user
    Question: (*@\itcomment{[few-shot question inserted here]}@*)

    Here is a list of functions in JSON format that can be invoked:
    (*@\itcomment{[few-shot function inserted here]}@*)
    
    Here are some brainstormed ideas:
    (*@\itcomment{[few-shot brainstormed ideas inserted here]}@*)
    
    Possible answer:
    (*@\itcomment{[few-shot incorrect answer inserted here]}@*)
    
    Is the possible answer:
    A) True
    B) False
    Respond with A or B only.<|im_end|>
    <|im_start|>assistant
    The possible answer is: B<|im_end|>
     \end{lstlisting}

    \pagebreak
    \item Correct few-shot example: 
    \begin{lstlisting}[style=noindent, language=Python, escapeinside={(*@}{@*)}, numbers=none, lineskip=-1pt]
    <|im_start|>user
    Question: (*@\itcomment{[few-shot question inserted here]}@*)
    
    Here is a list of functions in JSON format that can be invoked:
    (*@\itcomment{[few-shot function inserted here]}@*)
    
    Here are some brainstormed ideas:
    (*@\itcomment{[few-shot brainstormed ideas inserted here]}@*)
    
    Possible answer:
    (*@\itcomment{[few-shot correct answer inserted here]}@*)
    
    Is the possible answer:
    A) True
    B) False
    Respond with A or B only.<|im_end|>
    <|im_start|>assistant
    The possible answer is: A<|im_end|>
    \end{lstlisting}

    \item Actual example for which the correctness of $y^G$ is evaluated: 
    \begin{lstlisting}[style=noindent, language=Python, escapeinside={(*@}{@*)}, numbers=none, lineskip=-1pt]
    <|im_start|>user
    Question: (*@\itcomment{[question inserted here]}@*)
    
    Here is a list of functions in JSON format that can be invoked:
    (*@\itcomment{[function inserted here]}@*)
    
    Here are some brainstormed ideas:
    (*@\itcomment{[$\{y^1, \dots, y^J\}$ inserted here]}@*)
    
    Possible answer:
    (*@\itcomment{[$y^{G}$ inserted here]}@*)
    
    Is the possible answer:
    A) True
    B) False
    Respond with A or B only.<|im_end|>
    <|im_start|>assistant
    The possible answer is: 
    \end{lstlisting}
    
\end{enumerate}

\noindent
Unlike the section's other estimators, $u_{\text{P}_{true}}$ is a confidence: a higher P$_{true}$ score indicates lower uncertainty.

\noindent 
Throughout the P$_{true}$ experiments, we use the following few-shot example: 
\begin{enumerate}[leftmargin=*]
    \item Few-shot question:
    \begin{lstlisting}[style=noindent, language=Python, escapeinside={(*@}{@*)}, numbers=none]
    What is 19/53?
    \end{lstlisting}
    \item Few-shot function:
    \begin{lstlisting}[style=noindent, language=Python, escapeinside={(*@}{@*)}, numbers=none]
    [{'name': 'divide', 'description': 'Divides two numbers.', 'parameters': {'type': 'dict', 'properties': {'numerator': {'type': 'float', 'description': 'The numerator of the fraction.'}, 'denominator': {'type': 'float', 'description': 'The denominator of the fraction.'}}, 'required': ['numerator', 'denominator']}}, {'name': 'add', 'description': 'Adds two integers.', 'parameters': {'type': 'dict', 'properties': {'a': {'type': 'int', 'description': 'The first integer.'}, 'b': {'type': 'int', 'description': 'The second integer.'}}, 'required': ['a', 'b']}}]
    \end{lstlisting}
    \item Few-shot brainstormed ideas:
    \begin{lstlisting}[style=noindent, language=Python, escapeinside={(*@}{@*)}, numbers=none]
    [divide(denominator=53, numerator=19)]
    [divide(numerator=53, denominator=53)]
    [divide(numerator=19, denominator=19)]
    [divide(numerator=19, denominator=53)]
    \end{lstlisting}
     \item Few-shot incorrect answer:
    \begin{lstlisting}[style=noindent, language=Python, escapeinside={(*@}{@*)}, numbers=none]
    [divide(numerator=53, denominator=19)]
    \end{lstlisting}
    \item Few-shot correct answer:
    \begin{lstlisting}[style=noindent, language=Python, escapeinside={(*@}{@*)}, numbers=none]
    [divide(numerator=19, denominator=53)]
    \end{lstlisting}
\end{enumerate}

\section{Further Experimental Results}

\subsection{Entailment-Based Semantic Entropy}
\label{app:SEentail}

The original SE implementation clusters generations using an NLI-based entailment model. We report explicitly that this does not transfer to FC outputs. On \emph{All Combined} with Qwen2.5-7B-Instruct, DeBERTa-based \citep{he2021deberta} entailment clustering reaches an AUROC of $0.60$ for both SE$_{ENT}$ and DSE$_{ENT}$, against $0.75$ for the exact-match variants (\Cref{tab:entailment})---a substantial degradation, and at a considerably higher computational cost, since the pairwise entailment calls add hours of processing. We attribute this to textual entailment being the wrong equivalence relation for structured outputs: function-call equivalence is structural, determined by whether two calls invoke the same functions with equivalent arguments, rather than by whether their textual renderings entail one another. This is a diagnostic experiment on a single model and split, reported as a negative result rather than as an exhaustive evaluation of entailment-based clustering, and it is what led us to use exact string matching in \Cref{sec:benchmark} and to consider the structural alternative of \Cref{sec:SE_AST}.

\begin{table}[h!]
\centering
\footnotesize
\begin{tabular}{l|c}
\toprule
\textbf{Method} & \textbf{All Combined}       \\ %
 \hline
\textbf{SE}$_{ENT}$ & $0.60$ \\
\textbf{DSE}$_{ENT}$ & $0.60$ \\
\textbf{SE}$_{EXM}$ & \best{$0.75$} \\
\textbf{DSE}$_{EXM}$ & \best{$0.75$}  \\
\bottomrule
\end{tabular}
\caption{AUROC of entailment-based versus exact string matching-based SE and DSE on the \emph{All Combined} task for Qwen2.5-7B-Instruct. Entailment-based clustering performs substantially worse on FC outputs.}
\label{tab:entailment}
\end{table}

\subsection{Per-Model AUROC Results}
\label{app:detailed_results}

\begin{figure*}[h!]
    \centering
    \begin{subfigure}[c]{0.48\textwidth}
        \centering
        \includegraphics[width=\textwidth]{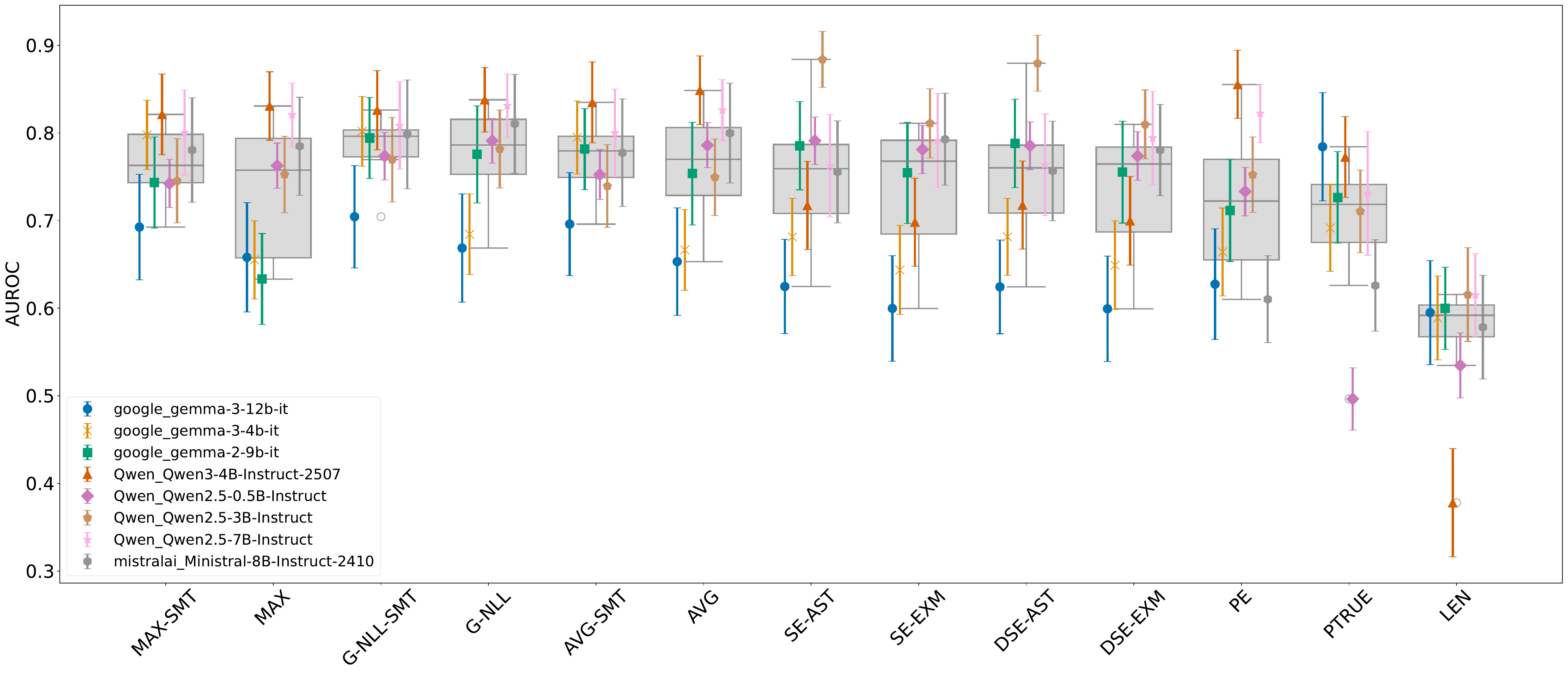}
        \caption{Simple}
        \label{fig:Simple}
    \end{subfigure}
    \hfill
    \begin{subfigure}[c]{0.48\textwidth}
        \centering
        \includegraphics[width=\textwidth]{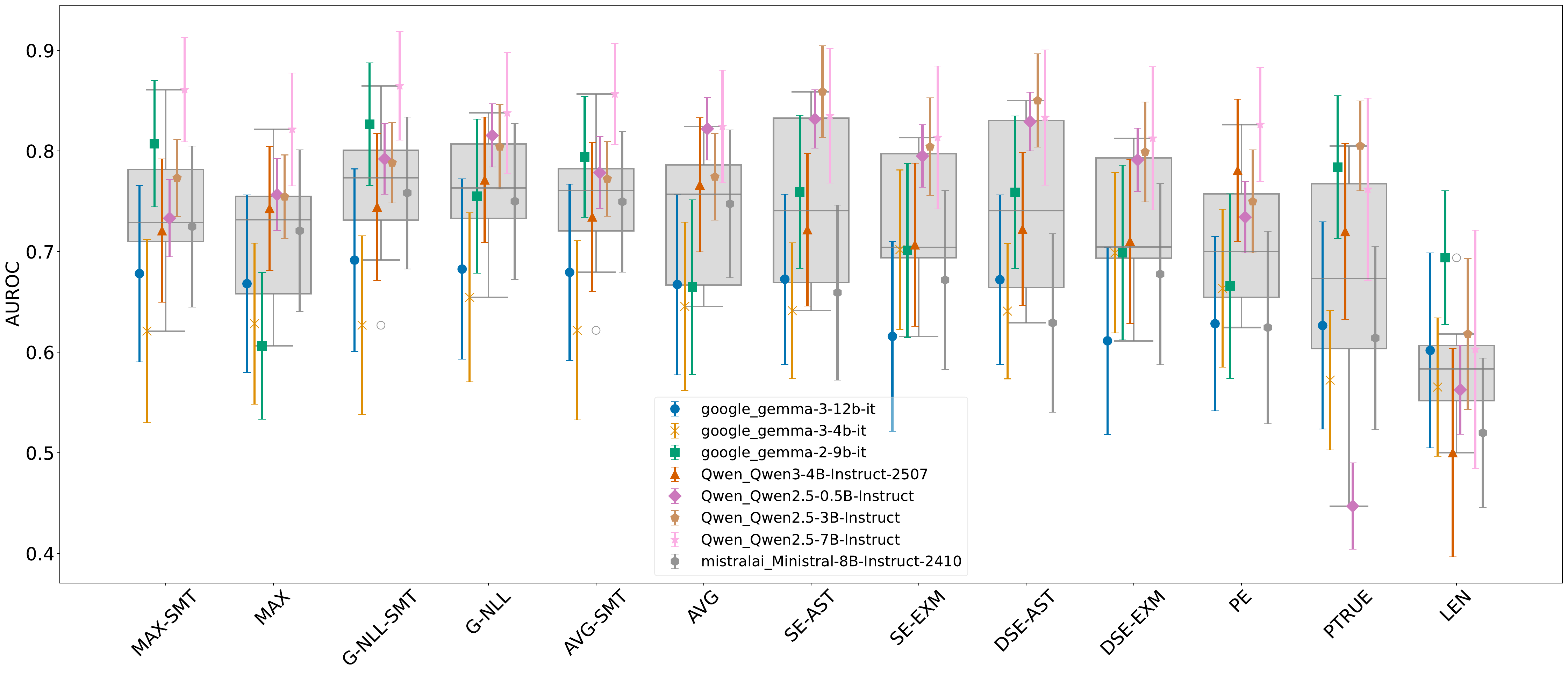}
        \caption{Multiple}
        \label{fig:Multiple}
    \end{subfigure}

    \vskip\baselineskip

    \begin{subfigure}[c]{0.48\textwidth}
        \centering
        \includegraphics[width=\textwidth]{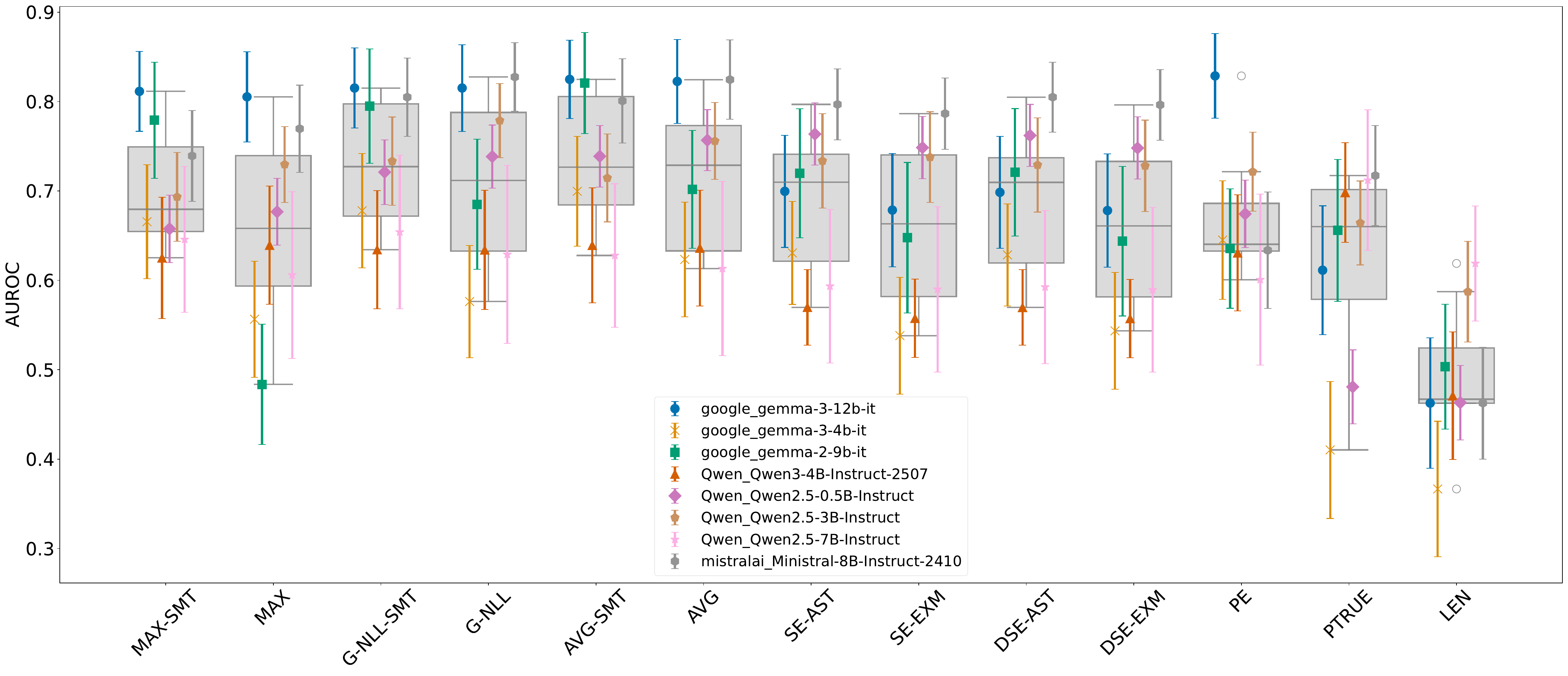}
        \caption{Parallel}
        \label{fig:Parallel}
    \end{subfigure}
    \hfill
    \begin{subfigure}[c]{0.48\textwidth}
        \centering
        \includegraphics[width=\textwidth]{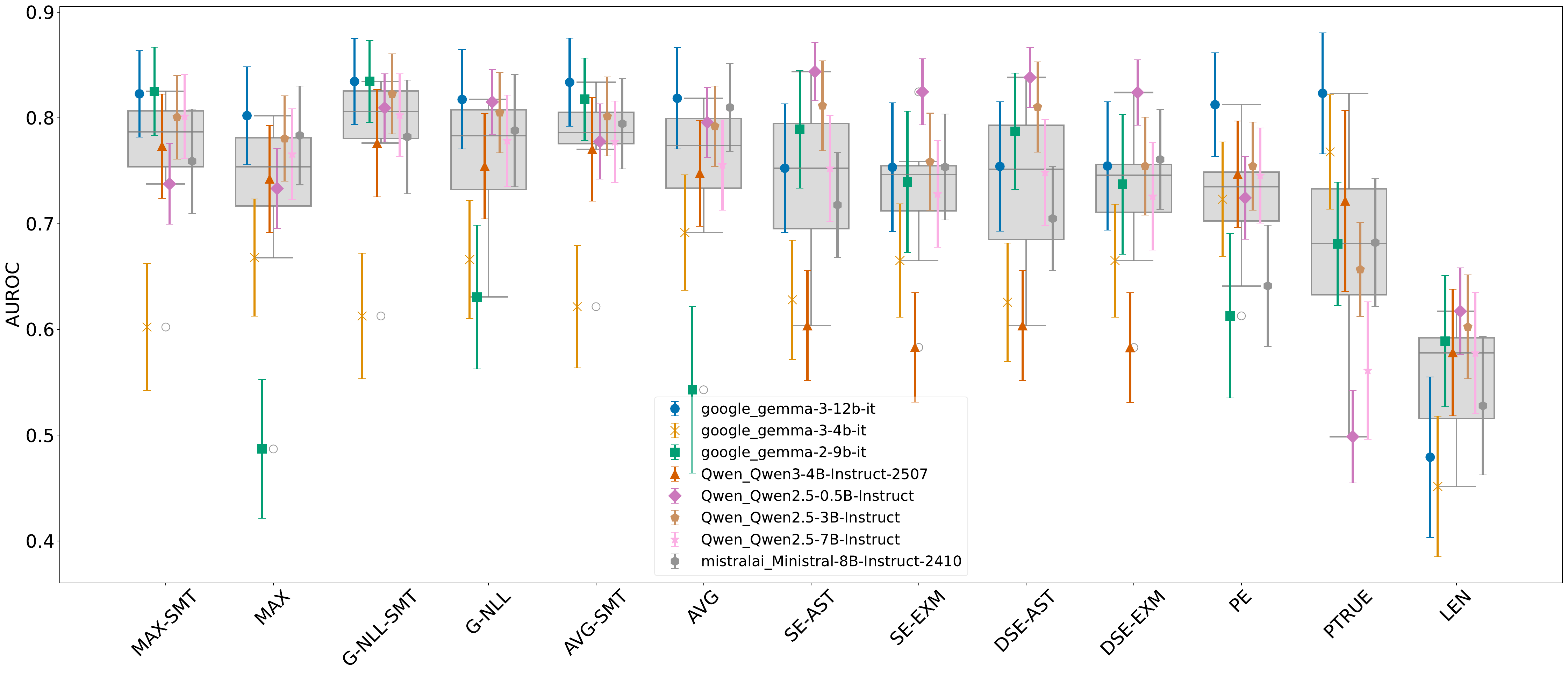}
        \caption{Parallel-Multiple}
        \label{fig:Parallel-Multiple}
    \end{subfigure}

    \caption{
        Individual tasks:
        AUROC $\pm$ standard error (bootstrap estimate with $n=1000$) for the individual tasks described in \Cref{sec:individual_tasks}. Boxplots summarize the variation across LLMs.
    }
    \label{fig:individual_tasks}
\end{figure*}

\begin{figure*}[h!]
    \centering
    \begin{subfigure}[c]{0.48\textwidth}
        \centering
        \includegraphics[width=\textwidth]{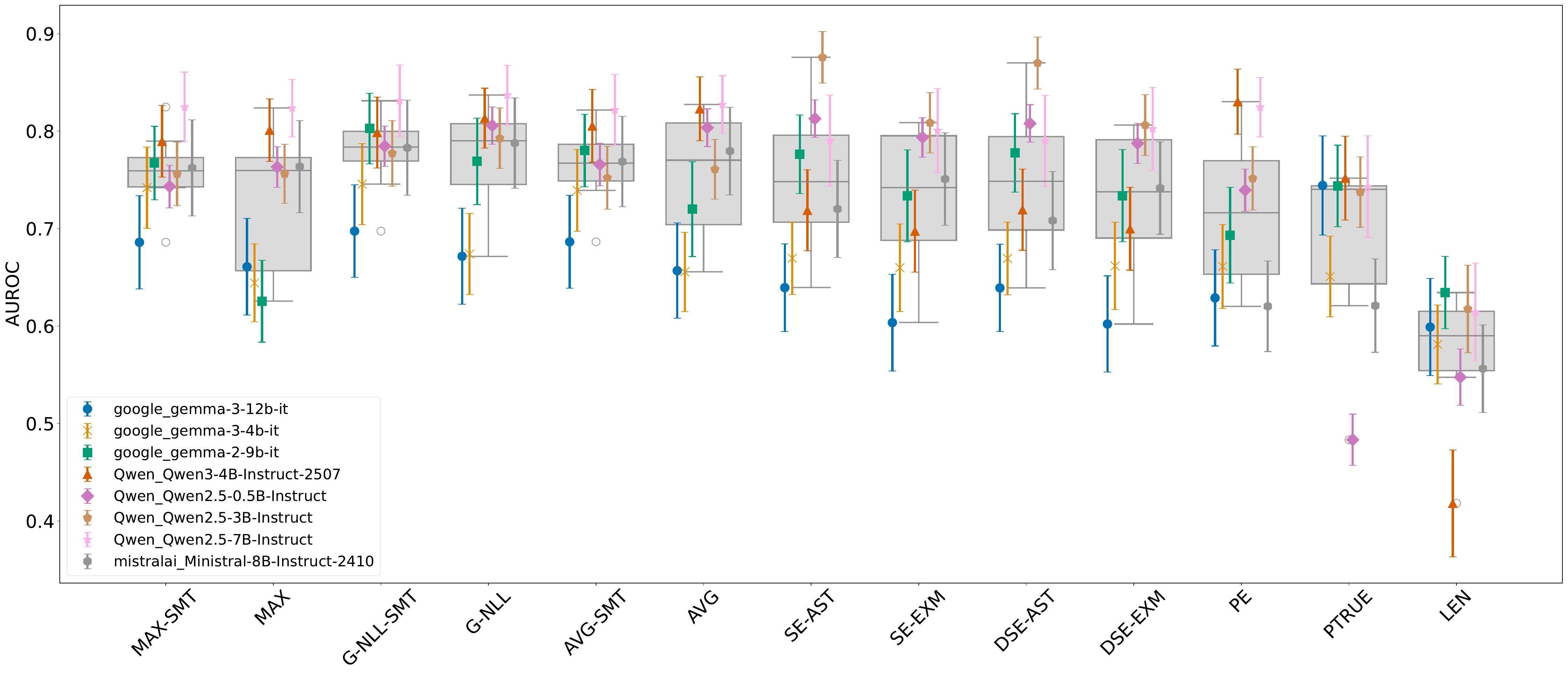}
        \caption{Simple+Multiple}
        \label{fig:Simple+Multiple}
    \end{subfigure}
    \hfill
    \begin{subfigure}[c]{0.48\textwidth}
        \centering
        \includegraphics[width=\textwidth]{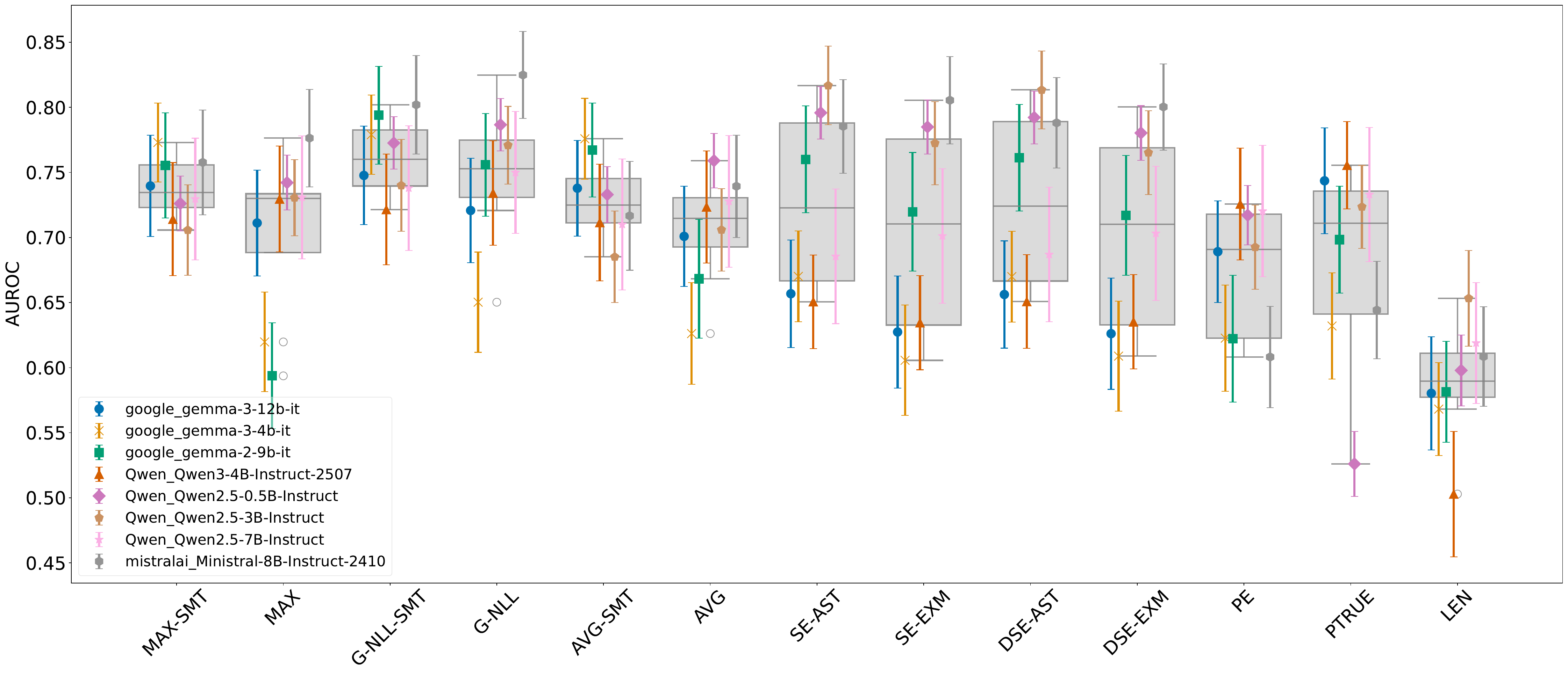}
        \caption{Simple+Parallel}
        \label{fig:Simple+Parallel}
    \end{subfigure}

    \vspace{0.5em}

    \begin{subfigure}[c]{0.48\textwidth}
        \centering
        \includegraphics[width=\textwidth]{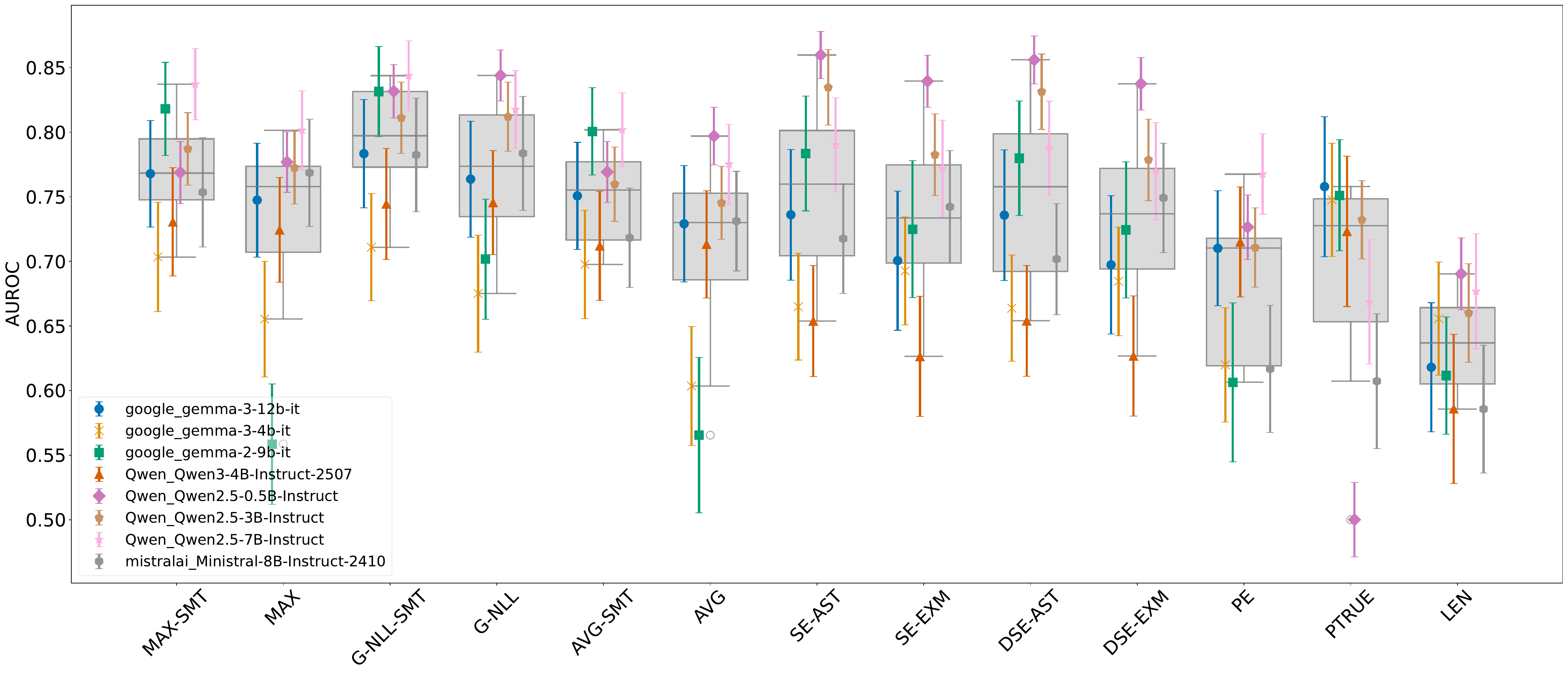}
        \caption{Multiple+\mbox{Parallel-Multiple}}
        \label{fig:Multiple+Parallel-Multiple}
    \end{subfigure}
    \hfill
    \begin{subfigure}[c]{0.48\textwidth}
        \centering
        \includegraphics[width=\textwidth]{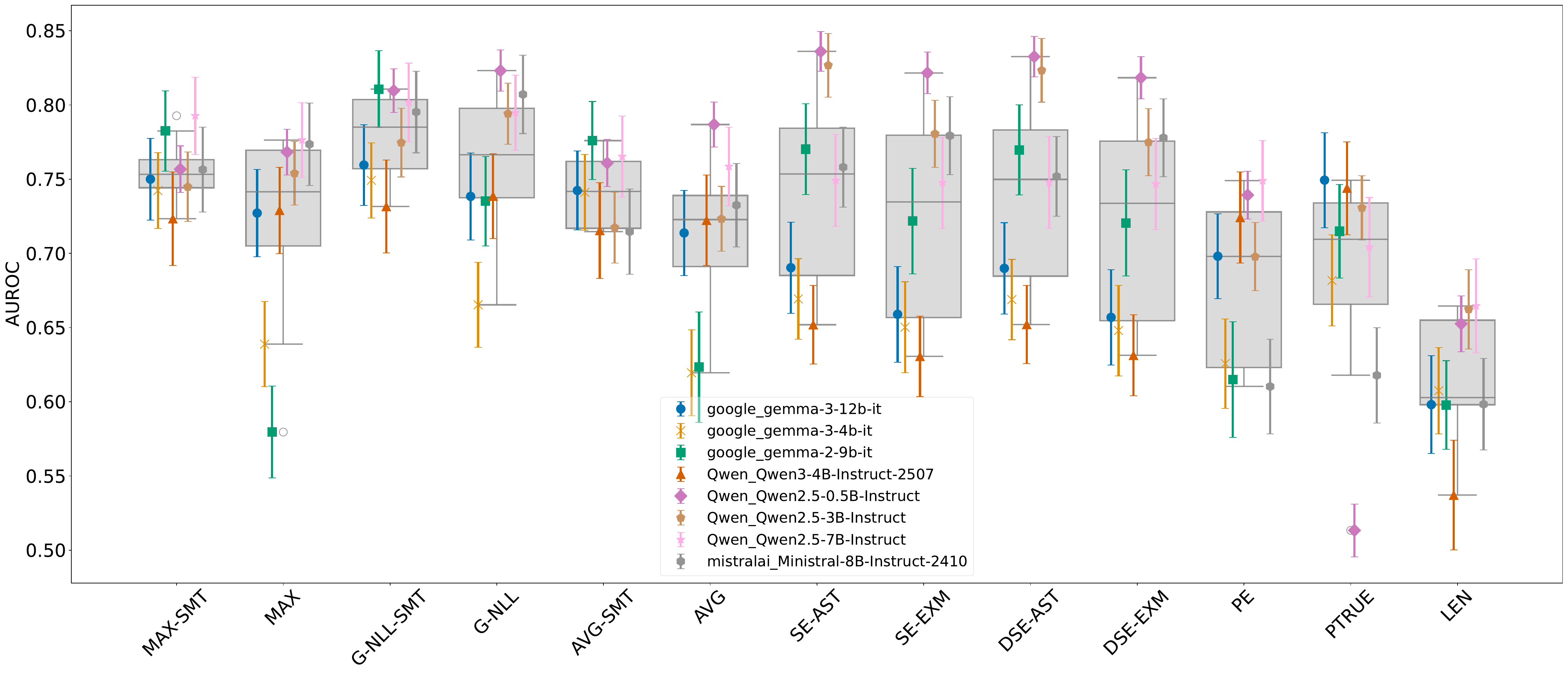}
        \caption{All Combined}
        \label{fig:All}
    \end{subfigure}

    \caption{
        Combinations of tasks: 
        AUROC $\pm$ standard error for the combinations of tasks described in \Cref{sec:combinations}. Boxplots summarize the variation across LLMs for a given UQ method.
    }
    \label{fig:combinations}
\end{figure*}

\begin{figure*}[h!]
\centering
    \begin{subfigure}[c]{0.48\textwidth}
        \centering
        \includegraphics[width=\textwidth]{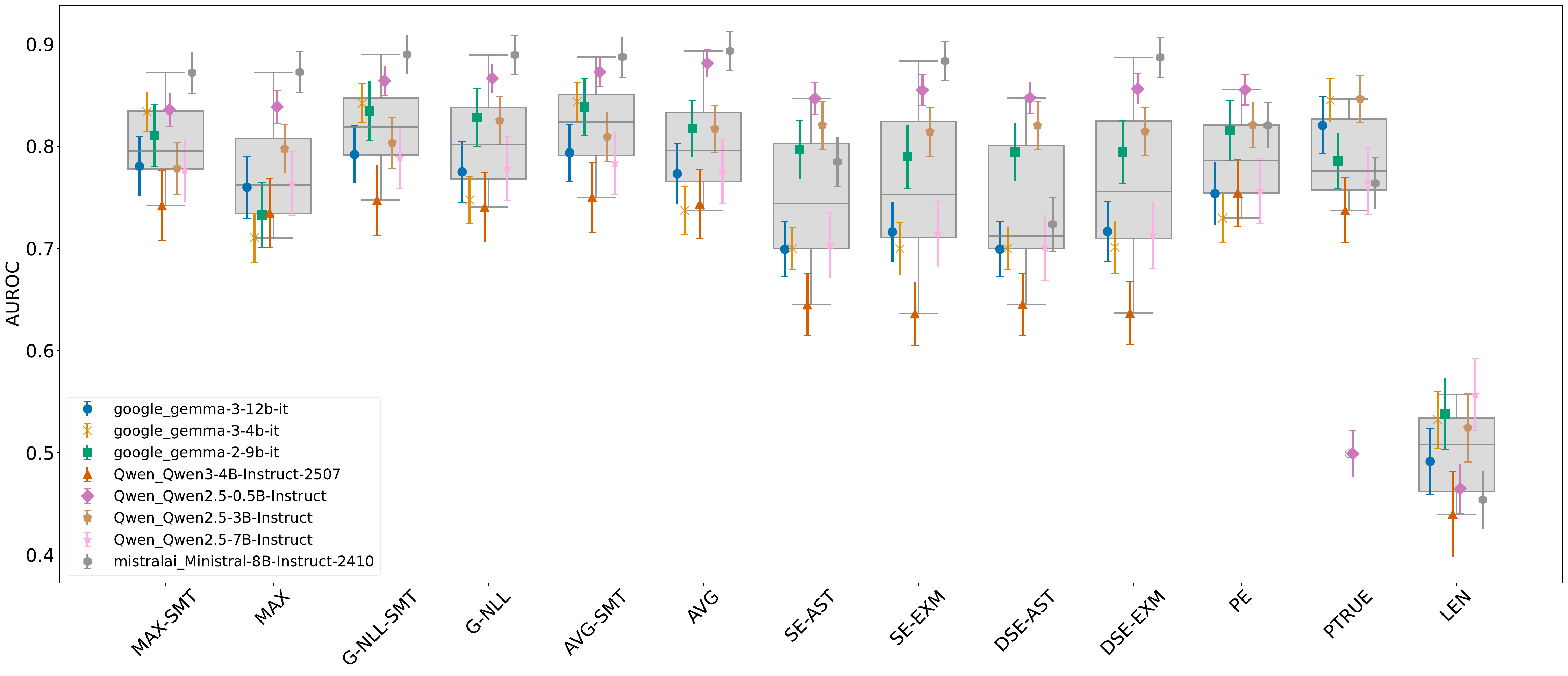}
        \caption{Simple+Irrelevance}
        \label{fig:Simple+Irrelevance}
    \end{subfigure}
    \hfill
    \begin{subfigure}[c]{0.48\textwidth}
        \centering
        \includegraphics[width=\textwidth]{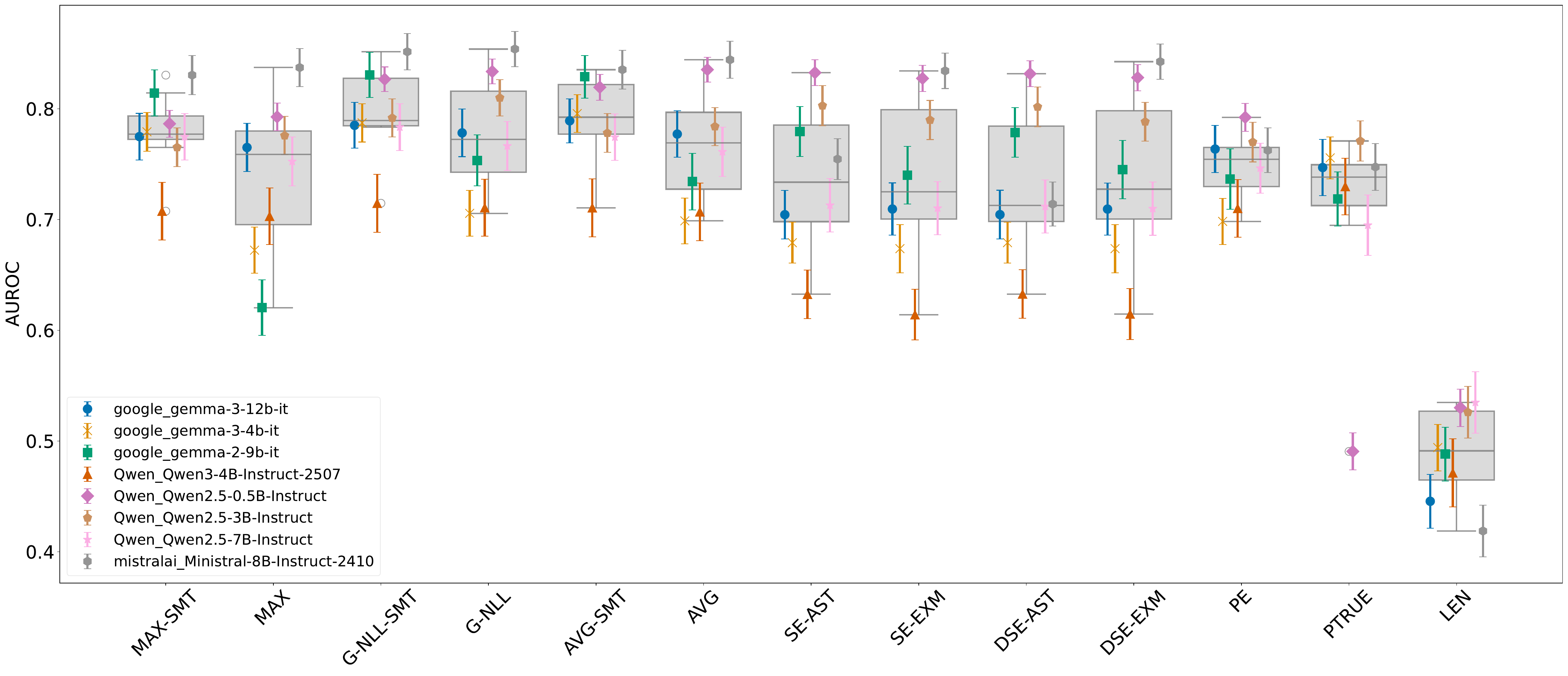}
        \caption{All Combined+Irrelevance}
        \label{fig:All+Irrelevance}
    \end{subfigure}

    \caption{
        Unanswerable requests: 
        AUROC $\pm$ standard error for the combinations of answerable and unanswerable tasks described in \Cref{sec:irrelevance_combinations}. Boxplots summarize the variation across LLMs for a given UQ method.
    }
    \label{fig:irrelevance_combinations}
\end{figure*}

We present the detailed results of each language model for the individual tasks in \Cref{fig:individual_tasks}, for the combined tasks in \Cref{fig:combinations}, and for the combinations that also include the irrelevance split in \Cref{fig:irrelevance_combinations}.

\subsection{Risk-Coverage Curves}
\label{sec:risk_coverage}

\begin{figure*}[b!]
    \centering
    \begin{subfigure}[c]{0.48\textwidth}
        \centering
        \includegraphics[width=\textwidth]{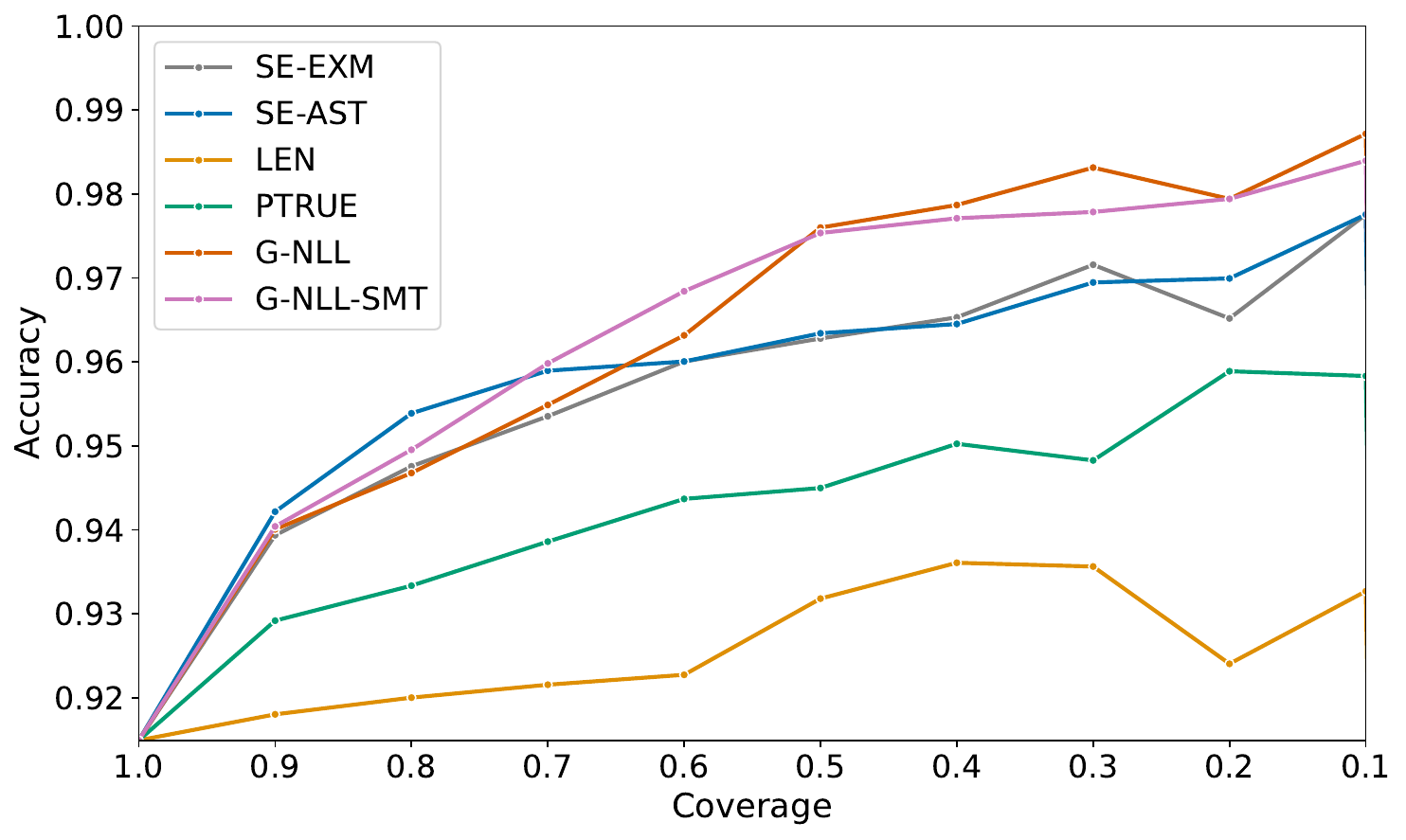}
        \caption{Simple}
        \label{fig:Simple_rc}
    \end{subfigure}
    \hfill
    \begin{subfigure}[c]{0.48\textwidth}
        \centering
        \includegraphics[width=\textwidth]{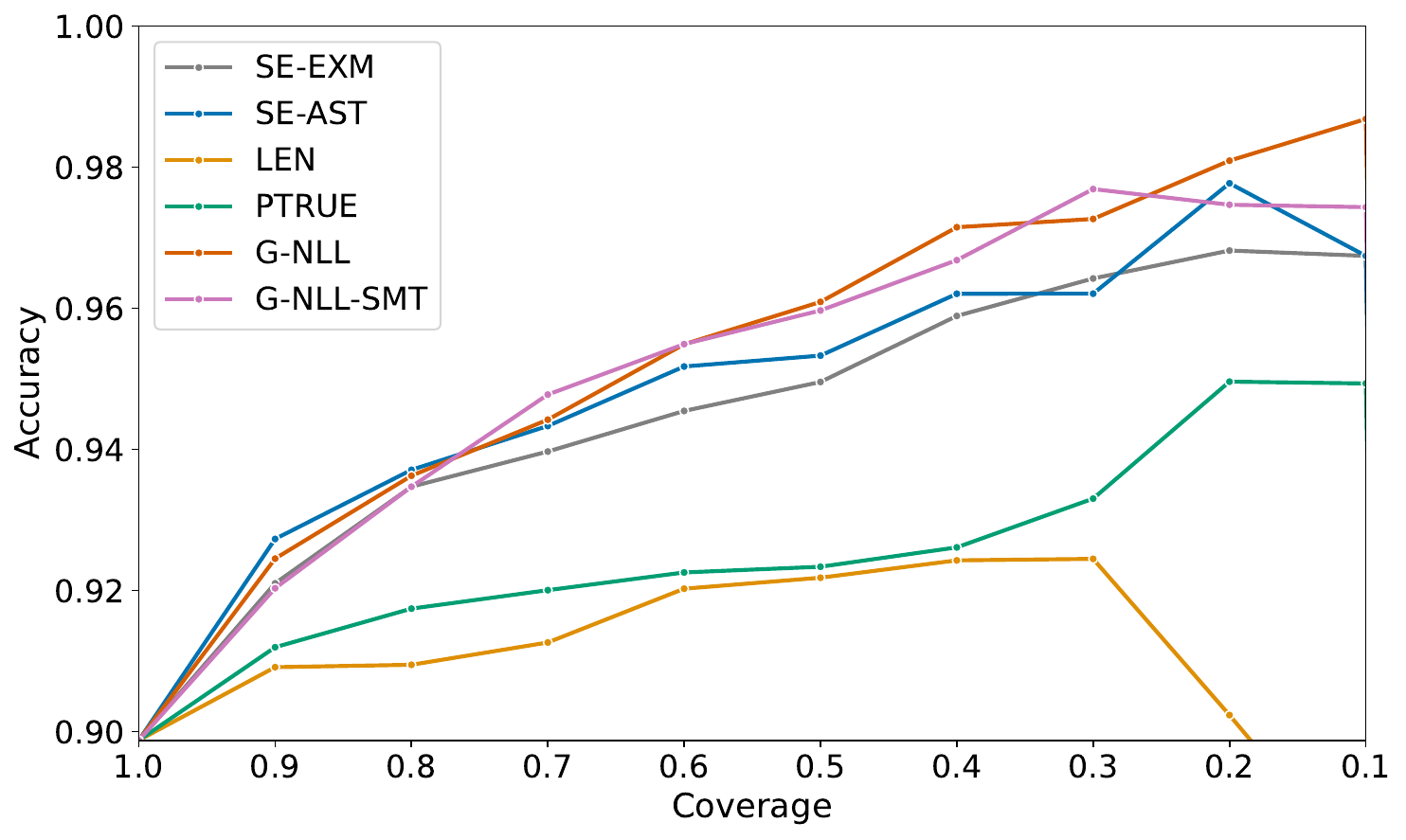}
        \caption{Multiple}
        \label{fig:Multiple_rc}
    \end{subfigure}
    
    \vskip\baselineskip

    \begin{subfigure}[c]{0.48\textwidth}
        \centering
        \includegraphics[width=\textwidth]{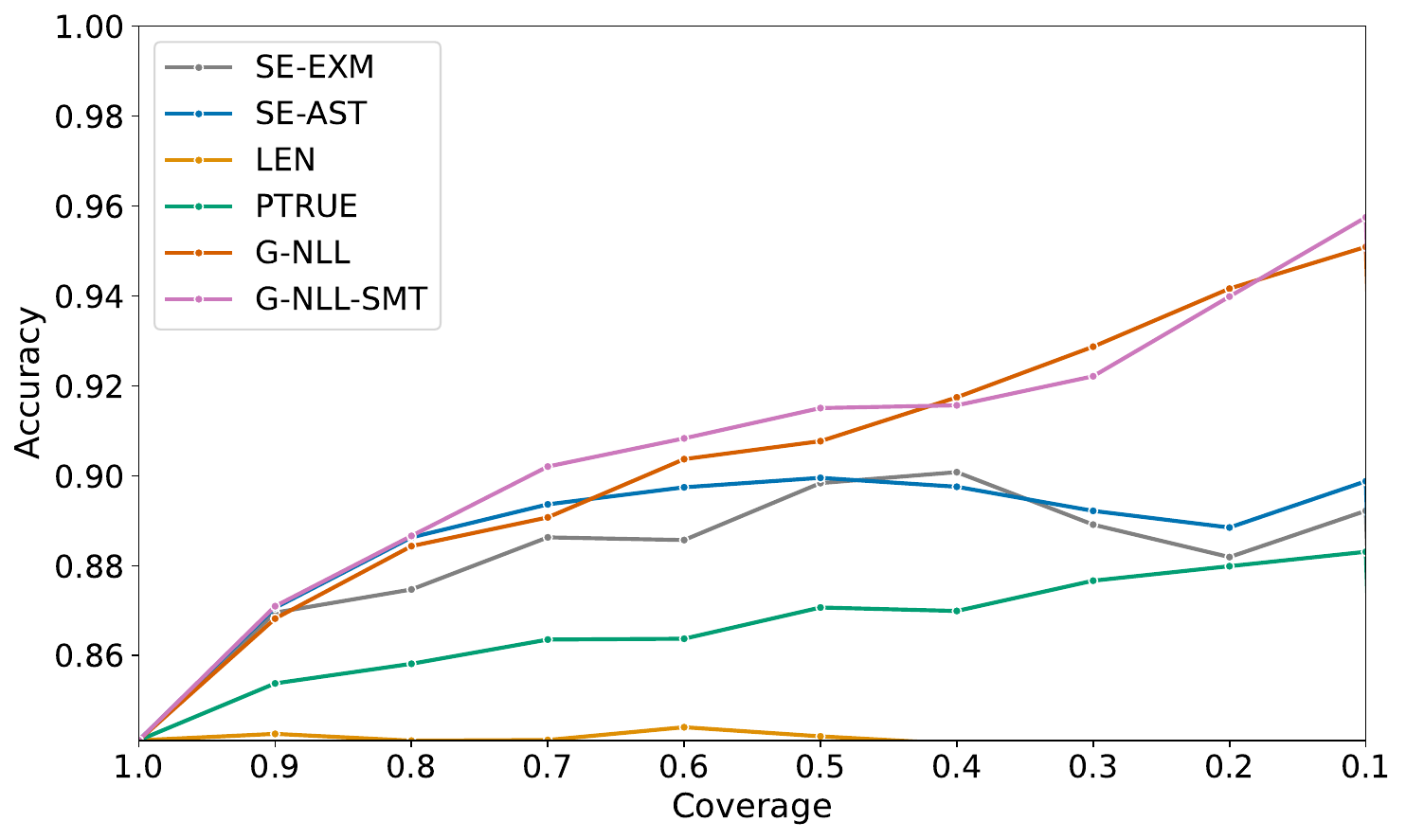}
        \caption{Parallel}
        \label{fig:Parallel_rc}
    \end{subfigure}
    \hfill
    \begin{subfigure}[c]{0.48\textwidth}
        \centering
        \includegraphics[width=\textwidth]{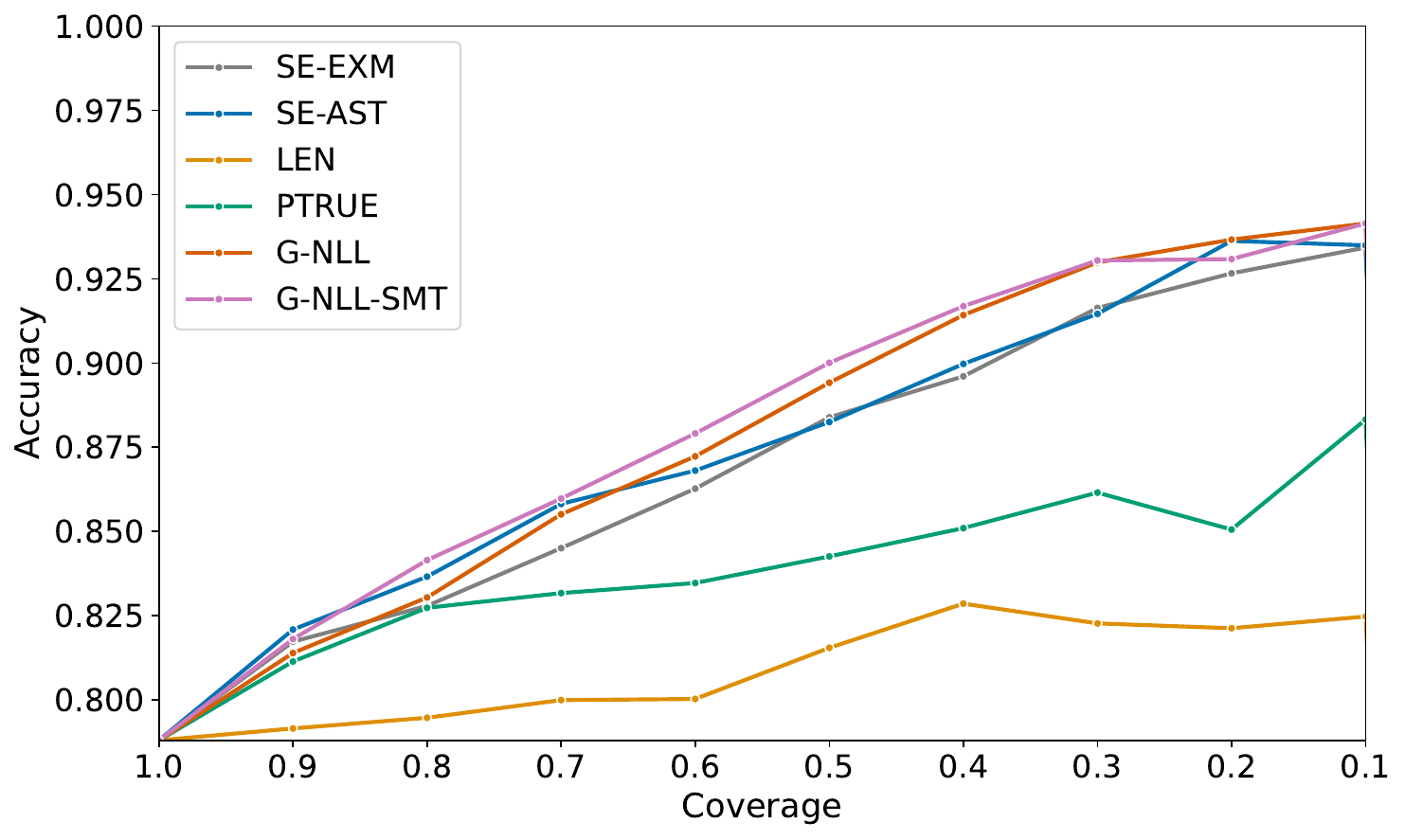}
        \caption{Parallel-Multiple}
        \label{fig:Parallel-Multiple_rc}
    \end{subfigure}

    \caption{
        Individual tasks: 
        Risk-coverage curves for the individual tasks described in \Cref{sec:individual_tasks}, averaged across models. Higher is better.
    }
    \label{fig:individual_tasks_riskcoverage}
\end{figure*}

\begin{figure*}[h!]
    \centering
        \begin{subfigure}[c]{0.48\textwidth}
        \centering
        \includegraphics[width=\textwidth]{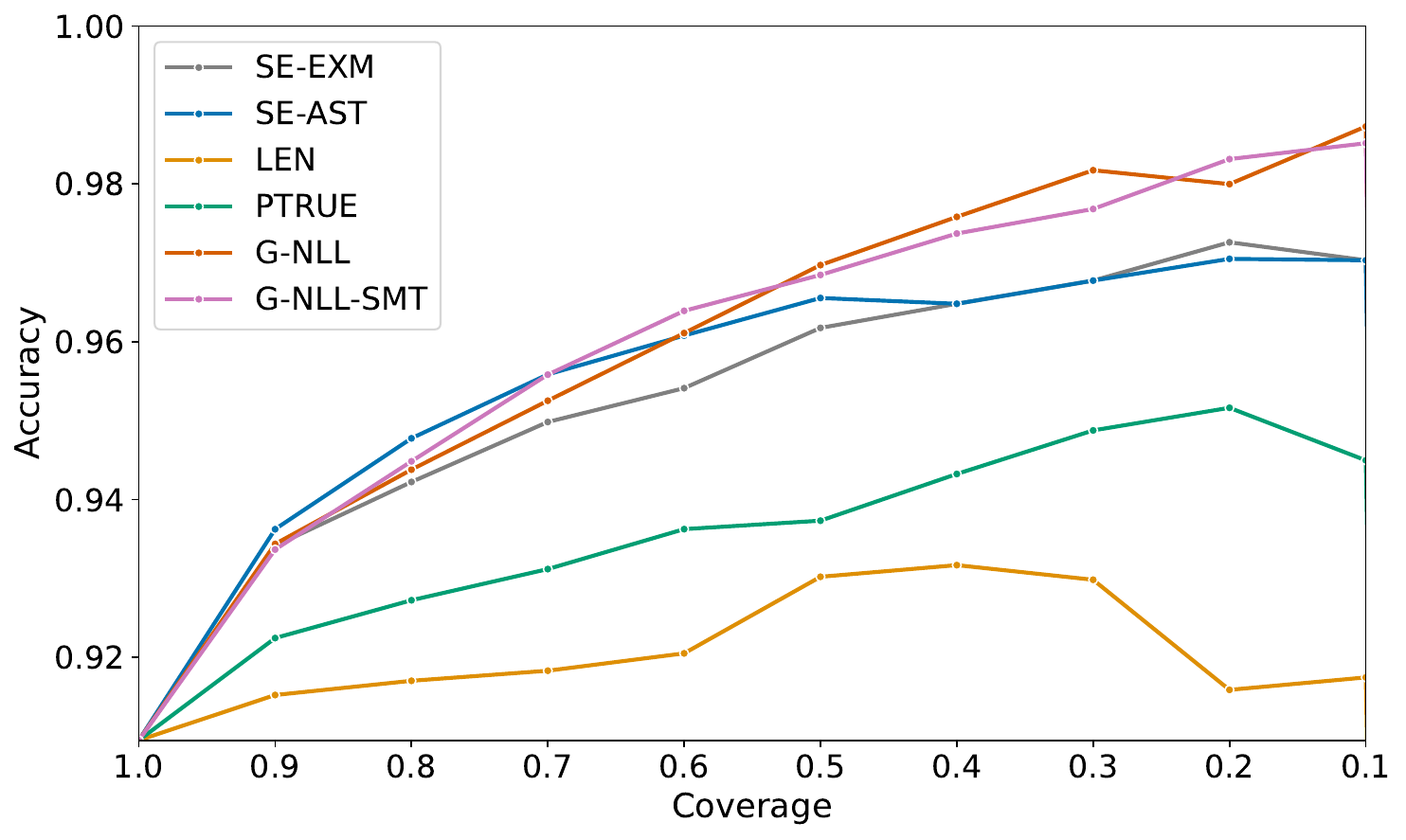}
        \caption{Simple+Multiple}
        \label{fig:Simple+Multiple_rc}
    \end{subfigure}
    \hfill
    \begin{subfigure}[c]{0.48\textwidth}
        \centering
        \includegraphics[width=\textwidth]{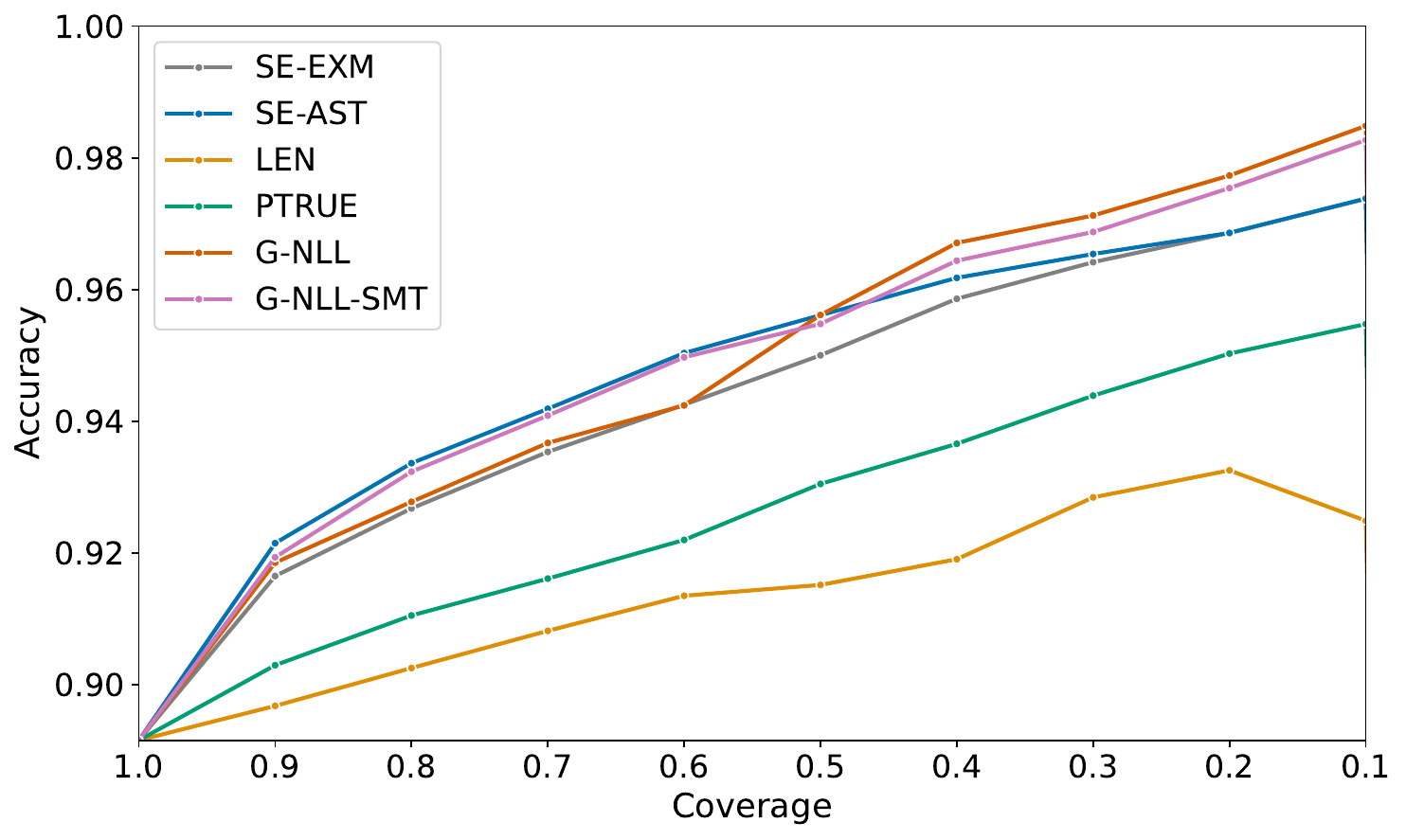}
        \caption{Simple+Parallel}
        \label{fig:Simple+Parallel_rc}
    \end{subfigure}

    \vspace{0.5em}

    \begin{subfigure}[c]{0.48\textwidth}
        \centering
        \includegraphics[width=\textwidth]{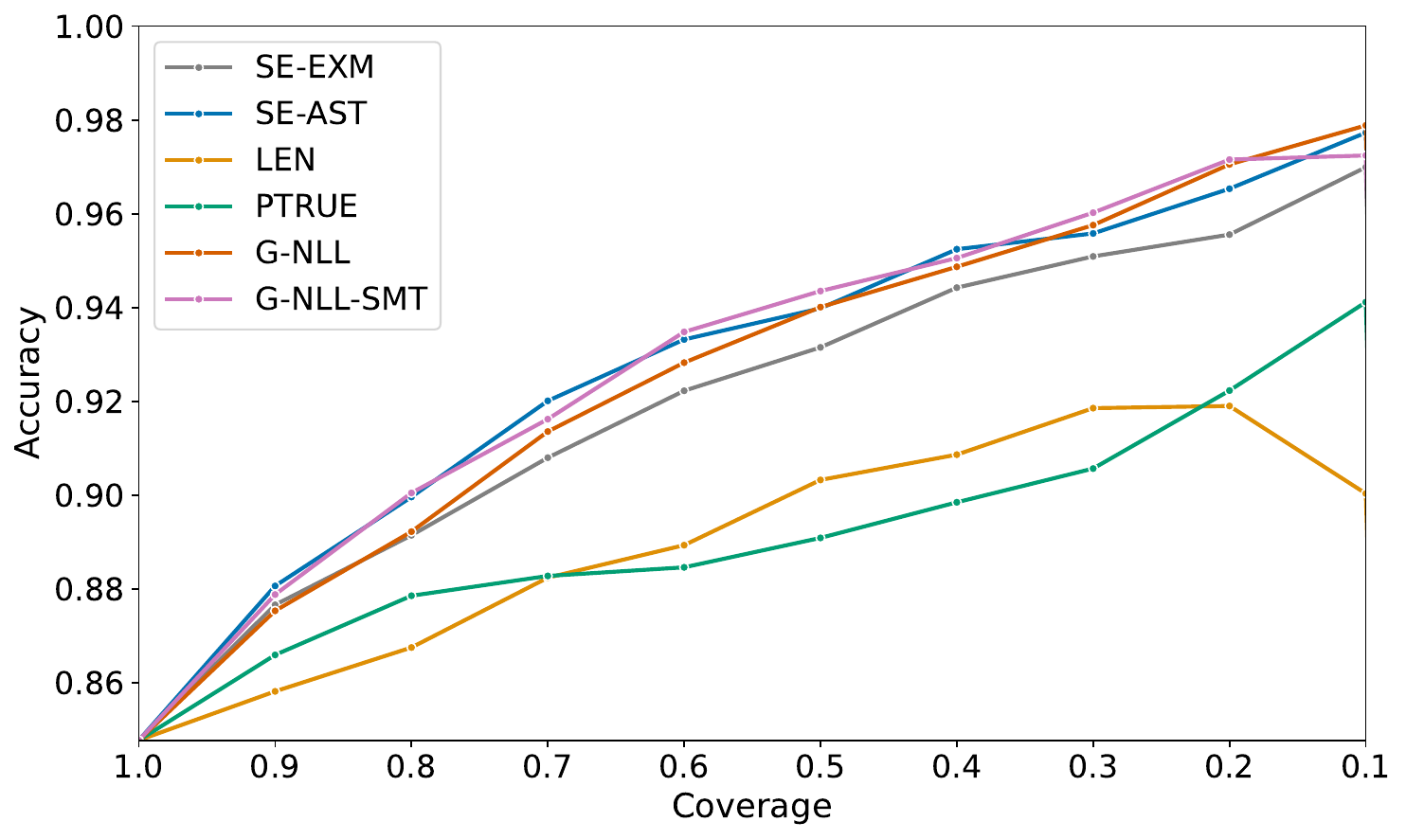}
        \caption{Multiple+\mbox{Parallel-Multiple}}
        \label{fig:Multiple+Parallel-Multiple_rc}
    \end{subfigure}
    \hfill
    \begin{subfigure}[c]{0.48\textwidth}
        \centering
        \includegraphics[width=\textwidth]{figures/rejection/simple-multiple-parallel-parallel_multiple_selective_accuracy_mean.pdf}
        \caption{All Combined}
        \label{fig:All_rc}
    \end{subfigure}

    \caption{
        Combination of tasks: 
        Risk-coverage curves for the combinations of tasks described in \Cref{sec:combinations}, averaged across models. Higher is better.
    }
    \label{fig:combined_tasks_riskcoverage}
\end{figure*}

\begin{figure*}[h!]
\centering
    \begin{subfigure}[c]{0.48\textwidth}
        \centering
        \includegraphics[width=\textwidth]{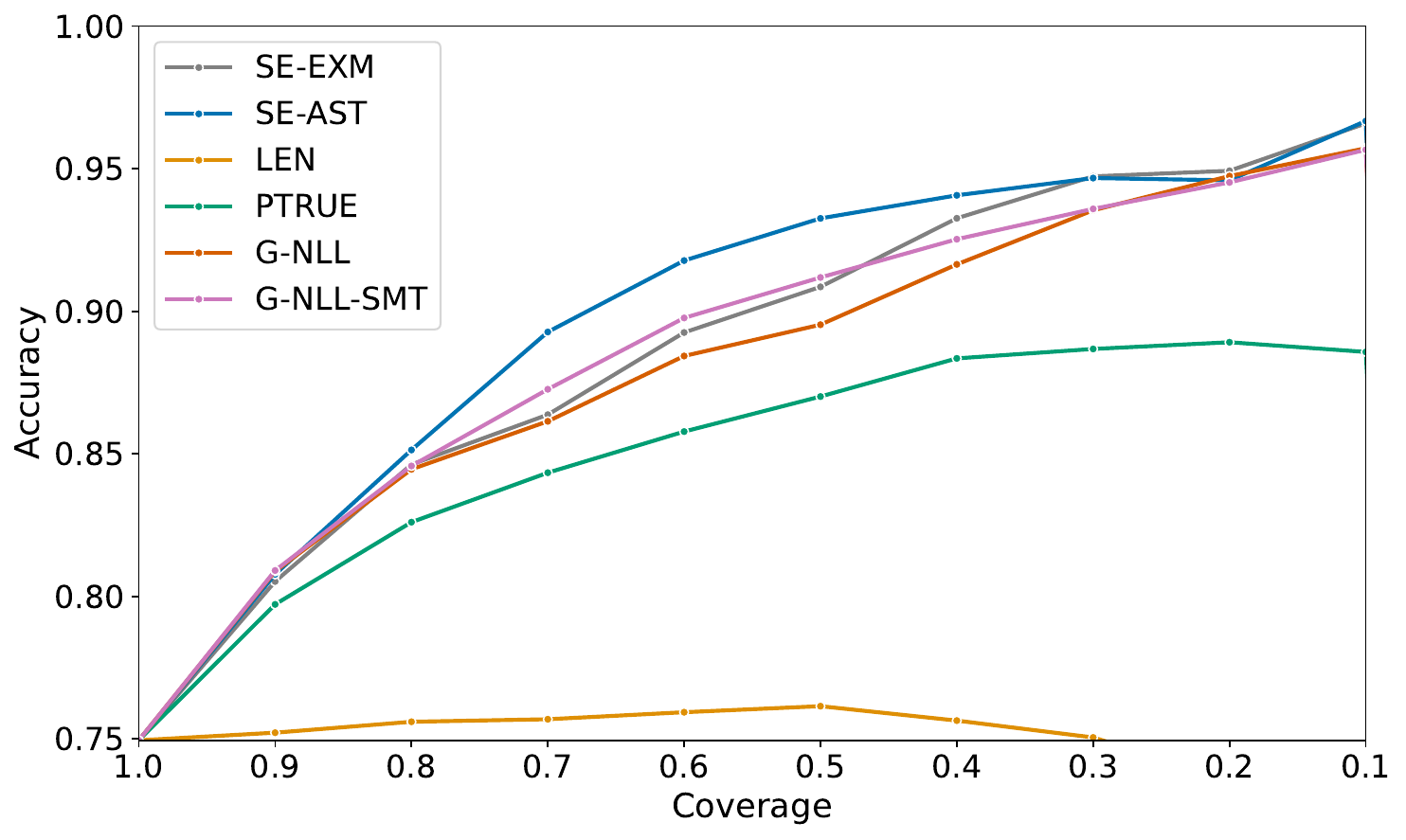}
        \caption{Simple+Irrelevance}
        \label{fig:Simple+Irrelevance_rc}
    \end{subfigure}
    \hfill
    \begin{subfigure}[c]{0.48\textwidth}
        \centering
        \includegraphics[width=\textwidth]{figures/rejection/simple-multiple-parallel-parallel_multiple-irrelevance_selective_accuracy_mean.pdf}
        \caption{All Combined+Irrelevance}
        \label{fig:All+Irrelevance_rc}
    \end{subfigure}

    \caption{
        Unanswerable requests: 
        Risk-coverage curves for the combinations of answerable and unanswerable tasks described in \Cref{sec:irrelevance_combinations}, averaged across models. Higher is better.
    }
    \label{fig:irrelevance_combinations_riskcoverage}
\end{figure*}

As an additional analysis, we include the risk-coverage curves (averaged across models) for the individual tasks in \Cref{fig:individual_tasks_riskcoverage}, for the combined tasks in \Cref{fig:combined_tasks_riskcoverage}, and for the combinations that also include the irrelevance split in \Cref{fig:irrelevance_combinations_riskcoverage}.

We see that indeed, UQ methods can be used to achieve higher accuracy through abstention. For example, not executing a function call for the 30\% of requests with highest G-NLL$_{SMT}$ UQ scores in their output, the accuracies increase from $<92 \%$ to $96 \%$  for \emph{Simple}, from $90 \%$ to $95 \%$ for \emph{Multiple}, from $84 \%$ to $90 \%$ for \emph{Parallel}, and from $79 \%$ to $86 \%$ for \emph{Parallel-Multiple} (\Cref{fig:individual_tasks_riskcoverage}). The gains are similar for the combinations of tasks, and even reach $\approx +10 \%$ once the irrelevance task is added (from $75 \%$ to $85 \%$ and $80\%$ to $90 \%$ in \Cref{fig:irrelevance_combinations_riskcoverage}). 
If one only executes the function calls with the lowest $10 \%$ of G-NLL$_{SMT}$ uncertainty scores, accuracies that can be reached are even higher: $ \approx 98 \%$ for  \emph{Simple} and \emph{Multiple}, $95 \%$ for \emph{Parallel}, and $93 \%$ for \emph{Parallel-Multiple} (\Cref{fig:individual_tasks_riskcoverage}), or $97 \%$ on \emph{All Combined} (\Cref{fig:All_rc}) and $\approx 95 \%$ on the combinations including the irrelevance task (\Cref{fig:irrelevance_combinations_riskcoverage}).

Also, we see some differences by method. P$_{true}$ and LEN clearly lag behind, corresponding to their markedly lower AUROCs in \Cref{sec:benchmark}. Between the remaining methods, the differences are more nuanced. However, for most tasks, the curves for the single-sample methods (G-NLL and G-NLL$_{SMT}$) are above those of the multi-sample methods (SE$_{EXM}$ and SE$_{AST}$), again corresponding to the AUROC results in \Cref{sec:benchmark}. Also, especially for the first half of coverage (until $0.5$), the methods adjusted to the FC setting (G-NLL$_{SMT}$ and SE$_{AST}$) have higher curves than those of their generic counterparts. However, this difference typically disappears in the second half of coverage (from $0.5$ to $0.1$).

\subsection{Detailed Calibration Results}
\label{app:calibration}

\begin{figure*}[h!]
    \centering
        \includegraphics[width=\textwidth, trim=10 10 10 35, clip]{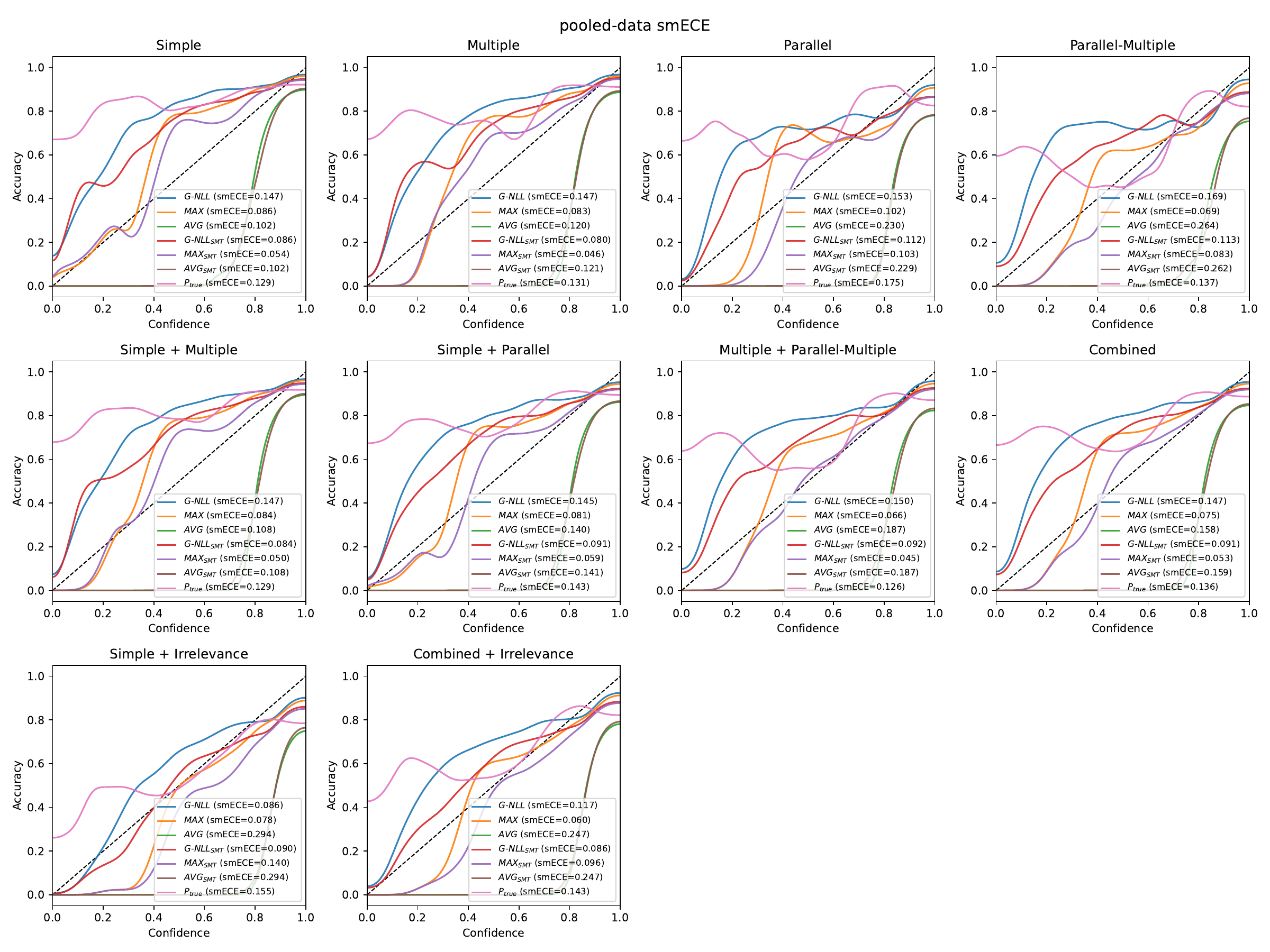}
    \caption{
        smECE for all tasks aggregated over models.
    }
    \label{fig:smoothECE_all}
\end{figure*}

In \Cref{fig:smoothECE_all} we show the calibration results for all individual and combined task datasets for \textbf{MAX}, \textbf{MAX}$_{SMT}$, \textbf{AVG}, \textbf{AVG}$_{SMT}$, \textbf{G-NLL}, \textbf{G-NLL}$_{SMT}$, and \textbf{P}$_{true}$. Other UQ methods do not output probabilities, and their calibration can as such not be evaluated. 

\subsection{Post-Hoc Calibration}
\label{app:posthoc}

As noted in \Cref{sec:calibration}, post-hoc calibration techniques can be applied to a UQ score after the fact. It is worth making precise what this can and cannot change about our conclusions. All of the discrimination and selective-prediction metrics we report---AUROC, as well as the NAURC and PRR results in \Cref{app:selection_pred_metrics}---depend on the UQ scores only through the ranking they induce. Any \emph{strictly} monotone map applied to the scalar uncertainty score, as Platt scaling and score-level temperature scaling are, therefore leaves all of these metrics exactly invariant. Rescaling the token logits before aggregation is not a map of this form, since it can reorder sequences of different lengths, and we leave logit-level recalibration to future work. Isotonic regression is only weakly monotone, and can affect rank-based metrics solely through the ties it introduces. Histogram binning, which maps each bin to its empirical accuracy, is not monotone in general and can therefore also reorder scores across bins. Consequently, post-hoc calibration changes \emph{how} a given uncertainty score should be thresholded, which matters when choosing an abstention threshold for deployment, but it cannot change \emph{which} score is preferable, which is the question our benchmark is designed to answer. The converse also holds: because post-hoc calibration is by construction designed to reduce ECE, it would compress the smECE differences reported in \Cref{sec:calibration}, so it is the calibration comparison between methods---rather than the discrimination comparison---that a calibration step would change. We report uncalibrated values because they characterize the scores as they are produced by the model.

Which methods are amenable to monotone recalibration can be read off \Cref{fig:calibration}. The reliability curves of {G-NLL} and {MAX} are close to monotone, and {AVG} is monotone although nearly flat over most of its range, so all three should recalibrate well. {P}$_{true}$, by contrast, is clearly non-monotone: its accuracy rises to $\approx\!0.75$ around a confidence of $0.2$, falls to $\approx\!0.63$ around $0.5$, peaks near $0.90$ around $0.8$, and then declines. No monotone map can calibrate such a score; doing so would require a non-monotone map such as histogram binning, which reorders the scores and thus effectively yields a different UQ score, outside the invariance argument above. This is consistent with the comparatively weak AUROC of {P}$_{true}$ throughout \Cref{sec:benchmark}.

\newpage
\section{Compute Infrastructure}
\label{app:compute}

\subsection{LLM Inference}
LLM inference to generate outputs for the BFCL tasks as described in \Cref{sec:benchmark} was performed using vLLM on a single node with 4 A100 GPUs. Total runtime to generate the greedy outputs for 1,240 requests across all task datasets, using eight LLMs, was roughly 40 minutes in total. Generating 10 samples at temperature 1 per request required 3 hours in total for all task datasets and models.    

\subsection{Computation of UQ Scores}
Given the outputs to requests generated by the different LLMs, computation of UQ scores was done locally on a 2021 Apple MacBook Pro with M1 Max chip and 64GB RAM. Computations of all UQ scores for any single model and any (individual or combined) task dataset in our benchmark (see \Crefrange{sec:individual_tasks}{sec:irrelevance_combinations}) took less than a minute. This excludes P$_{true}$, which requires an additional LLM forward pass per request, and the entailment-based clustering of \Cref{app:SEentail}, which requires pairwise entailment-model calls. 

\section{AST Decoding Errors and Effective Sample Sizes}
\label{app:ast_errors}

As described in \Cref{sec:individual_tasks}, we exclude requests for which greedy decoding results in an AST decoding error (i.e., the output cannot be interpreted by the Python interpreter). Here, we quantify the frequency of these errors and their impact on our conclusions.

On the \emph{All Combined} split (1,000 data points in total), $3.4\%$ of the outputs are AST decoding errors, averaged across the eight LLMs.
\Cref{tab:effective_sample_sizes} shows the effective sample size (number of successfully parsed outputs) per model.
We observe that the majority of models produce valid outputs for almost all requests, with the exception of gemma-3-4b-it, which has a notably lower effective sample size of 832.

To assess whether excluding these errors affects our conclusions, we compared the UQ method rankings when including AST decoding errors (marking them as incorrect) versus excluding them. The Spearman rank correlation between the two rankings is $0.94$, and the ordering of G-NLL and entropy-based methods stays the same for all.
We further note that a non-decodable function call does not result in an actual function execution, and hence it does not pose a risk from the perspective of preventing potentially harmful function calls.

\begin{table}[h!]
\centering
\setlength{\tabcolsep}{10pt}
\footnotesize
\begin{tabular}{lcc}
\toprule
\textbf{Model} & \textbf{Effective sample size} & \textbf{Retained \emph{Irrelevance} requests} \\
\midrule
Qwen2.5-0.5B-Instruct & 983 & 215 \\
Qwen2.5-3B-Instruct & 980 & 85 \\
Qwen2.5-7B-Instruct & 999 & 59 \\
Qwen3-4B-Instruct-2507 & 997 & 37 \\
Ministral-8B-Instruct-2410 & 987 & 102 \\
gemma-2-9b-it & 987 & 61 \\
gemma-3-4b-it & 832 & 121 \\
gemma-3-12b-it & 962 & 78 \\
\bottomrule
\end{tabular}
\caption{Effective sample sizes (number of successfully AST-parsed outputs) per model on the \emph{All Combined} split (1,000 data points total), and number of \emph{Irrelevance} requests (out of 240) for which the model emitted a function call and which are therefore retained in \emph{Simple+Irrelevance} and \emph{All Combined+Irrelevance} (\Cref{sec:irrelevance_combinations}).}
\label{tab:effective_sample_sizes}
\end{table}

\section{Implementation Details of the Algorithm Extracting Semantically Meaningful Tokens} \label{app:new_method}

In \Cref{fig:algo_python} we give the algorithmic implementation of the classification of semantically meaningful tokens as defined in \Cref{sec:new_method} for the Python-list-style function call outputs by the models \texttt{Qwen/Qwen2.5-\{0.5,3,7\}B-Instruct}, \texttt{Qwen/Qwen3-4B-Instruct-2507}, \texttt{google/gemma-2-9b-it} as well as \texttt{google/gemma-3-\{4,12\}b-it}.
In \Cref{fig:algo_json}, we do the same for the JSON-style function call outputs by
\texttt{mistralai/\allowbreak Ministral-8B-Instruct-2410}.
\Cref{alg:smt-python} restates the Python-style procedure of \Cref{fig:algo_python_p1} as pseudocode, using the same function names.

\begin{center}
\begin{minipage}{0.8\textwidth}
\begin{algorithm}[H]
\caption{SMT typing for Python-call tool outputs}
\label{alg:smt-python}
\begin{algorithmic}[1]
\REQUIRE Token list $T=[t_0,\dots,t_{N-1}]$, each token has \texttt{token\_text}
\ENSURE Same tokens with \texttt{token\_type} $\in \{\texttt{nfp},\texttt{nf},\texttt{np},\texttt{pv},\texttt{-}\}$

\FOR{$i \leftarrow 0$ \TO $N-1$}
  \STATE $t \leftarrow T[i]$
  \STATE $t^{-1} \leftarrow T[i-1]$ \textbf{if} $i>0$ \textbf{else} \texttt{None}
  \STATE $t^{-2} \leftarrow T[i-2]$ \textbf{if} $i>1$ \textbf{else} \texttt{None}

  \IF{$i=0$ \OR $\texttt{check\_is\_num\_of\_func\_or\_param}(t)$}
    \STATE $t.\texttt{token\_type} \leftarrow \texttt{nfp}$
  \ELSIF{$i>0$ \AND $\texttt{check\_is\_param\_value}(t, t^{-1})$}
    \STATE $t.\texttt{token\_type} \leftarrow \texttt{pv}$
  \ELSIF{$i>0$ \AND $\texttt{check\_is\_name\_of\_param}(t, t^{-1}, t^{-2})$}
    \STATE $t.\texttt{token\_type} \leftarrow \texttt{np}$
  \ELSIF{$i>0$ \AND $\texttt{check\_is\_name\_of\_func}(t, t^{-1}, t^{-2})$}
    \STATE $t.\texttt{token\_type} \leftarrow \texttt{nf}$
  \ELSE
    \STATE $t.\texttt{token\_type} \leftarrow \texttt{-}$
  \ENDIF
\ENDFOR

\STATE \textbf{return} $T$
\end{algorithmic}
\end{algorithm}
\end{minipage}
\end{center}

\begin{figure*}[htb]
\centering
    \begin{subfigure}[c]{0.99\textwidth}
        \centering
        \begin{lstlisting}[style=smtcode]
def classify_token_types(list_of_tokens):

    for idx, token in enumerate(list_of_tokens):
        if idx > 0:
            previous_token = list_of_tokens[idx-1]
        preprevious_token = list_of_tokens[idx-2] if idx > 1 else None

        if (idx == 0) or check_is_num_of_func_or_param(token):
            token['token_type'] = 'nfp'  # Aggregating Categories 1., 5., and 6. in (*@\textcolor{green!45!black}{\Cref{sec:new_method}}@*)
        elif idx > 0 and check_is_param_value(token, previous_token):
            token['token_type'] = 'pv' # Category 4. in (*@\textcolor{green!45!black}{\Cref{sec:new_method}}@*)
        elif idx > 0 and check_is_name_of_param(token, previous_token, preprevious_token):
            token['token_type'] = 'np' # Category 3. in (*@\textcolor{green!45!black}{\Cref{sec:new_method}}@*)
        elif idx > 0 and check_is_name_of_func(token, previous_token, preprevious_token):
            token['token_type'] = 'nf' # Category 2. in (*@\textcolor{green!45!black}{\Cref{sec:new_method}}@*)
        else:
            token['token_type'] = '-'

    return list_of_tokens
\end{lstlisting}
        
        \caption{The overall algorithm for classifying the tokens in a sequence of tokens.}
        \label{fig:algo_python_p1}
    \end{subfigure}
    
    \begin{subfigure}[c]{0.99\textwidth}
        \centering
        \begin{lstlisting}[style=smtcode]
def check_is_num_of_func_or_param(token):
    if (
        ('(' in token['token_text'])
        or (')' in token['token_text'])
        or (',' in token['token_text'])):
        return True
    else:
        return False

def check_is_param_value(token, previous_token):
    if (
        ('=' in previous_token['token_text']) or (previous_token['token_type'] == 'pv')
            and (not ('(' in token['token_text']) or (',' in token['token_text']))
        ):
        return True
    else:
        return False

def check_is_name_of_param(token, previous_token, preprevious_token):
    if (
        (previous_token['token_text'].endswith('('))
        or (('(' in token['token_text']) and (not token['token_text'].endswith('(')))
        or ((',' in previous_token['token_text']) and (not '(' in previous_token['token_text'])
            and (preprevious_token['token_type'] == 'pv'))
        ):
        return True
    else:
        return False

def check_is_name_of_func(token, previous_token, preprevious_token=None):
    if (
        ('[' in previous_token['token_text'])
        or ((',' in previous_token['token_text']) and (preprevious_token['token_type'] == 'nfp'))
        or ('),' in previous_token['token_text'])
        or ((',' in token['token_text']) and (token['token_text'] != ',') and (previous_token['token_type'] == 'nfp'))
        ):
        return True
    else:
        return False
\end{lstlisting}
        \caption{The individual token classification functions.}
        \label{fig:algo_python_p2}
    \end{subfigure}

    \caption{
        The algorithmic implementation of the classification of semantically meaningful tokens as defined in \Cref{sec:new_method} for the Python-list-style function call outputs by the models \texttt{Qwen/Qwen2.5-\{0.5,3,7\}B-Instruct}, \texttt{Qwen/Qwen3-4B-Instruct-2507}, \texttt{google/gemma-2-9b-it} as well as \texttt{google/gemma-3-\{4,12\}b-it}.
    }
    \label{fig:algo_python}
\end{figure*}

\begin{figure*}[htb]
\centering
    \begin{subfigure}[c]{0.99\textwidth}
        \centering

        \begin{lstlisting}[style=smtcode]
def classify_token_types_json(list_of_tokens):

    max_idx = len(list_of_tokens)

    for idx, token in enumerate(list_of_tokens):

        preprevious_token = list_of_tokens[idx-2] if idx > 1 else None
        previous_token = list_of_tokens[idx-1] if idx > 0 else None
        two_token_later = list_of_tokens[idx+2] if idx < (max_idx - 2) else None

        if (idx == 0) or check_is_num_of_func_or_param_json(token, two_token_later):
            token['token_type'] = 'nfp' # Aggregating Categories 1., 5., and 6. in (*@\textcolor{green!45!black}{\Cref{sec:new_method}}@*)
        elif idx > 1 and check_is_param_value_or_name_of_func_or_param_json(preprevious_token):
            token['token_type'] = 'pv' # Aggregating Categories 2., 3., and first tokens of Category 4. in (*@\textcolor{green!45!black}{\Cref{sec:new_method}}@*)
        elif idx > 0 and fill_param_value_json(token, previous_token):
            token['token_type'] = 'pv' # All remaining tokens of Category 4. in (*@\textcolor{green!45!black}{\Cref{sec:new_method}}@*)
        else:
            token['token_type'] = '-'

    return list_of_tokens
\end{lstlisting}
        \caption{The overall algorithm for classifying the tokens in a sequence of tokens.}
        \label{fig:algo_json_p1}
    \end{subfigure}
    \begin{subfigure}[c]{0.99\textwidth}
        \centering
        \begin{lstlisting}[style=smtcode]
def check_is_num_of_func_or_param_json(token, two_token_later):
    if two_token_later:
        if (
            ((']' in token['token_text']) and (token['token_text'] != '[TOOL_CALLS]'))
            or ('}}' in token['token_text'])
            or ((',' in token['token_text']) and (two_token_later['token_text'] != 'arguments'))
            ):
            return True
    else:
        if ((']' in token['token_text']) and (token['token_text'] != '[TOOL_CALLS]')) or ('}}' in token['token_text']):
            return True
    return False

def check_is_param_value_or_name_of_func_or_param_json(preprevious_token):
    if (
        (':' in preprevious_token['token_text'])
        or ((',' in preprevious_token['token_text']) and (preprevious_token['token_type'] == 'nfp') and
             ('}}' not in preprevious_token['token_text']) and (']}' not in preprevious_token['token_text']))
        ):
        return True
    else:
        return False

def fill_param_value_json(token, previous_token):
    if (not ('"' in token['token_text'])) and (previous_token['token_type'] == 'pv'):
        return True
    else:
        return False
\end{lstlisting}
        \caption{The individual token classification functions.}
        \label{fig:algo_json_p2}
    \end{subfigure}

    \caption{
        The algorithmic implementation of the classification of semantically meaningful tokens as defined in \Cref{sec:new_method} for the JSON-style function call outputs by \texttt{mistralai/\allowbreak Ministral-8B-Instruct-2410}.
    }
    \label{fig:algo_json}
\end{figure*}

\clearpage

\section{Sensitivity to Temperature and Number of Samples}
\label{app:sensitivity}

In \Cref{sec:SE_AST}, we evaluate multi-sample UQ methods using the default settings of \citet{farquhar2024detec}: temperature $T\!=\!1.0$ and $J\!=\!10$ samples. Here, we ablate these choices for the best-performing multi-sample method, SE$_{AST}$.

\Cref{tab:sensitivity_temperature} shows the change in mean AUROC (over all models) when varying $T$ while keeping $J\!=\!10$ fixed. Lowering $T$ to $0.5$ has a detrimental effect (up to $-0.09$), while increasing to $T\!=\!1.5$ can improve AUROC by up to $0.04$. However, increasing $T$ further undoes these improvements and can even lead to a decrease in AUROC compared to $T\!=\!1.0$. For splits involving the \emph{Irrelevance} task, increasing $T$ to $1.5$ yields almost no improvement and becomes detrimental at higher temperatures as well.

\Cref{tab:sensitivity_nsamples} shows the change in mean AUROC when varying $J$ while keeping $T\!=\!1.0$ fixed. Increasing $J$ to $30$ can improve AUROC by around $0.01$--$0.03$, but improvements flatten out beyond that (no better performance for $J\!=\!40$).

Importantly, adding these changes to the values in \Cref{tab:new_method_se} shows that tuning $T$ or $J$ lets SE$_{AST}$ match or slightly exceed G-NLL (by at most $0.02$) on the eight answerable splits. Against G-NLL$_{SMT}$ (\Cref{tab:new_method}), SE$_{AST}$ at $T\!=\!1.5$ matches or exceeds it on seven of the eight answerable splits (exceeding it only on \emph{Multiple} and \emph{Simple+Parallel}, and falling behind by $0.01$ on \emph{Parallel-Multiple}), whereas at $J\!=\!30$ it matches G-NLL$_{SMT}$ only on \emph{Multiple} and \emph{Multiple+Parallel-Multiple} and falls short by $0.01$--$0.02$ on the remaining six. On the two \emph{Irrelevance} splits, SE$_{AST}$ remains clearly behind both ($0.02$--$0.07$ lower). In light of the added computational cost required for multiple samples (\Cref{sec:SE_AST}), our main conclusion remains unchanged: SE as the best multi-sample UQ method offers no clear advantage over G-NLL as the best single-sample method.

\begin{table}[h!]
\centering
\setlength{\tabcolsep}{10pt}
\footnotesize
\begin{tabular}{lccccc}
\toprule
\textbf{Task} & $T\!=\!0.5$ & $T\!=\!1.0$ & $T\!=\!1.5$ & $T\!=\!2.0$ & $T\!=\!3.0$ \\
\midrule
Simple & $-0.09$ & $0$ & $0.03$ & $0.03$ & $0.02$ \\
Multiple & $-0.06$ & $0$ & $0.03$ & $0.02$ & $0$ \\
Parallel & $-0.01$ & $0$ & $0.04$ & $-0.01$ & $0.02$ \\
Parallel-Multiple & $-0.05$ & $0$ & $0.03$ & $-0.03$ & $-0.03$ \\
Simple+Multiple & $-0.07$ & $0$ & $0.03$ & $0.03$ & $0.01$ \\
Simple+Parallel & $-0.05$ & $0$ & $0.04$ & $0.01$ & $0.01$ \\
Multiple+Parallel-Multiple & $-0.06$ & $0$ & $0.03$ & $-0.01$ & $-0.03$ \\
All Combined & $-0.05$ & $0$ & $0.04$ & $0$ & $-0.01$ \\
Simple+Irrelevance & $-0.07$ & $0$ & $0$ & $0$ & $-0.03$ \\
All Combined+Irrelevance & $-0.06$ & $0$ & $0.01$ & $-0.01$ & $-0.03$ \\
\bottomrule
\end{tabular}
\caption{Differences of mean AUROC of SE$_{AST}$ w.r.t.\ its values for $T\!=\!1.0$. $J\!=\!10$ fixed for all.}
\label{tab:sensitivity_temperature}
\end{table}

\begin{table}[h!]
\centering
\setlength{\tabcolsep}{10pt}
\footnotesize
\begin{tabular}{lccccc}
\toprule
\textbf{Task} & $J\!=\!5$ & $J\!=\!10$ & $J\!=\!20$ & $J\!=\!30$ & $J\!=\!40$ \\
\midrule
Simple & $-0.03$ & $0$ & $0.01$ & $0.02$ & $0.02$ \\
Multiple & $-0.05$ & $0$ & $0.01$ & $0.01$ & $0.01$ \\
Parallel & $-0.04$ & $0$ & $0.02$ & $0.02$ & $0.02$ \\
Parallel-Multiple & $-0.03$ & $0$ & $0.01$ & $0.03$ & $0.03$ \\
Simple+Multiple & $-0.04$ & $0$ & $0.01$ & $0.02$ & $0.02$ \\
Simple+Parallel & $-0.03$ & $0$ & $0.02$ & $0.02$ & $0.02$ \\
Multiple+Parallel-Multiple & $-0.04$ & $0$ & $0.01$ & $0.03$ & $0.03$ \\
All Combined & $-0.03$ & $0$ & $0.02$ & $0.03$ & $0.03$ \\
Simple+Irrelevance & $-0.02$ & $0$ & $0.01$ & $0.02$ & $0.02$ \\
All Combined+Irrelevance & $-0.03$ & $0$ & $0.01$ & $0.02$ & $0.02$ \\
\bottomrule
\end{tabular}
\caption{Differences of mean AUROC of SE$_{AST}$ w.r.t.\ its values for $J\!=\!10$. $T\!=\!1.0$ fixed for all.}
\label{tab:sensitivity_nsamples}
\end{table}

\section{Paired Significance Analysis}
\label{app:paired_significance}
We perform a paired statistical analysis for G-NLL$_{SMT}$ vs.\ G-NLL, and SE$_{AST}$ vs.\ SE$_{EXM}$ to test whether our proposed SMT and AST clustering improve over their baselines. Additionally, we also include a test between the best single-sample method and the best multi-sample method, G-NLL vs.\ SE$_{AST}$. In each pair the first-named method is method A, so that a positive $\Delta$ favors the FC-adapted variant in the first two pairs and the single-sample method in the third. For each task and each pair of UQ methods, we test whether one method's AUROC is systematically higher across models. For every one of the eight LLMs we compute each method's AUROC, giving one paired difference per model, $\Delta\mathrm{AUROC} = \mathrm{AUROC}(\mathrm{A}) - \mathrm{AUROC}(\mathrm{B})$. We apply a two-sided Wilcoxon signed-rank test to these eight paired differences, separately for each task. We report the median across-model $\Delta$AUROC as the effect size together with the two-sided $p$-value (* denotes $p < 0.05$). Note that with eight models, the statistical power is limited, with the smallest attainable two-sided $p = \frac{2}{2^{8}} \approx 0.008$, which is reached only when all eight per-model differences share the same sign. Therefore, the effect sizes (median $\Delta$AUROC) should be considered alongside $p$-values when interpreting the results.

As shown in Table~\ref{tab:wilcoxon}, G-NLL$_{SMT}$ vs.\ G-NLL shows mixed results: the median $\Delta$AUROC is small on every task ($|\Delta| \leq 0.017$) and split in both directions, being negative on four tasks, positive on four, and vanishing to three decimal places on the remaining two, with no task reaching significance. We therefore find no statistical evidence that SMT improves over G-NLL across models. In contrast, SE$_{AST}$ vs.\ SE$_{EXM}$ is positive on nine of the ten task splits, indicating that AST-based clustering consistently improves over EXM-based clustering; the sole exception is \emph{Simple+Irrelevance}, where AST-based clustering is marginally worse ($\Delta = -0.004$, $p = 0.383$). The effect is modest in size but consistent in direction, reaching significance on \emph{Parallel} ($\Delta = +0.014$, $p = 0.023$) and falling just short of it on the largest split (\emph{All Combined}, $\Delta = +0.020$, $p = 0.055$). Finally, comparing the best single-sample method (G-NLL) against the best multi-sample method (SE$_{AST}$), the median $\Delta$AUROC favors G-NLL on all ten splits, significantly so on both unanswerable splits (\emph{Simple+Irrelevance}, $\Delta = +0.061$, $p = 0.008$; \emph{All Combined+Irrelevance}, $\Delta = +0.040$, $p = 0.039$); on the former, $p$ attains the minimum possible value for eight models, i.e.\ G-NLL is better on every model. This shows that, at the default $T$ and $J$, G-NLL, a single-sample greedy method, matches or exceeds the strongest sampling-based estimator, at a fraction of its inference cost.

\begin{table}[b!]
\centering
\setlength{\tabcolsep}{6pt}
\footnotesize
\begin{tabular}{llccc}
\toprule
\textbf{Category} & \textbf{Task} & \textbf{G-NLL$_{SMT}$ vs.\ G-NLL} & \textbf{SE$_{AST}$ vs.\ SE$_{EXM}$} & \textbf{G-NLL vs.\ SE$_{AST}$} \\
\midrule
\multirow{4}{*}{Individual} & Simple & $-0.012\phantom{^{*}}$ (0.844) & $+0.022\phantom{^{*}}$ (0.312) & $+0.023\phantom{^{*}}$ (0.383) \\
 & Multiple & $-0.004\phantom{^{*}}$ (0.945) & $+0.029\phantom{^{*}}$ (0.250) & $+0.006\phantom{^{*}}$ (0.641) \\
 & Parallel & $+0.000\phantom{^{*}}$ (0.578) & $+0.014^{*}$ (0.023) & $+0.033\phantom{^{*}}$ (0.312) \\
 & Parallel-Multiple & $+0.017\phantom{^{*}}$ (0.312) & $+0.020\phantom{^{*}}$ (0.461) & $+0.032\phantom{^{*}}$ (0.461) \\
\midrule
\multirow{4}{*}{Combinations} & Simple+Multiple & $-0.005\phantom{^{*}}$ (0.742) & $+0.020\phantom{^{*}}$ (0.148) & $+0.018\phantom{^{*}}$ (0.461) \\
 & Simple+Parallel & $-0.012\phantom{^{*}}$ (0.844) & $+0.023\phantom{^{*}}$ (0.109) & $+0.018\phantom{^{*}}$ (0.383) \\
 & Multiple+Parallel-Multiple & $+0.009\phantom{^{*}}$ (0.312) & $+0.024\phantom{^{*}}$ (0.195) & $+0.019\phantom{^{*}}$ (0.461) \\
 & All Combined & $-0.000\phantom{^{*}}$ (0.641) & $+0.020\phantom{^{*}}$ (0.055) & $+0.021\phantom{^{*}}$ (0.312) \\
\midrule
\multirow{2}{*}{Unanswerable} & Simple+Irrelevance & $+0.007\phantom{^{*}}$ (0.250) & $-0.004\phantom{^{*}}$ (0.383) & $+0.061^{*}$ (0.008) \\
 & All Combined+Irrelevance & $+0.005\phantom{^{*}}$ (0.383) & $+0.005\phantom{^{*}}$ (0.312) & $+0.040^{*}$ (0.039) \\
\bottomrule
\end{tabular}
\caption{Paired two-sided Wilcoxon signed-rank tests across the eight LLMs. Each cell reports the median across-model $\Delta$AUROC (first method minus second) and its $p$-value; $^{*}$ marks $p<0.05$. A positive $\Delta$ favors the first-named method. Columns: (i) whether SMT improves G-NLL, (ii) whether AST clustering improves semantic entropy, and (iii) the best single-sample (G-NLL) vs.\ the best multi-sample (SE$_{AST}$) method.}
\label{tab:wilcoxon}

\end{table}

\newpage
\section{Black-Box Methods Comparison}
\label{app:black_box}
Beyond the DSE-based methods, which require no log-probabilities, we add self-consistency (SC), or the majority-cluster frequency across the multiple samples under both clustering schemes, in Table~\ref{tab:blackbox}. All four scores are computed from ten sampled generations and their semantic clustering only, without access to token log-probabilities. A higher SC score indicates lower uncertainty. We note that SC matches or slightly underperforms DSE in every cell. This suggests that the black-box signal comes mostly from the clustering, rather than from the choice between an entropy and a majority-vote functional.

\begin{table}[h!]
\centering
\setlength{\tabcolsep}{6pt}
\footnotesize
\begin{tabular}{llcccc}
\toprule
& & \multicolumn{2}{c}{\textbf{EXM clustering}} & \multicolumn{2}{c}{\textbf{AST clustering}} \\
\cmidrule(lr){3-4} \cmidrule(lr){5-6}
\textbf{Category} & \textbf{Task} & $\mathbf{DSE}_{EXM}$ & $\mathbf{SC}_{EXM}$ & $\mathbf{DSE}_{AST}$ & $\mathbf{SC}_{AST}$ \\
\midrule
\multirow{4}{*}{Individual} & Simple & $0.73$ \tiny{±$.050$} & $0.73$ \tiny{±$.049$} & \best{$0.75$} \tiny{±$.047$} & \underline{$0.75$} \tiny{±$.046$} \\
 & Multiple & $0.73$ \tiny{±$.073$} & $0.72$ \tiny{±$.073$} & \best{$0.74$} \tiny{±$.067$} & \underline{$0.74$} \tiny{±$.066$} \\
 & Parallel & $0.66$ \tiny{±$.059$} & $0.66$ \tiny{±$.058$} & \best{$0.69$} \tiny{±$.056$} & \underline{$0.68$} \tiny{±$.055$} \\
 & Parallel-Multiple & $0.73$ \tiny{±$.051$} & $0.72$ \tiny{±$.052$} & \best{$0.73$} \tiny{±$.050$} & \underline{$0.73$} \tiny{±$.049$} \\
\midrule
\multirow{4}{*}{Combinations} & Simple+Multiple & $0.73$ \tiny{±$.041$} & $0.72$ \tiny{±$.040$} & \best{$0.75$} \tiny{±$.038$} & \underline{$0.74$} \tiny{±$.038$} \\
 & Simple+Parallel & $0.70$ \tiny{±$.038$} & $0.70$ \tiny{±$.037$} & \best{$0.73$} \tiny{±$.036$} & \underline{$0.72$} \tiny{±$.036$} \\
 & Multiple+Parallel-Multiple & $0.73$ \tiny{±$.041$} & $0.73$ \tiny{±$.040$} & \best{$0.75$} \tiny{±$.038$} & \underline{$0.75$} \tiny{±$.039$} \\
 & All Combined & $0.72$ \tiny{±$.028$} & $0.72$ \tiny{±$.027$} & \best{$0.74$} \tiny{±$.026$} & \underline{$0.74$} \tiny{±$.026$} \\
\midrule
\multirow{2}{*}{Unanswerable} & Simple+Irrelevance & \best{$0.76$} \tiny{±$.026$} & \best{$0.76$} \tiny{±$.026$} & $0.74$ \tiny{±$.026$} & $0.73$ \tiny{±$.026$} \\
 & All Combined+Irrelevance & \best{$0.74$} \tiny{±$.021$} & $0.73$ \tiny{±$.020$} & \underline{$0.73$} \tiny{±$.020$} & $0.73$ \tiny{±$.020$} \\
\bottomrule
\end{tabular}
\caption{Black-box uncertainty methods: Mean AUROC $\pm$ mean standard error (mean over eight LLMs). All four scores are computed from ten sampled generations and their semantic clustering only, without access to token log-probabilities: discrete semantic entropy (DSE) and self-consistency (SC; the frequency of the majority cluster), each under exact-match (EXM) and AST-based clustering. The best result in \best{green}, \underline{underlined} second best. Note that the slight variation of numbers compared to \Cref{tab:combined_all_tasks} and \Cref{tab:new_method_se} is due to high temperature sampling variations.}
\label{tab:blackbox}
\end{table}

\section{Selective Prediction Metrics}
\label{app:selection_pred_metrics}
Metrics such as AUROC as shown in Table~\ref{tab:combined_all_tasks} measure discrimination, or how well an uncertainty score ranks correct over incorrect predictions, but they do not reflect the utility of the score under abstention. Here we report two selective prediction metrics, the normalized area under the risk-coverage curve (NAURC) and the prediction rejection ratio (PRR).

\paragraph{NAURC.} The area under the risk-coverage curve (AURC) summarizes selective prediction performance across all abstention thresholds from a coverage $c$ of 0 to 1. The model answers a fraction $c$ of inputs with the lowest uncertainty and we compute the corresponding selective risk, which is the error rate among the answered inputs. AURC integrates this risk over coverage. As AURC depends on the model's base error rate, we report a normalized version of this metric, NAURC~\citep{cattelan2024fixbroken}:
$$\mathrm{NAURC} = \frac{\mathrm{AURC} - \mathrm{AURC}^{\star}}{\mathrm{AURC}_{\text{rand}} - \mathrm{AURC}^{\star}},$$
where $\text{AURC}^{\star}$ is the oracle AURC and $\text{AURC}_{\text{rand}}$ is the AURC under random ordering (equal to the base error rate). A lower NAURC is better.

\paragraph{PRR.} We also report the prediction-rejection ratio (PRR) \citep{malinin2021uncertainty}. For a given rejection rate $k$, PRR$@k$ measures the fraction of the oracle's risk reduction that an uncertainty score realizes when rejecting up to the $k\%$ most uncertain inputs:
$$  \mathrm{PRR}@k =
  \frac{A_{\mathrm{random}} - A_{\mathrm{uncertainty}}}
       {A_{\mathrm{random}} - A_{\mathrm{oracle}}},$$
where $A_{\mathrm{uncertainty}}$, $A_{\mathrm{random}}$ and $A_{\mathrm{oracle}}$ are the areas under the error-rejection curve, i.e., the error rate of the retained inputs as a function of the rejection rate, integrated from $0$ to $k\%$, when inputs are rejected in order of decreasing uncertainty, in random order, and by an oracle that rejects incorrect outputs first, respectively.
A PRR of 1 matches the oracle, while 0 matches random rejection. To give an example, a PRR$@k$ of 0.5 means that, of all the error reduction that was achievable by rejecting up to $k$ percent of examples, the uncertainty score captures 50 percent. Here we report PRR$@30\%$ and PRR$@50\%$, which integrate the error-rejection curve as rejection increases from $0$ up to $30\%$ and $50\%$.

We report NAURC in Table~\ref{tab:naurc} and PRR in Tables~\ref{tab:prr30} and~\ref{tab:prr50}. For NAURC, we find a similar trend to the AUROC result in Table~\ref{tab:combined_all_tasks}: the single-sample scores dominate, with G-NLL$_{SMT}$ the best method on 9 of the 10 splits (strictly best on 7, tied with G-NLL on \emph{Simple} and \emph{Simple+Parallel}) and G-NLL the best of the remaining methods on all splits except \emph{Parallel}, where AVG is marginally better ($0.48$ vs.\ $0.49$), while every multi-sample score is clearly behind. The PRR tables refine this picture in a way the AUROC results do not reveal. At PRR$@30\%$, $\text{SE}_{AST}$ outperforms plain G-NLL on all eight answerable splits, so between the two unadapted scores the multi-sample one is indeed preferable in the low-rejection regime; this is exactly the regime in which a deployed system would abstain. Once the FC-adapted single-sample score is included, however, G-NLL$_{SMT}$ recovers most of that gap: it is the best method on 7 of the 10 splits at PRR$@30\%$, with $\text{SE}_{AST}$ retaining an advantage only on \emph{Multiple}, \emph{Simple+Multiple} and \emph{Simple+Parallel}, and it is the best method on all 10 splits at PRR$@50\%$. The FC-specific adaptation therefore matters most precisely where the generic single-sample score was weakest.

Two observations are worth separating here. First, the ordering between G-NLL$_{SMT}$ and G-NLL is far more consistent under these selective-prediction metrics than under AUROC: SMT improves PRR on all ten splits at both rejection rates and NAURC on seven (with two ties and a $0.01$ deterioration on \emph{Simple+Multiple}), whereas its AUROC differences are small and split in both directions (\Cref{app:paired_significance}). This is not a contradiction. AUROC summarizes the ranking over the whole score range, including regions no deployed system operates in, whereas PRR$@k$ depends only on the $k\%$ most uncertain outputs, i.e., on the high-uncertainty region where abstention decisions are actually made; NAURC, which integrates over the full coverage range, accordingly shows a less uniform improvement. A modification that sharpens the ordering among the most uncertain outputs, which is what restricting the score to semantically meaningful tokens is intended to do, can therefore move PRR appreciably while leaving AUROC almost unchanged. Second, we report effect sizes and the consistency of their direction rather than $p$-values for these two metrics. We deliberately do not convert the split-level win counts into a test: the ten splits are overlapping combinations of the same four base task sets, so they are not independent observations and a count-based $p$-value over splits would be invalid. This is precisely why the paired test in \Cref{app:paired_significance} is taken across models rather than across splits. We note finally that NAURC values greater than 1, as well as a negative PRR$@50\%$ for LEN, indicate performance worse than random.

\begin{table}[h!]
\centering
\scalebox{0.58}{
\begin{tabular}{llcccccccccc}
\toprule
& & \multicolumn{4}{c}{\textbf{Single-sample}} & \multicolumn{4}{c}{\textbf{Multi-sample}} & \multicolumn{2}{c}{\textbf{Other}} \\
\cmidrule(lr){3-6} \cmidrule(lr){7-10} \cmidrule(lr){11-12}
\textbf{Category} & \textbf{Task} & \textbf{MAX} & \textbf{AVG} & \textbf{G-NLL} & \textbf{G-NLL}$_{SMT}$ & $\mathbf{SE}_{EXM}$ & $\mathbf{SE}_{AST}$ & $\mathbf{DSE}_{EXM}$ & \textbf{PE} & $\mathbf{P}_{true}$ & \textbf{LEN} \\
\midrule
\multirow{4}{*}{Individual} & Simple & $0.41$ \tiny{±$.098$} & $0.39$ \tiny{±$.103$} & \best{$0.36$} \tiny{±$.097$} & \best{$0.36$} \tiny{±$.113$} & $0.47$ \tiny{±$.145$} & $0.40$ \tiny{±$.150$} & $0.45$ \tiny{±$.144$} & $0.45$ \tiny{±$.106$} & $0.54$ \tiny{±$.128$} & $0.84$ \tiny{±$.155$} \\
 & Multiple & $0.46$ \tiny{±$.144$} & $0.42$ \tiny{±$.142$} & \underline{$0.39$} \tiny{±$.136$} & \best{$0.38$} \tiny{±$.129$} & $0.56$ \tiny{±$.209$} & $0.49$ \tiny{±$.211$} & $0.56$ \tiny{±$.212$} & $0.50$ \tiny{±$.160$} & $0.59$ \tiny{±$.182$} & $0.83$ \tiny{±$.238$} \\
 & Parallel & $0.59$ \tiny{±$.143$} & \underline{$0.48$} \tiny{±$.125$} & $0.49$ \tiny{±$.125$} & \best{$0.45$} \tiny{±$.121$} & $0.72$ \tiny{±$.180$} & $0.68$ \tiny{±$.177$} & $0.73$ \tiny{±$.181$} & $0.56$ \tiny{±$.136$} & $0.74$ \tiny{±$.163$} & $1.15$ \tiny{±$.215$} \\
 & Parallel-Multiple & $0.45$ \tiny{±$.101$} & $0.42$ \tiny{±$.104$} & \underline{$0.38$} \tiny{±$.096$} & \best{$0.34$} \tiny{±$.083$} & $0.49$ \tiny{±$.144$} & $0.47$ \tiny{±$.144$} & $0.49$ \tiny{±$.142$} & $0.45$ \tiny{±$.102$} & $0.57$ \tiny{±$.122$} & $0.87$ \tiny{±$.160$} \\
\midrule
\multirow{4}{*}{Combinations} & Simple+Multiple & $0.42$ \tiny{±$.080$} & $0.40$ \tiny{±$.081$} & \best{$0.36$} \tiny{±$.078$} & \underline{$0.37$} \tiny{±$.090$} & $0.51$ \tiny{±$.118$} & $0.51$ \tiny{±$.121$} & $0.53$ \tiny{±$.120$} & $0.47$ \tiny{±$.088$} & $0.56$ \tiny{±$.109$} & $0.85$ \tiny{±$.136$} \\
 & Simple+Parallel & $0.47$ \tiny{±$.084$} & $0.49$ \tiny{±$.088$} & \best{$0.39$} \tiny{±$.075$} & \best{$0.39$} \tiny{±$.080$} & $0.58$ \tiny{±$.111$} & $0.59$ \tiny{±$.116$} & $0.60$ \tiny{±$.113$} & $0.54$ \tiny{±$.091$} & $0.53$ \tiny{±$.087$} & $0.77$ \tiny{±$.111$} \\
 & Multiple+Parallel-Multiple & $0.43$ \tiny{±$.081$} & $0.47$ \tiny{±$.086$} & \underline{$0.36$} \tiny{±$.078$} & \best{$0.32$} \tiny{±$.069$} & $0.49$ \tiny{±$.120$} & $0.50$ \tiny{±$.118$} & $0.53$ \tiny{±$.120$} & $0.51$ \tiny{±$.087$} & $0.54$ \tiny{±$.101$} & $0.68$ \tiny{±$.127$} \\
 & All Combined & $0.45$ \tiny{±$.059$} & $0.48$ \tiny{±$.061$} & \underline{$0.37$} \tiny{±$.054$} & \best{$0.35$} \tiny{±$.056$} & $0.53$ \tiny{±$.081$} & $0.52$ \tiny{±$.081$} & $0.54$ \tiny{±$.082$} & $0.52$ \tiny{±$.064$} & $0.54$ \tiny{±$.070$} & $0.72$ \tiny{±$.086$} \\
\midrule
\multirow{2}{*}{Unanswerable} & Simple+Irrelevance & $0.37$ \tiny{±$.058$} & \underline{$0.33$} \tiny{±$.055$} & \underline{$0.33$} \tiny{±$.055$} & \best{$0.32$} \tiny{±$.060$} & $0.41$ \tiny{±$.075$} & $0.42$ \tiny{±$.080$} & $0.51$ \tiny{±$.076$} & $0.36$ \tiny{±$.058$} & $0.44$ \tiny{±$.066$} & $1.07$ \tiny{±$.106$} \\
 & All Combined+Irrelevance & $0.42$ \tiny{±$.045$} & $0.38$ \tiny{±$.044$} & \underline{$0.37$} \tiny{±$.042$} & \best{$0.34$} \tiny{±$.043$} & $0.48$ \tiny{±$.060$} & $0.51$ \tiny{±$.061$} & $0.51$ \tiny{±$.059$} & $0.42$ \tiny{±$.046$} & $0.52$ \tiny{±$.054$} & $1.09$ \tiny{±$.082$} \\
\bottomrule
\end{tabular}
}
\caption{Mean NAURC $\pm$ mean standard error (mean over eight LLMs) for individual tasks, task combinations, and unanswerable requests. \textbf{Lower is better}; best result in \best{green}, \underline{underlined} second best.}
\label{tab:naurc}
\end{table}

\begin{table}[h!]
\centering
\scalebox{0.58}{
\begin{tabular}{llcccccccccc}
\toprule
& & \multicolumn{4}{c}{\textbf{Single-sample}} & \multicolumn{4}{c}{\textbf{Multi-sample}} & \multicolumn{2}{c}{\textbf{Other}} \\
\cmidrule(lr){3-6} \cmidrule(lr){7-10} \cmidrule(lr){11-12}
\textbf{Category} & \textbf{Task} & \textbf{MAX} & \textbf{AVG} & \textbf{G-NLL} & \textbf{G-NLL}$_{SMT}$ & $\mathbf{SE}_{EXM}$ & $\mathbf{SE}_{AST}$ & $\mathbf{DSE}_{EXM}$ & \textbf{PE} & $\mathbf{P}_{true}$ & \textbf{LEN} \\
\midrule
\multirow{4}{*}{Individual} & Simple & $0.29$ \tiny{±$.091$} & $0.37$ \tiny{±$.092$} & $0.39$ \tiny{±$.094$} & \best{$0.43$} \tiny{±$.092$} & $0.38$ \tiny{±$.096$} & \underline{$0.42$} \tiny{±$.094$} & $0.38$ \tiny{±$.095$} & $0.29$ \tiny{±$.089$} & $0.28$ \tiny{±$.092$} & $0.05$ \tiny{±$.075$} \\
 & Multiple & $0.26$ \tiny{±$.122$} & $0.34$ \tiny{±$.130$} & $0.39$ \tiny{±$.134$} & \underline{$0.40$} \tiny{±$.129$} & $0.38$ \tiny{±$.133$} & \best{$0.43$} \tiny{±$.132$} & $0.38$ \tiny{±$.137$} & $0.31$ \tiny{±$.128$} & $0.24$ \tiny{±$.124$} & $0.13$ \tiny{±$.123$} \\
 & Parallel & $0.24$ \tiny{±$.107$} & $0.36$ \tiny{±$.109$} & $0.35$ \tiny{±$.111$} & \best{$0.38$} \tiny{±$.113$} & $0.32$ \tiny{±$.113$} & \underline{$0.37$} \tiny{±$.117$} & $0.31$ \tiny{±$.112$} & $0.28$ \tiny{±$.113$} & $0.17$ \tiny{±$.110$} & $0.00$ \tiny{±$.089$} \\
 & Parallel-Multiple & $0.31$ \tiny{±$.106$} & $0.37$ \tiny{±$.105$} & $0.39$ \tiny{±$.106$} & \best{$0.44$} \tiny{±$.105$} & $0.35$ \tiny{±$.111$} & \underline{$0.42$} \tiny{±$.102$} & $0.35$ \tiny{±$.111$} & $0.33$ \tiny{±$.114$} & $0.29$ \tiny{±$.121$} & $0.09$ \tiny{±$.103$} \\
\midrule
\multirow{4}{*}{Combinations} & Simple+Multiple & $0.29$ \tiny{±$.071$} & $0.36$ \tiny{±$.073$} & $0.40$ \tiny{±$.075$} & \underline{$0.42$} \tiny{±$.074$} & $0.38$ \tiny{±$.076$} & \best{$0.44$} \tiny{±$.075$} & $0.38$ \tiny{±$.075$} & $0.30$ \tiny{±$.072$} & $0.26$ \tiny{±$.073$} & $0.07$ \tiny{±$.066$} \\
 & Simple+Parallel & $0.25$ \tiny{±$.069$} & $0.29$ \tiny{±$.070$} & $0.37$ \tiny{±$.072$} & \underline{$0.40$} \tiny{±$.070$} & $0.35$ \tiny{±$.072$} & \best{$0.41$} \tiny{±$.072$} & $0.35$ \tiny{±$.071$} & $0.24$ \tiny{±$.069$} & $0.23$ \tiny{±$.068$} & $0.09$ \tiny{±$.061$} \\
 & Multiple+Parallel-Multiple & $0.29$ \tiny{±$.076$} & $0.27$ \tiny{±$.078$} & $0.41$ \tiny{±$.075$} & \best{$0.46$} \tiny{±$.076$} & $0.39$ \tiny{±$.079$} & \underline{$0.45$} \tiny{±$.077$} & $0.39$ \tiny{±$.079$} & $0.24$ \tiny{±$.080$} & $0.28$ \tiny{±$.085$} & $0.17$ \tiny{±$.073$} \\
 & All Combined & $0.29$ \tiny{±$.050$} & $0.29$ \tiny{±$.051$} & $0.40$ \tiny{±$.051$} & \best{$0.44$} \tiny{±$.051$} & $0.38$ \tiny{±$.052$} & \underline{$0.43$} \tiny{±$.052$} & $0.38$ \tiny{±$.052$} & $0.25$ \tiny{±$.052$} & $0.24$ \tiny{±$.053$} & $0.15$ \tiny{±$.048$} \\
\midrule
\multirow{2}{*}{Unanswerable} & Simple+Irrelevance & $0.50$ \tiny{±$.055$} & $0.58$ \tiny{±$.049$} & \underline{$0.59$} \tiny{±$.051$} & \best{$0.61$} \tiny{±$.050$} & $0.54$ \tiny{±$.053$} & $0.52$ \tiny{±$.055$} & $0.55$ \tiny{±$.052$} & $0.55$ \tiny{±$.053$} & $0.44$ \tiny{±$.055$} & $0.03$ \tiny{±$.055$} \\
 & All Combined+Irrelevance & $0.39$ \tiny{±$.040$} & $0.46$ \tiny{±$.039$} & \underline{$0.48$} \tiny{±$.039$} & \best{$0.52$} \tiny{±$.039$} & $0.45$ \tiny{±$.041$} & $0.45$ \tiny{±$.041$} & $0.45$ \tiny{±$.040$} & $0.43$ \tiny{±$.042$} & $0.32$ \tiny{±$.043$} & $0.01$ \tiny{±$.036$} \\
\bottomrule
\end{tabular}
}
\caption{Mean PRR@30\% $\pm$ mean standard error (mean over eight LLMs) for individual tasks, task combinations, and unanswerable requests. PRR@30\% is the area-based Prediction Rejection Ratio integrated up to 30\% rejection (oracle $=1$, random $=0$; higher is better). Best result in \best{green}, \underline{underlined} second best.}
\label{tab:prr30}
\end{table}

\begin{table}[h!]
\centering
\scalebox{0.58}{
\begin{tabular}{llcccccccccc}
\toprule
& & \multicolumn{4}{c}{\textbf{Single-sample}} & \multicolumn{4}{c}{\textbf{Multi-sample}} & \multicolumn{2}{c}{\textbf{Other}} \\
\cmidrule(lr){3-6} \cmidrule(lr){7-10} \cmidrule(lr){11-12}
\textbf{Category} & \textbf{Task} & \textbf{MAX} & \textbf{AVG} & \textbf{G-NLL} & \textbf{G-NLL}$_{SMT}$ & $\mathbf{SE}_{EXM}$ & $\mathbf{SE}_{AST}$ & $\mathbf{DSE}_{EXM}$ & \textbf{PE} & $\mathbf{P}_{true}$ & \textbf{LEN} \\
\midrule
\multirow{4}{*}{Individual} & Simple & $0.41$ \tiny{±$.092$} & $0.47$ \tiny{±$.092$} & \underline{$0.49$} \tiny{±$.092$} & \best{$0.53$} \tiny{±$.091$} & $0.44$ \tiny{±$.103$} & \underline{$0.49$} \tiny{±$.101$} & $0.43$ \tiny{±$.103$} & $0.38$ \tiny{±$.095$} & $0.35$ \tiny{±$.100$} & $0.07$ \tiny{±$.094$} \\
 & Multiple & $0.36$ \tiny{±$.132$} & $0.43$ \tiny{±$.135$} & \underline{$0.46$} \tiny{±$.137$} & \best{$0.47$} \tiny{±$.135$} & $0.41$ \tiny{±$.145$} & $0.45$ \tiny{±$.143$} & $0.41$ \tiny{±$.147$} & $0.37$ \tiny{±$.139$} & $0.30$ \tiny{±$.142$} & $0.16$ \tiny{±$.141$} \\
 & Parallel & $0.26$ \tiny{±$.115$} & \underline{$0.38$} \tiny{±$.113$} & $0.37$ \tiny{±$.115$} & \best{$0.41$} \tiny{±$.114$} & $0.30$ \tiny{±$.120$} & $0.35$ \tiny{±$.124$} & $0.30$ \tiny{±$.120$} & $0.29$ \tiny{±$.120$} & $0.21$ \tiny{±$.118$} & $0.01$ \tiny{±$.106$} \\
 & Parallel-Multiple & $0.37$ \tiny{±$.104$} & $0.44$ \tiny{±$.104$} & \underline{$0.46$} \tiny{±$.105$} & \best{$0.52$} \tiny{±$.100$} & $0.41$ \tiny{±$.110$} & \underline{$0.46$} \tiny{±$.106$} & $0.40$ \tiny{±$.112$} & $0.37$ \tiny{±$.113$} & $0.31$ \tiny{±$.119$} & $0.08$ \tiny{±$.111$} \\
\midrule
\multirow{4}{*}{Combinations} & Simple+Multiple & $0.39$ \tiny{±$.074$} & $0.45$ \tiny{±$.074$} & $0.48$ \tiny{±$.075$} & \best{$0.51$} \tiny{±$.075$} & $0.43$ \tiny{±$.082$} & \underline{$0.49$} \tiny{±$.081$} & $0.42$ \tiny{±$.082$} & $0.38$ \tiny{±$.077$} & $0.33$ \tiny{±$.080$} & $0.10$ \tiny{±$.079$} \\
 & Simple+Parallel & $0.34$ \tiny{±$.073$} & $0.36$ \tiny{±$.074$} & \underline{$0.43$} \tiny{±$.073$} & \best{$0.47$} \tiny{±$.071$} & $0.37$ \tiny{±$.078$} & $0.42$ \tiny{±$.078$} & $0.37$ \tiny{±$.077$} & $0.29$ \tiny{±$.076$} & $0.30$ \tiny{±$.073$} & $0.13$ \tiny{±$.073$} \\
 & Multiple+Parallel-Multiple & $0.38$ \tiny{±$.078$} & $0.35$ \tiny{±$.082$} & \underline{$0.48$} \tiny{±$.079$} & \best{$0.53$} \tiny{±$.075$} & $0.44$ \tiny{±$.084$} & \underline{$0.48$} \tiny{±$.082$} & $0.43$ \tiny{±$.084$} & $0.30$ \tiny{±$.084$} & $0.33$ \tiny{±$.088$} & $0.23$ \tiny{±$.082$} \\
 & All Combined & $0.37$ \tiny{±$.052$} & $0.36$ \tiny{±$.053$} & \underline{$0.46$} \tiny{±$.052$} & \best{$0.51$} \tiny{±$.051$} & $0.41$ \tiny{±$.055$} & $0.45$ \tiny{±$.056$} & $0.41$ \tiny{±$.056$} & $0.30$ \tiny{±$.055$} & $0.31$ \tiny{±$.057$} & $0.19$ \tiny{±$.054$} \\
\midrule
\multirow{2}{*}{Unanswerable} & Simple+Irrelevance & $0.50$ \tiny{±$.054$} & $0.56$ \tiny{±$.051$} & \underline{$0.57$} \tiny{±$.052$} & \best{$0.60$} \tiny{±$.049$} & $0.50$ \tiny{±$.054$} & $0.47$ \tiny{±$.056$} & $0.52$ \tiny{±$.054$} & $0.53$ \tiny{±$.053$} & $0.47$ \tiny{±$.052$} & $0.02$ \tiny{±$.056$} \\
 & All Combined+Irrelevance & $0.42$ \tiny{±$.041$} & $0.48$ \tiny{±$.041$} & \underline{$0.50$} \tiny{±$.040$} & \best{$0.55$} \tiny{±$.039$} & $0.45$ \tiny{±$.043$} & $0.44$ \tiny{±$.043$} & $0.45$ \tiny{±$.042$} & $0.45$ \tiny{±$.042$} & $0.37$ \tiny{±$.043$} & $-0.01$ \tiny{±$.040$} \\
\bottomrule
\end{tabular}
}
\caption{Mean PRR@50\% $\pm$ mean standard error (mean over eight LLMs) for individual tasks, task combinations, and unanswerable requests. PRR@50\% is the area-based Prediction Rejection Ratio integrated up to 50\% rejection (oracle $=1$, random $=0$; higher is better). Best result in \best{green}, \underline{underlined} second best.}
\label{tab:prr50}
\end{table}

\end{document}